\documentclass{article}
\usepackage{ijcai24}

\usepackage{times}
\usepackage{soul}
\usepackage{url}
\usepackage[hidelinks]{hyperref}
\usepackage[utf8]{inputenc}
\usepackage[small]{caption}
\usepackage{graphicx}
\usepackage{amsmath}
\usepackage{amsthm}
\usepackage{booktabs}
\usepackage{algorithm}
\usepackage{algorithmic}
\usepackage[switch]{lineno}

\usepackage{wrapfig}
\usepackage{float}
\usepackage{subfigure}
\usepackage{amssymb}
\usepackage{mathtools}

\usepackage{xcolor}
\usepackage{color}
\definecolor{seabornBlue}{RGB}{5, 68, 223}
\definecolor{seabornGreen}{RGB}{21, 176, 26}
\definecolor{seabornOrange}{RGB}{249, 115, 6}
\newcommand{\seabornBlue}[1]{\textcolor{seabornBlue}{#1}}
\newcommand{\seabornGreen}[1]{\textcolor{seabornGreen}{#1}}
\newcommand{\seabornOrange}[1]{\textcolor{seabornOrange}{#1}}
\newcommand{\gray}[1]{\textcolor{gray}{#1}}

\usepackage{tcolorbox}
\tcbuselibrary{most}
\newtcolorbox{mybox}[2][]{colbacktitle=red!10!white, colback=gray!10!white,coltitle=black!70!black, title={#2},fonttitle=\bfseries,#1}

\title{ENOTO: Improving Offline-to-Online Reinforcement Learning with Q-Ensembles}

\author{
Kai Zhao$^{1,2}$
\and
Jianye Hao$^{1,}$\thanks{Corresponding author.}\and
Yi Ma$^{1}$\and
Jinyi Liu$^1$\and
Yan Zheng$^1$\And
Zhaopeng Meng$^1$\\
\affiliations
$^1$College of Intelligence and
 Computing, Tianjin University\\
$^2$Bilibili\\
\emails
\{kaizhao, jianye.hao, mayi, jyliu, yanzheng, mengzp\}@tju.edu.cn
}

\begin{document}

\maketitle

\begin{abstract}
Offline reinforcement learning (RL) is a learning paradigm where an agent learns from a fixed dataset of experience. However, learning solely from a static dataset can limit the performance due to the lack of exploration. To overcome it, offline-to-online RL combines offline pre-training with online fine-tuning, which enables the agent to further refine its policy by interacting with the environment in real-time. Despite its benefits, existing offline-to-online RL methods suffer from performance degradation and slow improvement during the online phase. To tackle these challenges, we propose a novel framework called \textbf{EN}semble-based \textbf{O}ffline-\textbf{T}o-\textbf{O}nline (ENOTO) RL. By increasing the number of Q-networks, we seamlessly bridge offline pre-training and online fine-tuning without degrading performance. Moreover, to expedite online performance enhancement, we appropriately loosen the pessimism of Q-value estimation and incorporate ensemble-based exploration mechanisms into our framework. Experimental results demonstrate that ENOTO can substantially improve the training stability, learning efficiency, and final performance of existing offline RL methods during online fine-tuning on a range of locomotion and navigation tasks, significantly outperforming existing offline-to-online RL methods.
\end{abstract}

\section{Introduction}

Reinforcement learning (RL) has shown remarkable success in solving complex decision-making problems, from playing virtual games~\cite{silver2017mastering,vinyals2019alphastar} to controlling tangible robots~\cite{mnih2015human,tsividis2021human,schrittwieser2020mastering}. In RL, an agent learns to maximize the cumulative return from large amount of experience data obtained by interacting with an environment. However, in many real-world applications, collecting experience data can be expensive, time-consuming, or even dangerous. This challenge has motivated the development of offline RL, where an agent learns from a fixed dataset of experience collected prior to learning~\cite{fujimoto2019off,wu2019behavior,bai2022pessimistic,ptr,yu2020mopo,kidambi2020morel}.

Offline RL has several advantages over online RL, including the ability to reuse existing data, the potential for faster learning, and the possibility of learning from experiences that are too risky or costly to collect online~\cite{silver2018general}. However, offline RL also poses significant challenges, such as the potential for overfiting to the training data and the difficulty of ensuring that the learned policy is safe and optimal in the real-world environment. To address these challenges, offline-to-online RL has emerged as an attractive research direction. This approach combines offline pre-training with online fine-tuning using RL, with the goal of learning from a fixed dataset of offline experience and then continuing to learn online in the real-world environment~\cite{nair2020awac,lee2022offline}. Offline-to-online RL has the potential to address the limitations of offline RL, such as the sub-optimality of learned policy. Furthermore, starting with an offline RL policy can achieve strong performance with fewer online environment samples, compared to collecting large amounts of training data by rolling out policies from scratch.

Prior researches have shown that directly initializing an agent with an offline RL method for online fine-tuning can impede efficient policy improvement due to pessimistic learning~\cite{nair2020awac,zhao2022adaptive}. A naive solution to this problem is directly removing the pessimistic term during online training. However, this approach can lead to unstable learning or degraded performance in that the distributional shift between offline datasets and online interactions creates large initial temporal difference errors, causing the oblivion of information learned from offline RL~\cite{lee2022offline,mark2022fine}. Existing offline-to-online RL methods have attempted to address these challenges through implicit policy constraints~\cite{nair2020awac}, filtering offline data used for online fine-tuning~\cite{lee2022offline,mao2022moore,mark2022fine}, adjusting policy constraint weights carefully~\cite{zhao2022adaptive}, or training more online policies~\cite{zhang2023policy}. Nevertheless, these methods still face performance degradation in some tasks and settings, and their performance improvement in the online phase is limited.

\begin{figure*}[t]
    \centering
    \subfigure[Normalized Return]{
        \label{Fig:Motivation_norm_return}
        \includegraphics[width=0.3\textwidth]{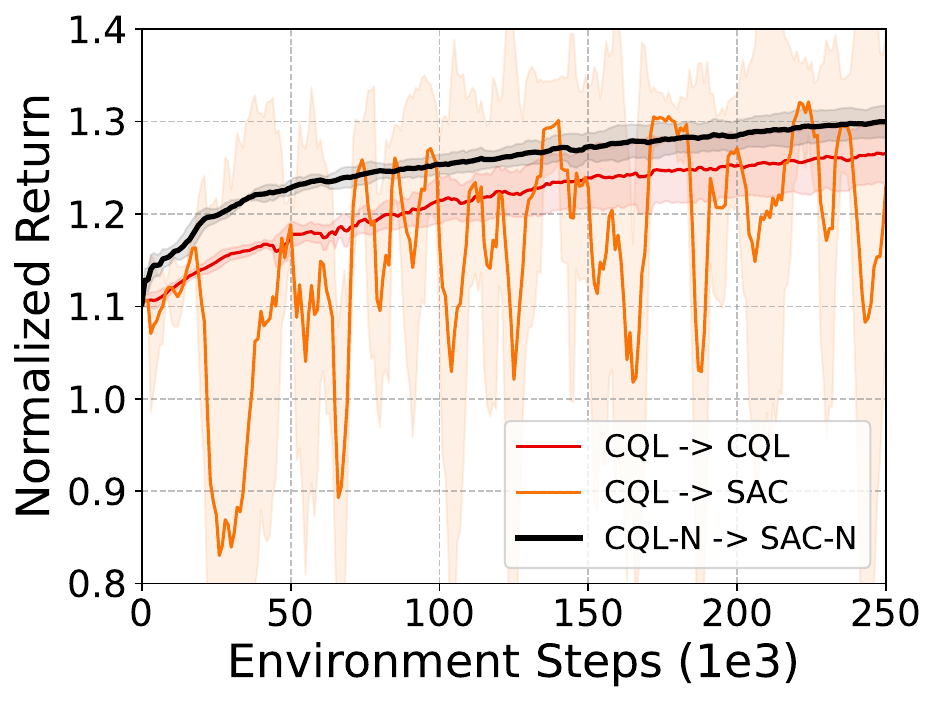}}
    \subfigure[Average Q-value]{
        \label{Fig:Motivation_avg_q}
        \includegraphics[width=0.3\textwidth] {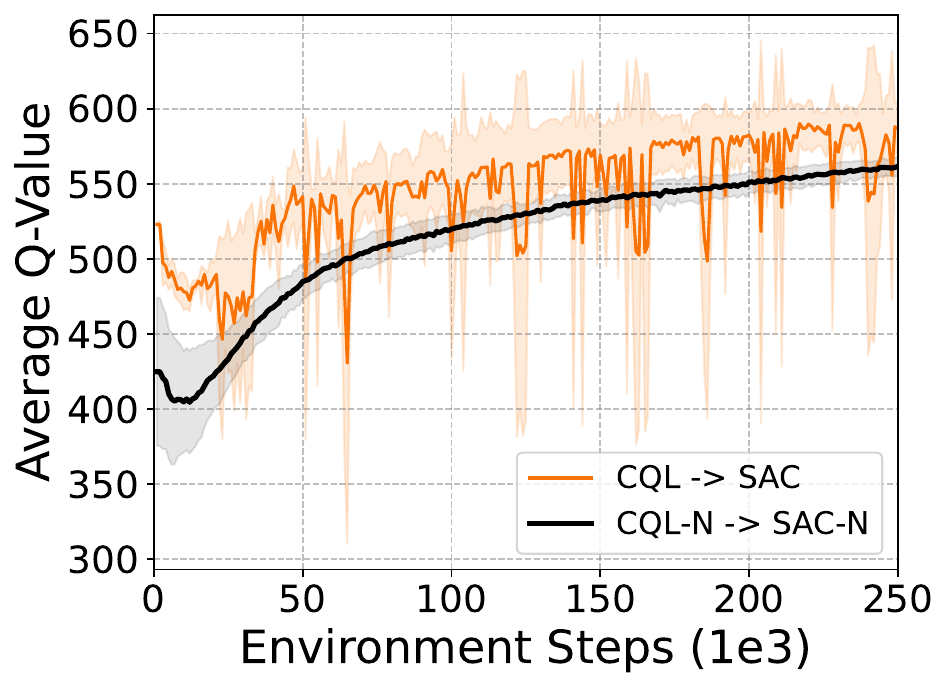}}
    \subfigure[Action Distance]{
        \label{Fig:Motivation_distance}
        \includegraphics[width=0.3\textwidth]{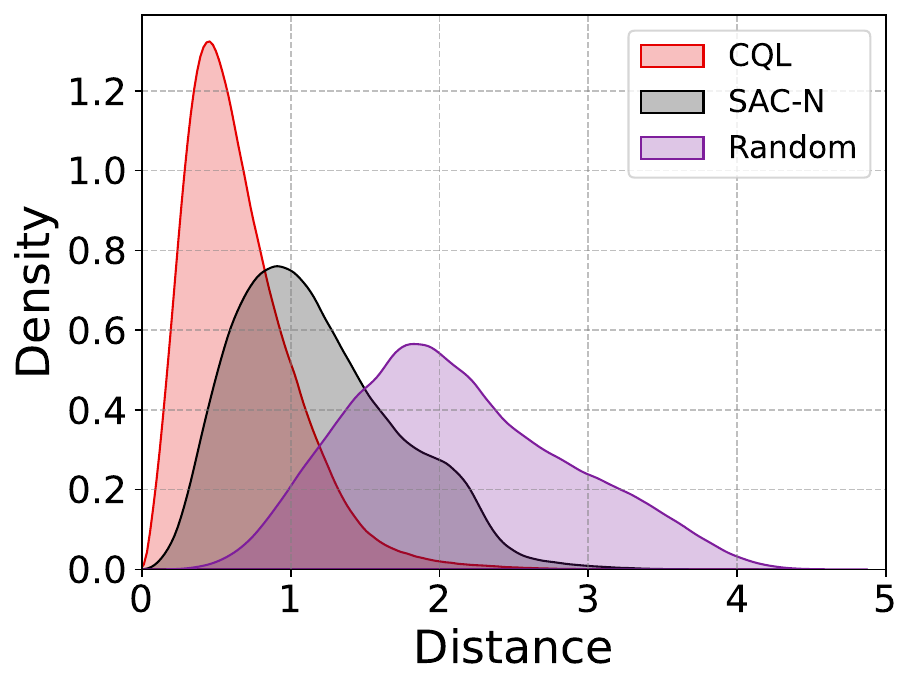}}
    \caption{(a) Normalized return curves of some motivated examples while performing online fine-tuning with offline policy trained on Walker2d-medium-expert-v2 dataset. (b) Comparison of the average Q-values of SAC and SAC-N. (c) Histograms of the distances between the actions from each method (CQL, SAC-N, and a random policy) and the actions from the dataset.}
    \label{Fig:Motivation_analysis}
\end{figure*}

Taking inspiration from leveraging Q-ensembles in offline RL~\cite{an2021uncertainty}, we propose a novel approach to address the challenges of offline-to-online RL. Specifically, we conduct comprehensive experiments by discarding the pessimistic term in existing offline RL algorithms and increasing the number of Q-networks in both offline and online phases. We find that Q-ensembles help to alleviate unstable training and performance degradation, and can serve as a more flexible pessimistic term by encompassing various target computation and exploration methods during the online fine-tuning phase. Based on this discovery, we propose an \textbf{EN}semble-based \textbf{O}ffline-\textbf{T}o-\textbf{O}nline (ENOTO) RL framework that bridges offline pre-training and online fine-tuning. We demonstrate the effectiveness of ENOTO framework by instantiating it on existing offline RL algorithms~\cite{kumar2020conservative,chen2022latent} across diverse benchmark tasks. The main contributions of this work are summarized as follows:

\begin{itemize}
    \item We demonstrate the effectiveness of Q-ensembles in bridging the gap between offline pre-training and online fine-tuning, providing a solution for mitigating the common problem of unstable training and performance drop.
    
    \item We propose a unified framework ENOTO for offline-to-online RL, which enables a wide range of offline RL algorithms to transition from pessimistic offline pre-training to optimistic online fine-tuning, leading to stable and efficient performance improvement. 
    
    \item We empirically validate the effectiveness of ENOTO on various benchmark tasks, including locomotion and navigation tasks, and verify that ENOTO achieves state-of-the-art performance in comparison to all baseline methods.
\end{itemize}

\section{Why Can Q-Ensembles Help Offline-to-Online RL?}
\label{sec_motivation}

To get a better understanding of our ensemble-based framework, we begin with examples that highlight the advantages of Q-ensembles for offline-to-online RL. A natural starting point for offline-to-online RL is to simply initialize the agent with the one trained by an existing offline RL method and then directly perform online fine-tuning without using the offline dataset. However, this approach can hinder efficient online performance improvement due to the inherent pessimism of the offline learning paradigm~\cite{lee2022offline,mark2022fine}. To support this claim, we present CQL~\cite{kumar2020conservative} as a representative and conduct preliminary experiments on the D4RL Walker2d-medium-expert-v2 dataset. The learning curve of CQL during online fine-tuning in Fig.~\ref{Fig:Motivation_norm_return} shows that CQL can maintain the offline performance at the initial stage of online fine-tuning and steadily improve during the training process. This can be attributed to the use of pessimistic Q-functions, which enables the agent to visit states resembling those in the offline dataset and maintain pessimistic towards unseen actions during the initial stage of online fine-tuning. However, the pessimistic objective impedes proper exploration in the online stage and restrict the agent from efficiently improving its performance~\cite{lee2022offline,mark2022fine,exploraton2023jianye,ghasemipour2022so}.

To tackle the aforementioned issue of limited exploration, one might be tempted to remove the conservative estimation component in order to reduce the conservatism of the learning process. However, as shown in Fig.~\ref{Fig:Motivation_norm_return}, this naive solution leads to unstable training or performance degradation when switching from CQL to Soft Actor-Critic (SAC)~\cite{haarnoja2018soft} during online fine-tuning, which has also been reported in previous offline-to-online RL works~\cite{lu2021aw,nair2020awac,lee2022offline,mark2022fine}. The reason is that SAC lacks accurate estimation of Q-values for unknown state-action pairs. Without the conservative constraints of CQL, the Q-values tend to be overestimated, leading to policy misguidance.

So is it possible to find a method that retains suitable pessimistic constraints to mitigate performance degradation, while also tailoring these constraints to be more conducive to exploration during the online phase, rather than being as conservative as traditional offline RL algorithms such as CQL? Inspired by increasing the number of Q-networks in~\cite{an2021uncertainty}, we introduce Q-ensembles and set the number of Q functions in CQL and SAC to N. Specifically, the target Q value is estimated by selecting the minimum value from all the Q-ensembles. We refer to these intermediate methods as CQL-N and SAC-N. Fig.~\ref{Fig:Motivation_norm_return} shows the effectiveness of using SAC-N for online fine-tuning of an offline policy pre-trained with CQL-N. Surprisingly, after incorporating Q-ensembles, we observe that the training becomes more stable and performance drop is no longer observed when switching to online fine-tuning. Moreover, this constraint method not only enhances the final performance of the offline stage, but also improves the efficiency of online learning.

To comprehend the reason behind how Q-ensembles help alleviate unstable training and performance drop, we examine the averaged Q-values over the dataset of different algorithms in Fig.~\ref{Fig:Motivation_avg_q}. We observe that if we directly remove the pessimistic constraints during the online fine-tuning stage (i.e. CQL$\to$SAC), the estimation of the Q-value will fluctuate violently, resulting in unstable training and performance drop, as depicted in Fig.~\ref{Fig:Motivation_norm_return}. However, with our integration of Q-ensembles, SAC-N still has the ability to conservatively estimate, and the variation range of Q-value in CQL-N$\to$SAC-N is much smaller than that of CQL$\to$SAC. This phenomenon indicates that appropriately retaining the conservative capabilities is crucial in avoiding unstable training and performance drop.

We have seen that both SAC-N and CQL can prevent performance drop during online fine-tuning, but why does SAC-N exhibit better performance compared to CQL? To answer this question, we analyze the distance between the actions selected by each method and the actions in the dataset, as shown in Fig.~\ref{Fig:Motivation_distance}. Specifically, we measure  for SAC-N, CQL and a random policy by performing online fine-tuning on the Walker2d-medium-expert-v2 dataset. Our findings reveal that SAC-N has a wider range of action choices compared to CQL, and a more diverse set of actions can lead to improved performance, as stated in previous exploration methods~\cite{ecoffet2021first,lee2021sunrise,ovde,savinov2018episodic,houthooft2016vime}. Therefore, we can incorporate Q-ensembles into existing offline RL algorithms like CQL, and discard the original conservative term designed for offline algorithms during the online phase to improve the online learning efficiency.

To summarize, our primary empirical analysis indicates the following observation:

\begin{mybox}[detach title,before upper={\tcbtitle}]
Q-ensembles can maintain certain conservative capabilities to mitigate unstable training and performance drop, functioning as a more versatile constraint method for exploring more diverse actions during online fine-tuning compared to offline RL algorithms such as CQL.
\end{mybox}

With Q-ensembles in hand, we can further improve online learning efficiency by flexibly leveraging various approaches based on this mechanism, which will be presented in our proposed framework in the following section.

\section{Ensemble-based Offline-to-Online Reinforcement Learning}
\label{sec_method}

Based on the empirical observations discussed earlier, we propose our \textbf{EN}nsemble-based \textbf{O}ffline-\textbf{T}o-\textbf{O}nline (ENOTO) RL Framework. In this section, we first present merits of Q-ensemble using additional empirical results and then progressively introduce more ensemble-based mechanisms into our framework. Although each individual design decision in ENOTO may seem relatively simple, their specific combination outperforms baselines in terms of training stability, learning efficiency and final performance.

\subsection{Q-Ensembles}

\begin{figure}[t]
    \centerline{
        \includegraphics[width=0.8\columnwidth]{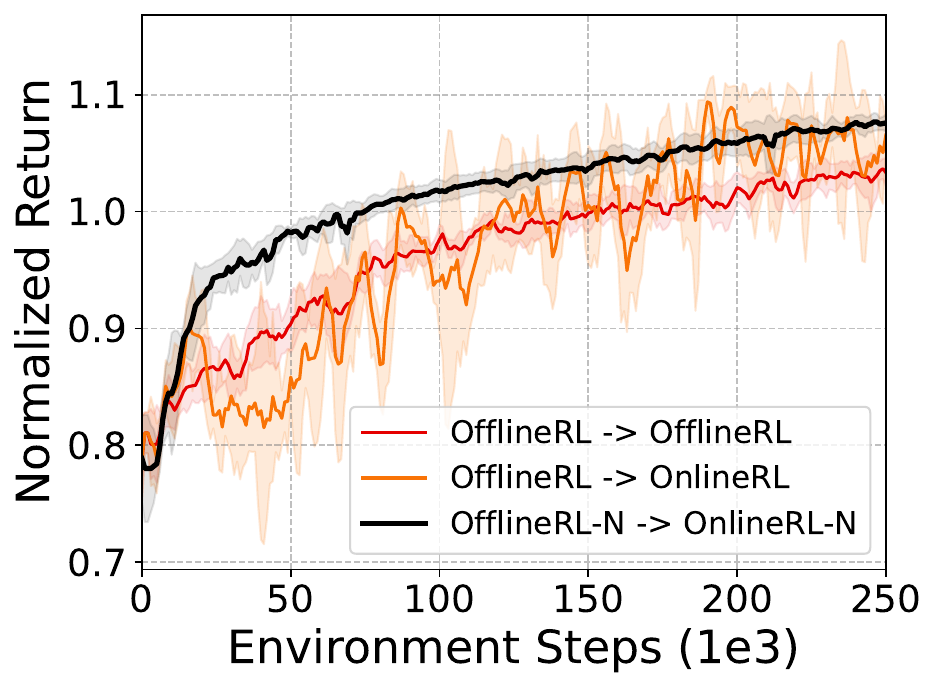}}
    \caption{Aggregated learning curves of different offline-to-online RL approaches on all considered MuJoCo datasets.}
    \label{Fig:Q-ensembles}
\end{figure}

As discussed in the previous section, Q-ensembles can bridge offline and online phases to help pre-trained offline agents perform stable online fine-tuning. In this section, we present comprehensive empirical results to further verify its advantages.

Given an offline RL algorithm named \textit{OfflineRL}, we introduce Q-ensembles to get \textit{OfflineRL-N}, indicating that the algorithm uses $N$ Q-networks and takes the minimum value of all the Q-networks in the ensemble to compute the target. With the pre-trained \textit{OfflineRL-N} agent, we load it as the initialization of the online agent and remove the originally designed pessimistic term (if possible) to obtain \textit{OnlineRL-N}. Then \textit{OnlineRL-N} is trained online. In all methodology sections, we instantiate \textit{OfflineRL} as CQL, and thus \textit{OfflineRL-N} refers to CQL-N, and \textit{OnlineRL-N} refers to SAC-N. To comprehensively verify the effectiveness of Q-ensembles in stabilizing training process and mitigating performance drop, we consider three MuJoCo locomotion tasks~\cite{todorov2012mujoco}: HalfCheetah, Hopper, and Walker2d from the D4RL benchmark suite~\cite{fu2020d4rl}. Specifically, we consider the medium, medium-replay and medium-expert datasets, as in typical real-world scenarios, we rarely use a random policy or have an expert policy for system control.

Fig.~\ref{Fig:Q-ensembles} shows the aggregated normalized return across all nine datasets. Consistent with the results of the previous illustrative experiment, online training of \textit{OfflineRL} is stable but leads to slower asymptotic performance. Directly switching to \textit{OnlineRL} causes unstable training process and performance drop. In contrast, \textit{OfflineRL-N $\to$ OnlineRL-N} no longer experiences performance collapse after switching to online fine-tuning, and the training process is relatively stable. Additionally, \textit{OfflineRL-N $\to$ OnlineRL-N} achieves better fine-tuned performance than \textit{OfflineRL $\to$ OfflineRL}.

Although the ensemble-based method \textit{OfflineRL-N $\to$ OnlineRL-N} has made certain improvements compared to existing method \textit{OfflineRL $\to$ OfflineRL}, it still fails to be improved rapidly in the online stage compared with standard online RL algorithms. Therefore, we shift our focus to analyzing whether we can appropriately loosen the pessimistic estimation of Q-values in the online phase to further improve learning efficiency while ensuring stable training.

\subsection{Loosing Pessimism}

\begin{figure}[t]
    \centerline{
        \includegraphics[width=0.8\columnwidth]{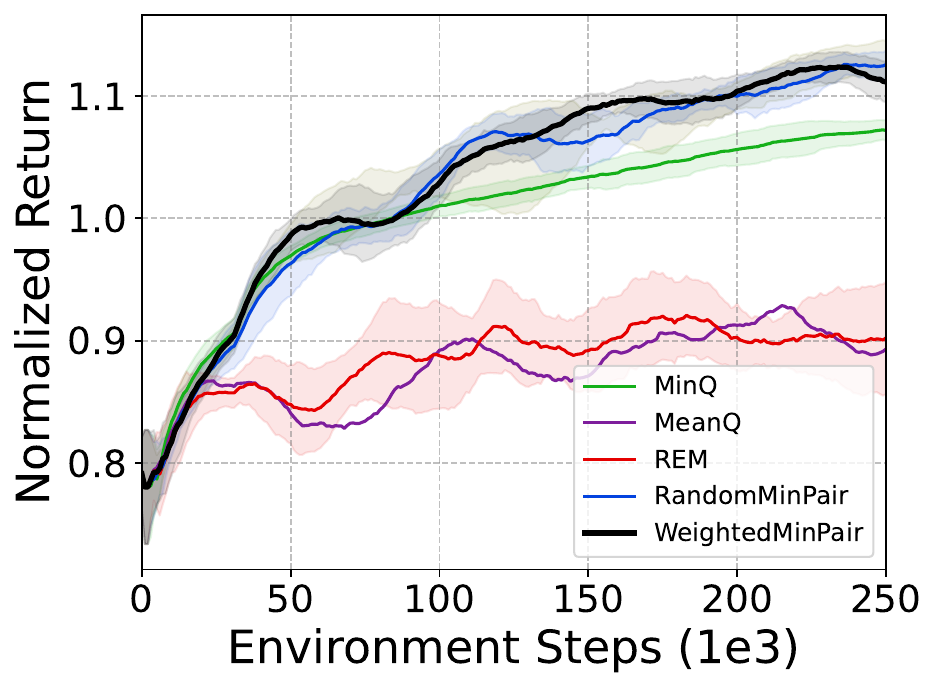}}
    \caption{Aggregated learning curves of \textit{OnlineRL-N} using different Q-target computation methods on all considered MuJoCo datasets.}
    \label{Fig:Target-Computation}
\end{figure}

In the previous section, we employ \textit{OnlineRL-N} as our primary method for the online phase. This method selects the minimum value of $N$ parallel Q-networks as the Bellman target to enforce their Q-value estimates to be conservative. While \textit{OfflineRL-N} $\to$ \textit{OnlineRL-N} has achieved satisfactory performance, selecting the minimum of $N$ Q-networks in the ensemble to compute the Q-target is still too conservative for online training, compared with standard online RL algorithms without pessimistic constraint. Consequently, while ensuring that the online training process is stable, we consider to appropriately loosen the pessimistic estimation of Q-values by modifying the Q-target computation method in \textit{OnlineRL-N} to efficiently improve online performance.

Specifically, we compare several Q-target computation methods. \textbf{(a) MinQ} is what we use in \textit{OnlineRL-N}, where the minimum value of all the Q-networks in the ensemble is taken to compute the target. \textbf{(b) MeanQ} leverages the average of all the Q-values to compute the target. \textbf{(c) REM} is a method originally proposed to boost performance of DQN in the discrete-action setting, which uses the random convex combination of Q-values to compute the target~\cite{agarwal2020optimistic}. It is similar to ensemble average (MeanQ), but with more randomization. \textbf{(d) RandomMinPair} uses a minimization over a random subset 2 of the $N$ Q-functions, which is proposed in prior methods~\cite{chen2021randomized}. \textbf{(e) WeightedMinPair} computes the target as the expectation of all the RandomMinPair targets, where the expectation is taken over all $N$-choose-2 pairs of Q-functions. RandomMinPair can be considered as a uniform-sampled version of WeightedMinPair.

Fig.~\ref{Fig:Target-Computation} presents the results of using different Q-target computation methods in the online phase based on \textit{OnlineRL-N}. With MinQ, which is originally used in \textit{OnlineRL-N}, as the bound, both MeanQ and REM exhibit poor performance, while RandomMinPair and WeightedMinPair outperform the other candidates with their efficient and stable online learning process. As the WeightedMinPair method is more stable on many datasets than the RandomMinPair method, we adopt the WeightedMinPair. Proceeding here, we refer to this intermediate algorithm as \textit{OnlineRL-N + WeightedMinPair}. Despite the superior online fine-tuning performance of this approach, we continue to explore ways to further improve the online learning efficiency by leveraging the ensemble characteristics.

\subsection{Optimistic Exploration}

\begin{figure}[t]
    \centerline{
        \includegraphics[width=0.8\columnwidth]{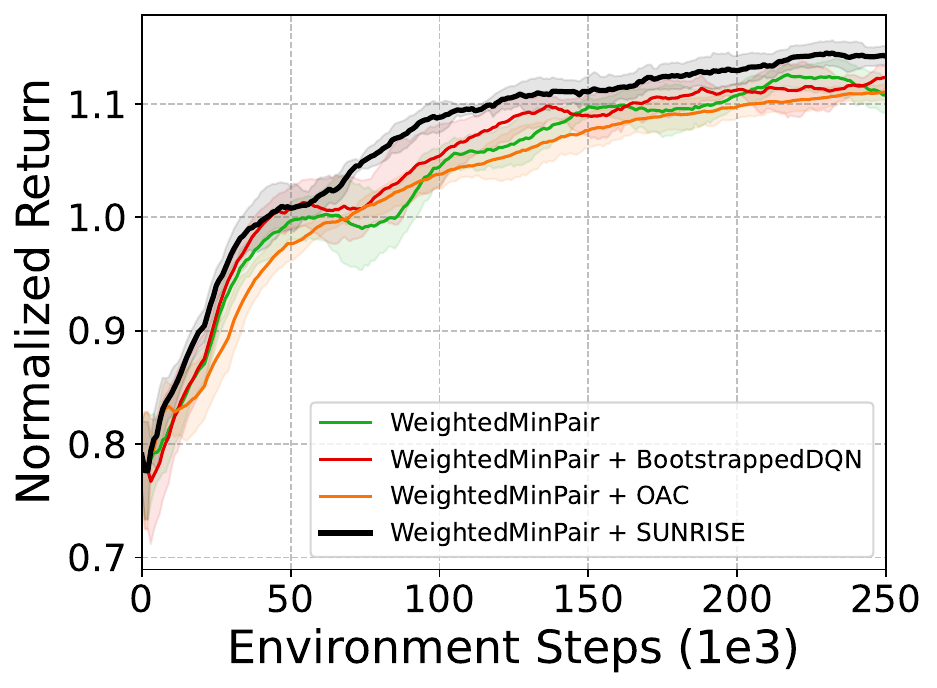}}
    \caption{Aggregated learning curves of \textit{OnlineRL-N + WeightedMinPair} using different exploration methods on all considered MuJoCo datasets.}
    \label{Fig:Exploration}
\end{figure}

In the previous sections, we use pessimistic learning to obtain a satisfactory start point for online learning and gradually loosen the pessimistic constraint to improve online learning. In this section, we investigate the use of ensemble-based exploration methods to further improve performance and learning efficiency.

Specifically, we compare three ensemble-based exploration methods. \textbf{(a) Bootstrapped DQN}~\cite{osband2016deep} uses ensembles to address some shortcomings of alternative posterior approximation schemes, whose network consists of a shared architecture with N bootstrapped “heads” branching off independently. \textbf{(b) OAC}~\cite{ciosek2019better} proposes an off-policy exploration strategy that adjusts to maximize an upper confidence bound to the critic, obtained from an epistemic uncertainty estimate on the Q-function computed with the bootstrap through Q-ensembles. \textbf{(c) SUNRISE}~\cite{lee2021sunrise} presents ensemble-based weighted Bellman backups that improve the learning process by re-weighting target Q-values based on uncertainty estimates.

The results of different exploration methods is presented in Fig.~\ref{Fig:Exploration}. Among them,  \textit{OnlineRL-N + WeightedMinPair + SUNRISE} achieves the highest aggregated return. Consequently, we turn \textit{OnlineRL-N + WeightedMinPair + SUNRISE} into our final ensemble-based framework ENOTO. Algorithm~\ref{algo:main_algorithm} summarizes the offline and online procedures of ENOTO. Note that as many offline RL algorithms can integrate ensemble technique in Q-functions, ENOTO can thus serve as a common plugin. We will further show the plug-and-play character of ENOTO by applying \textbf{\textit{OfflineRL-N} $\to$ \textit{OnlineRL-N + WeightedMinPair + SUNRISE}} on different offline RL algorithms in the experiments. For a comprehensive view of the detailed results of this section, appending the combination of RandomMinPair and different exploration methods, please refer to appendix.

\begin{algorithm}[t]
    \caption{ENOTO: \textbf{EN}semble-based \textbf{O}ffline-\textbf{T}o-\textbf{O}nline RL Framework}
    \label{algo:main_algorithm}
\begin{algorithmic}
    \STATE \textbf{Input:} Offline dataset $D_{offline}$, offline RL algorithm \textit{OfflineRL}
    
    \textbf{Output:} Offline to online learning algorithm
    
    \gray{// \textbf{Offline Phase}}
    
    Turning offline RL algorithm \textit{OfflineRL} into \textit{OfflineRL-N} with integration of Q-ensembles.
    
    Training \textit{OfflineRL-N} using $D_{offline}$

    \gray{// \textbf{Online Phase}}
    
    Removing original pessimistic term in \textit{OfflineRL} (if possible) and thus turn \textit{OfflineRL-N} to \textit{OnlineRL-N} 

    Setting the Q-target computation method to \textit{WeightedMinPair} and obtain \textit{OnlineRL-N + WeightedMinPair}

    Introducing \textit{SUNRISE} to encourage exploration and obtain \textit{OnlineRL-N + WeightedMinPair + SUNRISE}

    \textbf{return} \textit{OfflineRL-N} $\to$ \textit{OnlineRL-N + WeightedMinPair + SUNRISE}
\end{algorithmic}
\end{algorithm}

\begin{figure*}[t]
    \centering
    \includegraphics[width=0.82\textwidth]{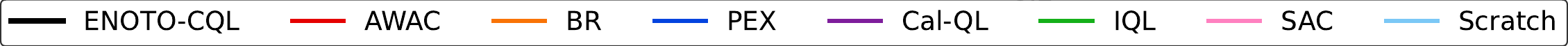}
    \includegraphics[width=0.3\textwidth]{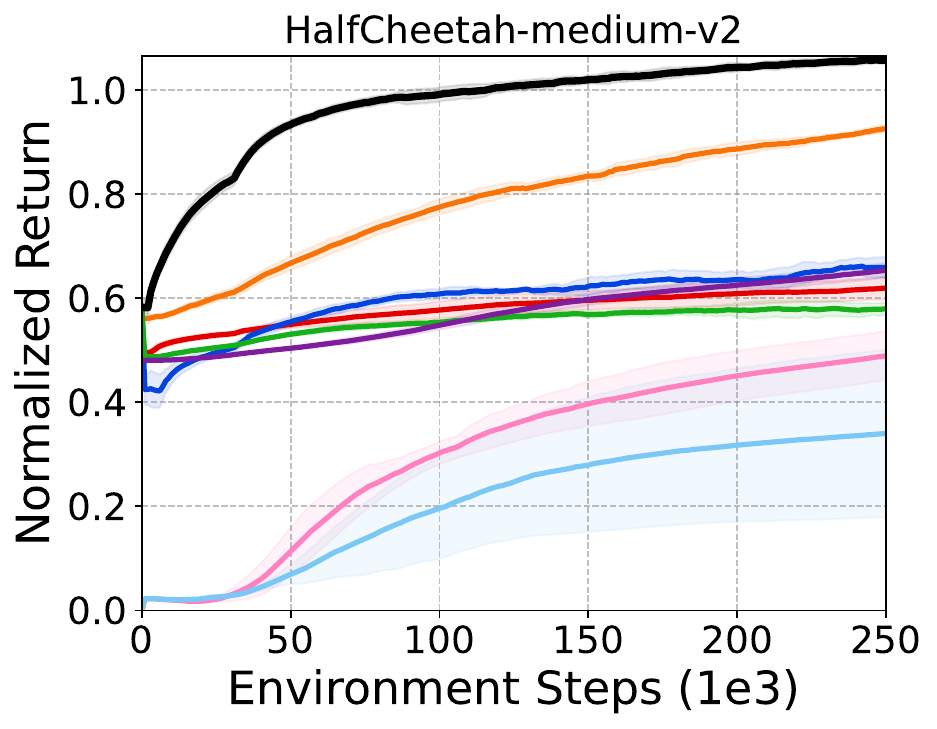}
    \includegraphics[width=0.3\textwidth]{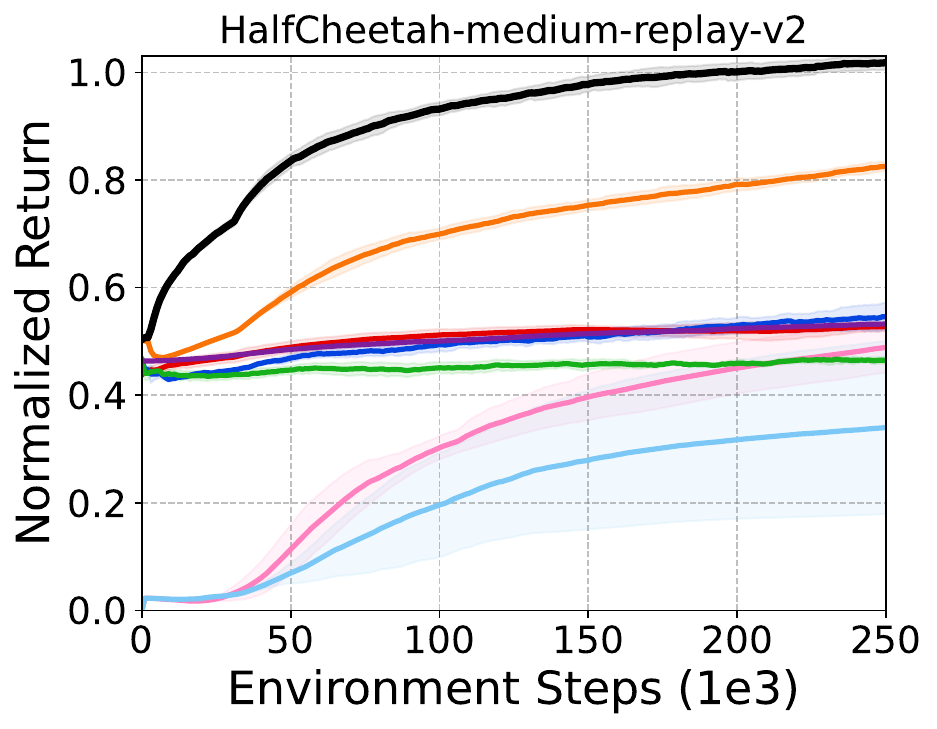}
    \includegraphics[width=0.3\textwidth]{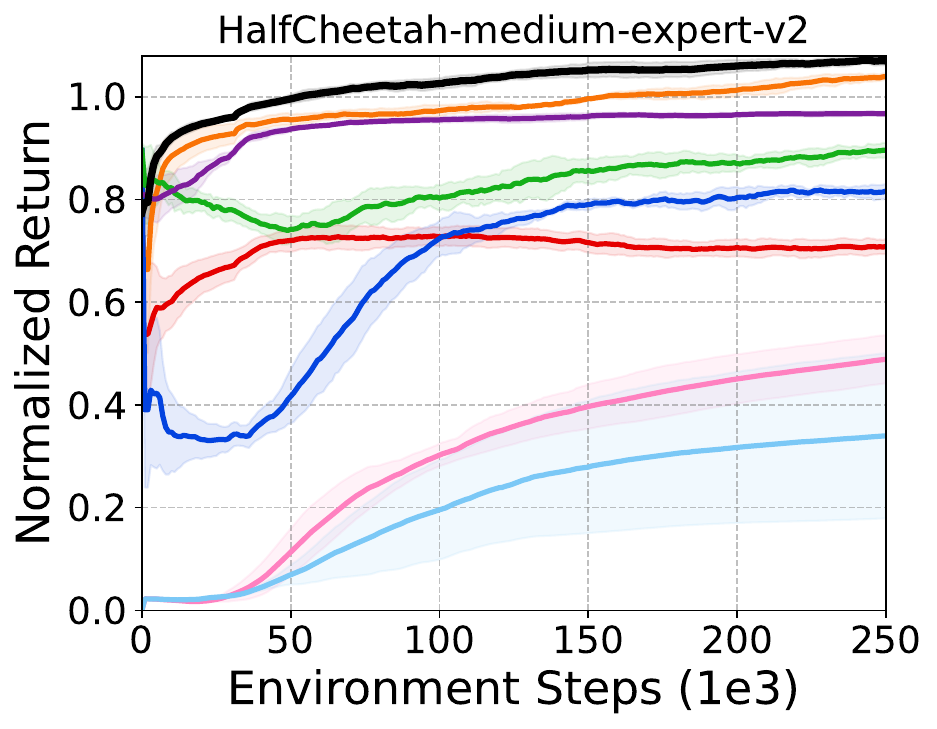}
    \includegraphics[width=0.3\textwidth]{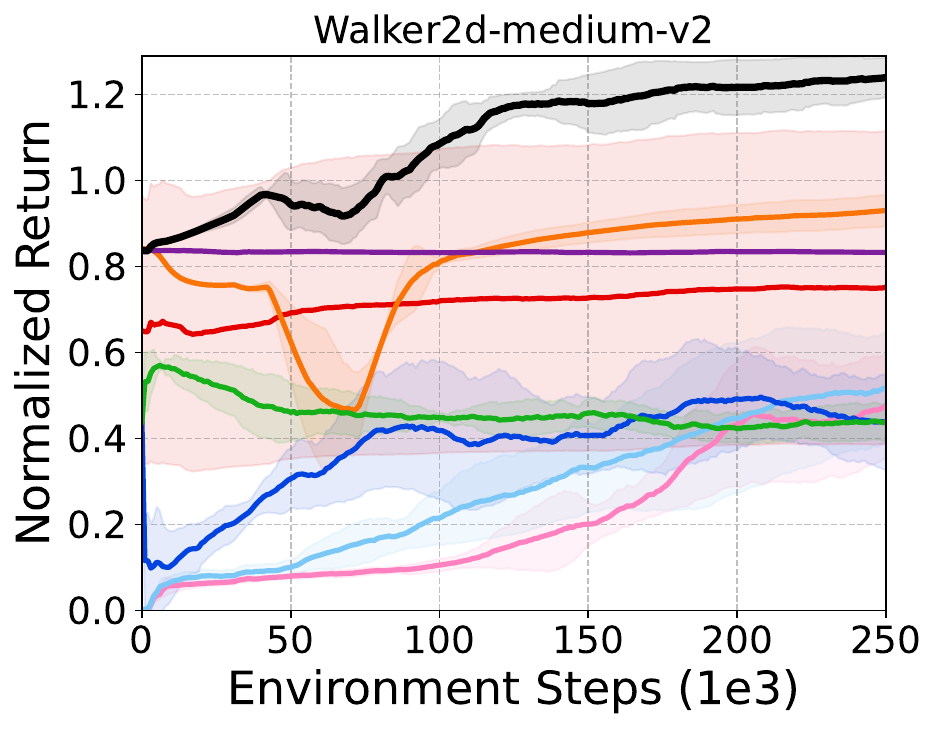}
    \includegraphics[width=0.3\textwidth]{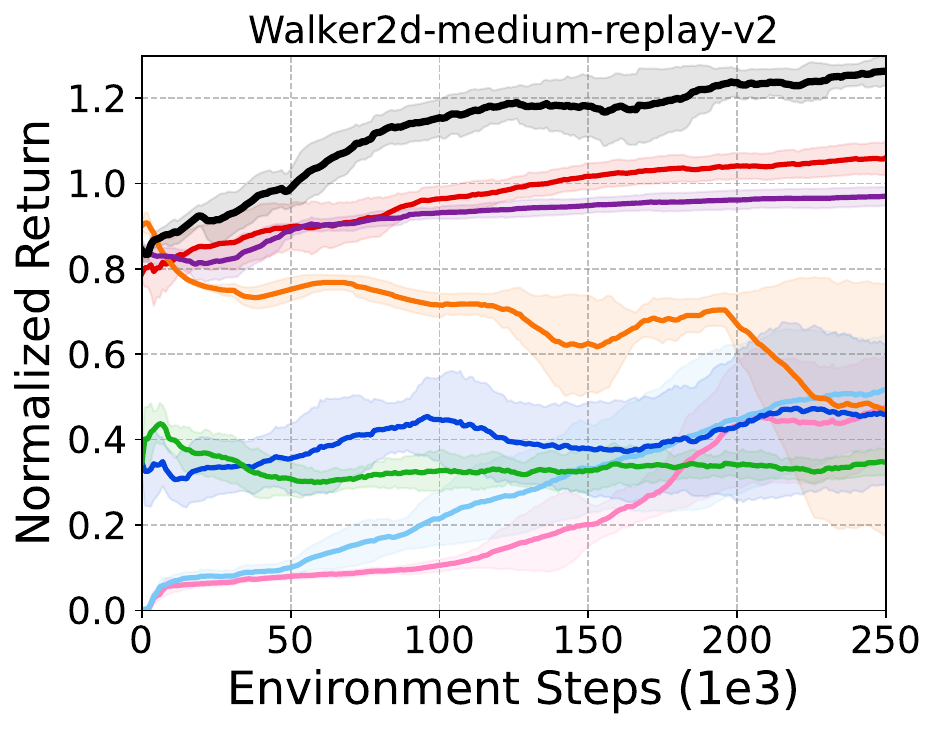}
    \includegraphics[width=0.3\textwidth]{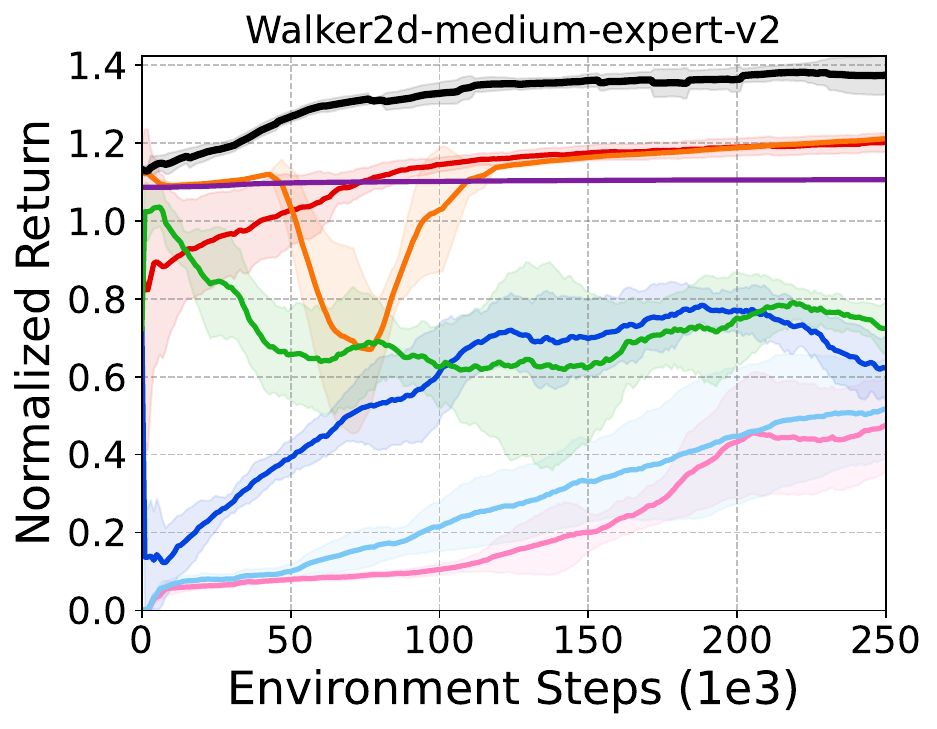}
    \includegraphics[width=0.3\textwidth]{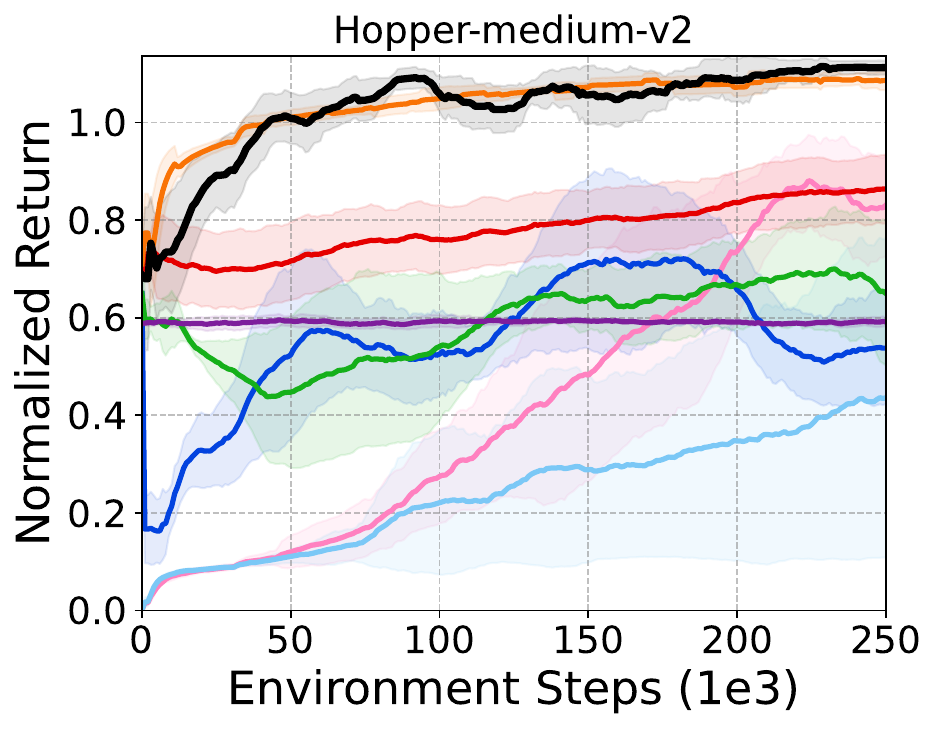}
    \includegraphics[width=0.3\textwidth]{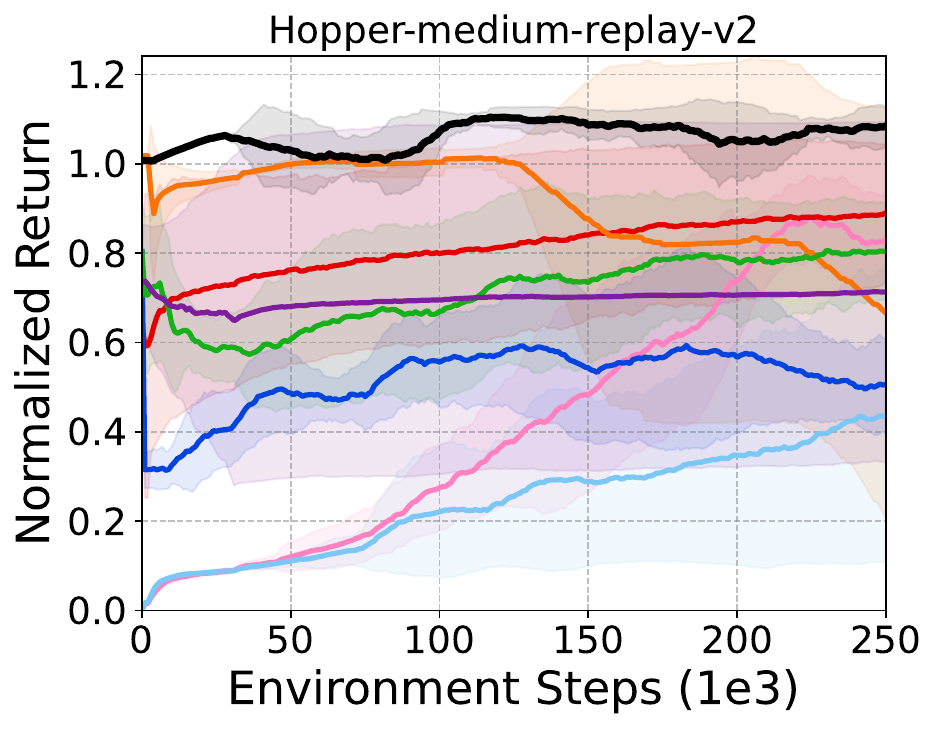}
    \includegraphics[width=0.3\textwidth]{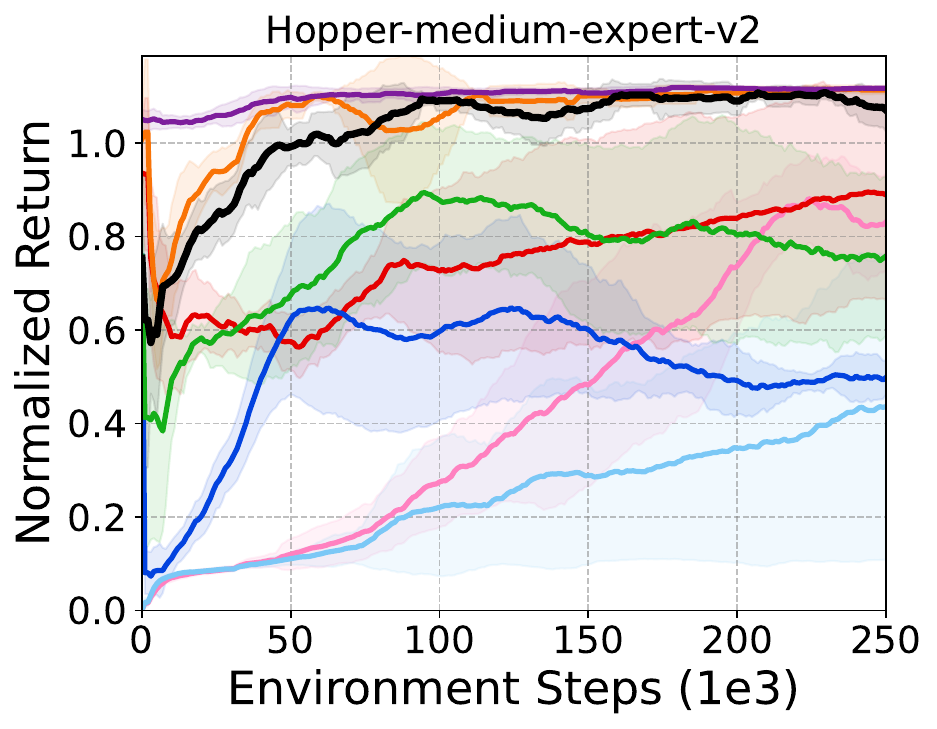}
    \caption{Online learning curves of different methods across five seeds on MuJoCo locomotion tasks. The solid lines and shaded regions represent mean and standard deviation, respectively.}
    \label{fig:Mujoco-ENOTO-CQL}
\end{figure*}

\section{Experiments}

In this section, we present the empirical evaluations of our ENOTO framework. We begin with locomotion tasks from D4RL~\cite{fu2020d4rl} to measure the training stability, learning efficiency and final performance of ENOTO by comparing it with several state-of-the-art offline-to-online RL methods. Additionally, we evaluate ENOTO on more challenging navigation tasks to verify its versatility.

\subsection{Locomotion Tasks}
\label{exp_locomotion}

We first evaluate our ENOTO framework on MuJoCo \cite{todorov2012mujoco} locomotion tasks, i.e., HalfCheetah, Walker2d, and Hopper from the D4RL benchmark suite \cite{fu2020d4rl}. To demonstrate the applicability of ENOTO on various suboptimal datasets, we use three dataset types: medium, medium-replay, and medium-expert. Specifically, medium datasets contain samples collected by a medium-level policy, medium-replay datasets include all samples encountered while training a medium-level agent from scratch, and medium-expert datasets consist of samples collected by both medium-level and expert-level policies. We pre-train the agent for 1M training steps in the offline phase and perform online fine-tuning for 250K environmental steps. Additional experimental details can be found in the appendix.

\begin{figure*}[t]
    \centering
    \includegraphics[width=0.3\textwidth]{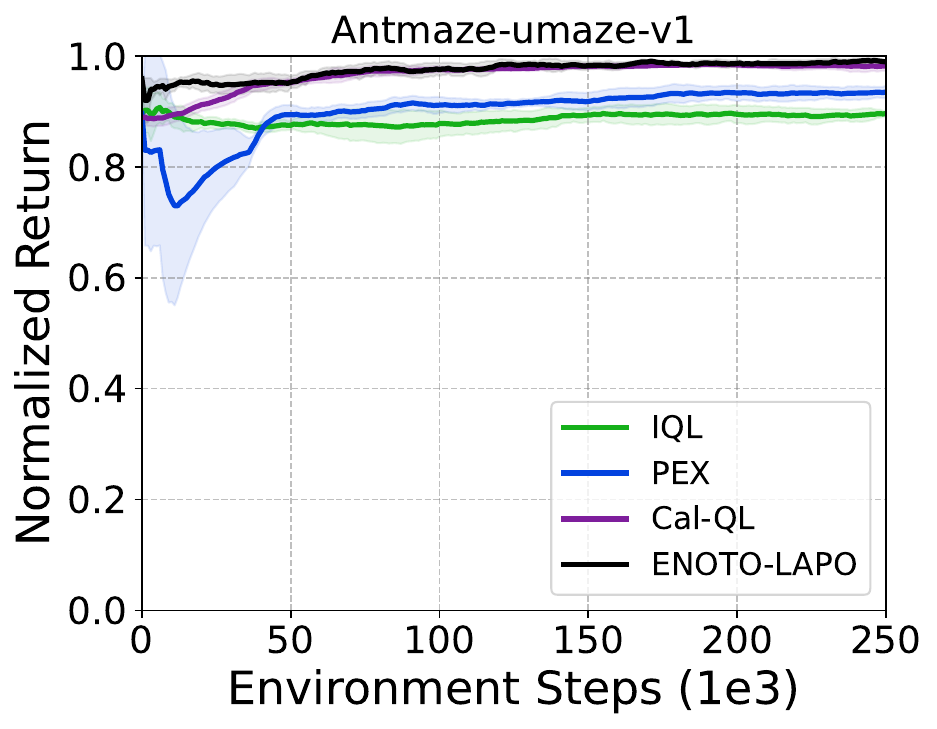}
    \includegraphics[width=0.3\textwidth]{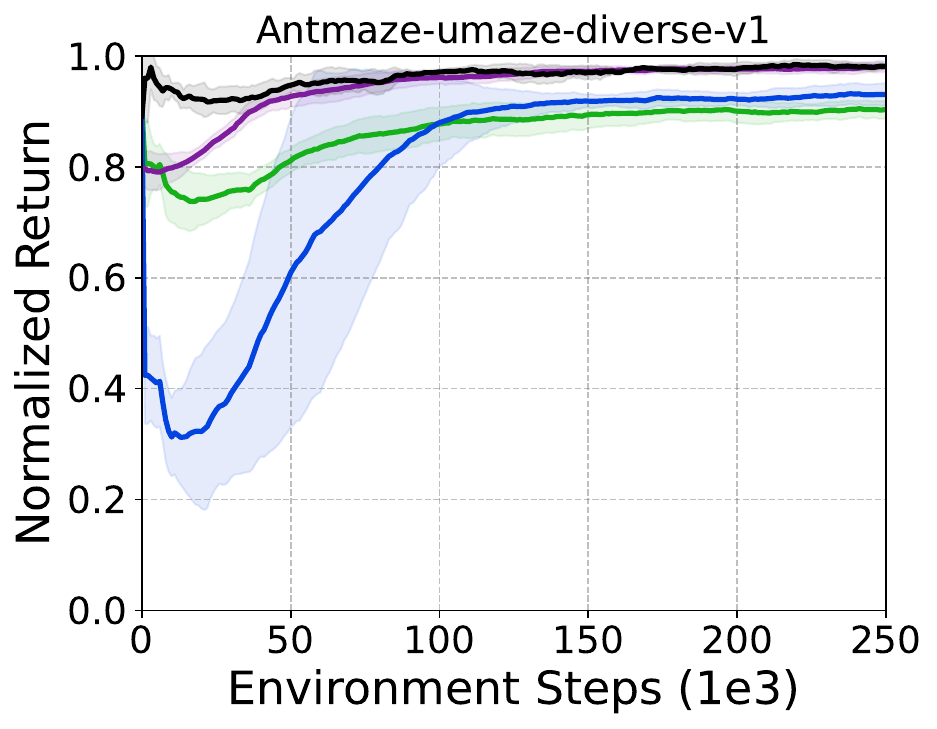}
    \includegraphics[width=0.3\textwidth]{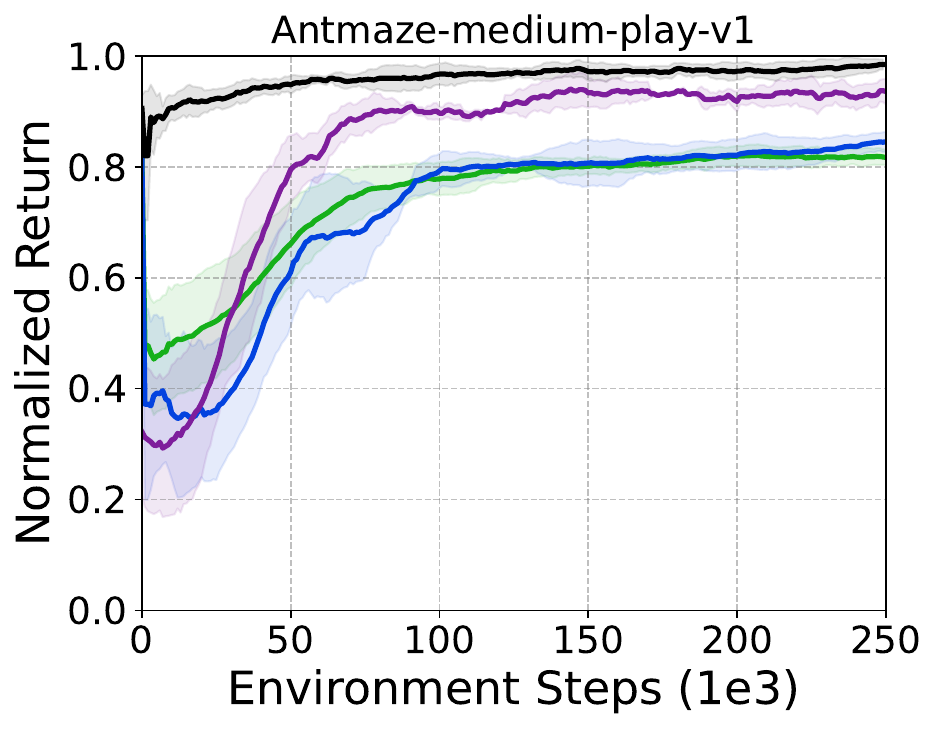}
    \includegraphics[width=0.3\textwidth]{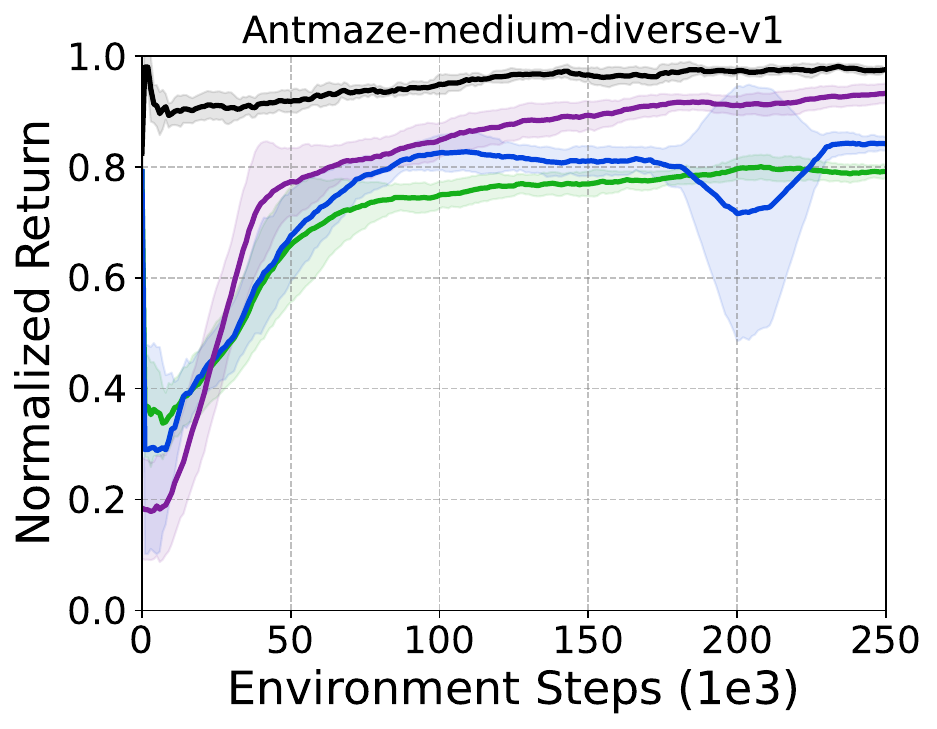}
    \includegraphics[width=0.3\textwidth]{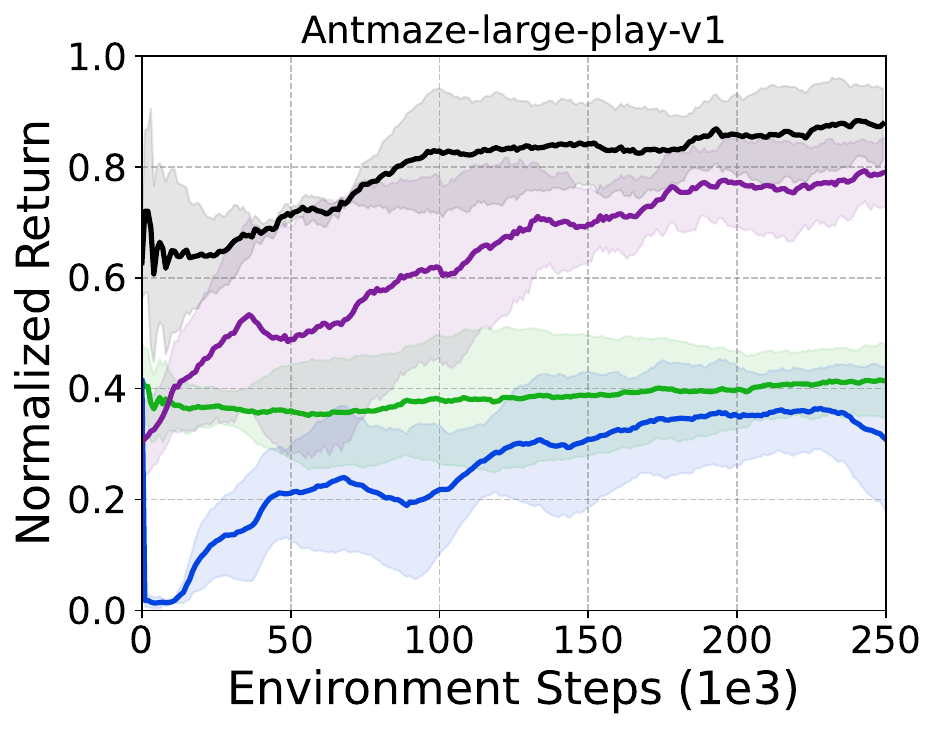}
    \includegraphics[width=0.3\textwidth]{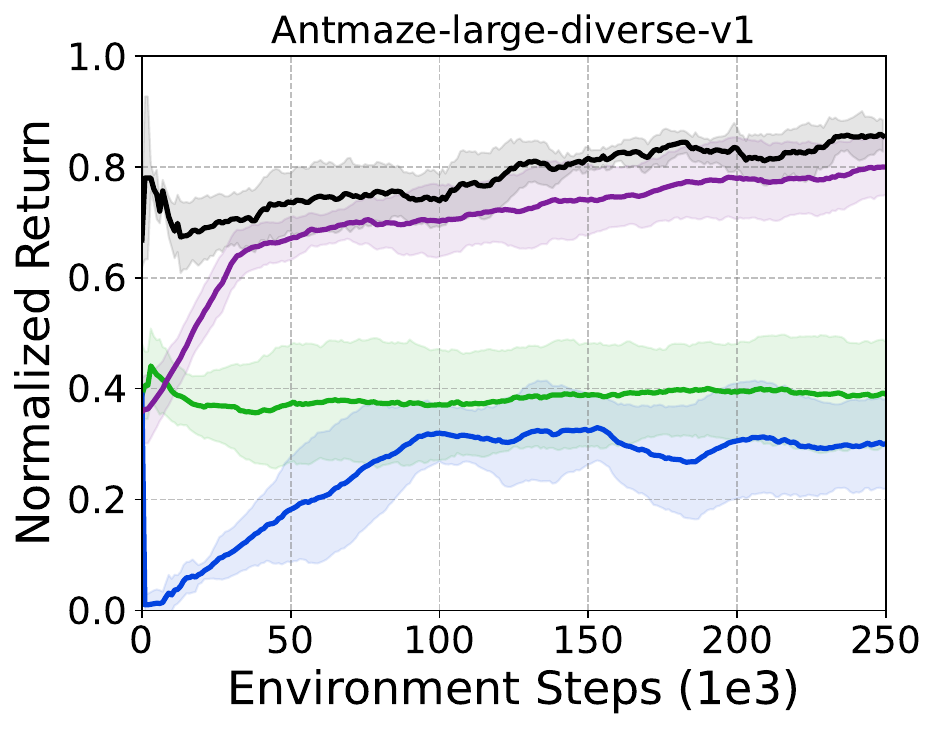}
    \caption{Online learning curves of different methods across five seeds on Antmaze navigation tasks. The solid lines and shaded regions represent mean and standard deviation, respectively.}
    \label{fig:LAPO-10_Antmaze}
\end{figure*}

\paragraph{Comparative Evaluation.} We consider the following methods as baselines.

\begin{itemize}
    \item \textbf{AWAC}~\cite{nair2020awac}: an offline-to-online RL method that forces the policy to imitate actions with high advantage estimates in the dataset.
    \item \textbf{BR}~\cite{lee2022offline}: an offline-to-online RL method that trains an additional network to prioritize samples in order to effectively use new data as well as near-on-policy samples in the offline dataset.
    \item \textbf{PEX}~\cite{zhang2023policy}: a recent offline-to-online RL method utilizing an offline policy within a policy set, expanding it with additional policies, and constructing a categorical distribution based on their values at the current state to select the final action.
    \item \textbf{Cal-QL}~\cite{nakamoto2023cal}: a recent offline-to-online RL method learning a conservative value function initialization that underestimates the value of the learned policy from offline data, while also being calibrated, in the sense that the learned Q-values are at a reasonable scale.
    \item \textbf{IQL}~\cite{kostrikov2021offline}: a representative RL algorithm demonstrating superior offline performance and enabling seamless online fine-tuning through direct parameter transfer.
    \item \textbf{SAC}~\cite{haarnoja2018soft}: a SAC agent trained from scratch. This baseline highlights the benefit of offline-to-online RL, as opposed to fully online RL, in terms of learning efficiency.
    \item \textbf{Scratch}: training SAC-N + WeightedMinPair + SUNRISE online from scratch without offline pre-training, as opposed to our ENOTO framework.
\end{itemize}

Fig.~\ref{fig:Mujoco-ENOTO-CQL} shows the performance of the ENOTO-CQL method (ENOTO instantiated on CQL) and baseline methods during the online fine-tuning phase. Compared with pure online RL methods such as SAC and Scratch, ENOTO-CQL starts with a well-performed policy and learns quickly and stably, proving the benefits of offline pre-training. For offline RL methods, IQL shows limited improvement as complete pessimistic training is no longer suitable for online fine-tuning, while ENOTO-CQL displays fast fine-tuning. Among other offline-to-online RL methods, the performance of AWAC is limited by the quality of the dataset due to the operation of training its policy to imitate actions with high advantage estimates, resulting in slow improvement during the online phase. While BR can attain performance second only to ENOTO-CQL on some datasets, it also suffers from unstable training. PEX exhibits a notable decline in performance during the initial stages of online fine-tuning across various datasets, attributed to the randomness of newly trained policies in the early phase, which negatively affects training stability. Although the original PEX paper does not explicitly address this phenomenon, a meticulous examination of its experimental section reveals that performance drop indeed affects PEX. We contend that the phenomenon of performance drop is a pivotal concern in the domain of offline-to-online RL, warranting significant attention. Turning to the Cal-QL algorithm, while its efficacy is prominently showcased in intricate tasks such as Antmaze, Adroit, and Kitchen, as emphasized in the paper, we note a more subdued performance in traditional MuJoCo tasks. The enhancements during the online phase appear relatively constrained. However, its most salient attribute lies in its exceptional stability, effectively circumventing the issue of performance drop. It is worth noting that the Hopper-medium-expert-v2 dataset represents a special case where most considered offline-to-online RL methods exhibit varying degrees of performance drop, except for Cal-QL, which maintains its offline-stage performance while remaining stable.

It is important to underscore that due to the partial incompleteness of code provided by certain baseline algorithms, our experiments partially rely on publicly available and widely accepted code repositories from GitHub~\cite{seno2022d3rlpy,tarasov2022corl}. Consequently, the experimental results may exhibit slight deviations from the reported outcomes in the original papers, which will be comprehensively detailed in the appendix. Nevertheless, through rigorous comparisons encompassing both the baseline papers' original performance metrics and the results obtained from our code implementation, our ENOTO method consistently surpasses the baseline approaches in terms of training stability, learning efficiency, and final performance across most tasks. Unfortunately, due to constraints within this text, we can only present the results attained from executing the code, as graphical representations from the source papers cannot be seamlessly incorporated.


\subsection{Navigation Tasks}

We further verify the effectiveness of ENOTO on D4RL navigation task Antmaze \cite{fu2020d4rl} by integrating another offline RL algorithm LAPO \cite{chen2022latent}. In detail, we specialize ENOTO as LAPO-N + WeightedMinPair + SUNRISE, i.e., ENOTO-LAPO. For the Antmaze task, we consider three types of mazes: umaze, medium and large mazes, and two data compositions: play and diverse. The data compositions vary in their action coverage of different regions of the state space and the sub-optimality of the behavior policy.

\paragraph{Comparative Evaluation.}

Since Antmaze is a more challenging task, most offline RL methods struggle to achieve satisfactory results in the offline phase, we only compare our ENOTO-LAPO method on this task with three effective baseline methods, IQL, PEX and Cal-QL. Specifically, for the D4RL Antmaze tasks, these methods apply a reward modification following previous works. This modification effectively introduces a survival penalty that encourages the agent to complete the maze as quickly as possible. In the online phase, we maintain the same reward modification as the offline phase during training but keep the rewards unchanged during evaluation.

Fig.~\ref{fig:LAPO-10_Antmaze} presents the performance of ENOTO-LAPO and baseline methods during the online fine-tuning phase. First, LAPO demonstrates better offline performance than IQL, providing a higher starting point for the online phase, especially in the umaze and medium maze environments where it almost reaches the performance ceiling. In the online stage, IQL shows slower asymptotic performance due to offline policy constraints. Building upon IQL, PEX enhances the degree of exploration by incorporating additional new policies trained from scratch, but the strong randomness of these policies in the early online stage causes performance drop. Note that although both IQL and PEX share the same starting point, PEX exhibits more severe performance drop on most tasks. Regarding the Cal-QL algorithm, akin to the outcomes portrayed in the original paper, it demonstrates robust performance in the Antmaze environment, outperforming significantly its MuJoCo counterparts. Notably, it exhibits superior stability and learning efficiency compared to the two baseline methods, IQL and PEX. For our proposed ENOTO framework, we demonstrate that ENOTO-LAPO can not only enhance the offline performance, but also facilitate stable and rapid performance improvement while maintaining the offline performance without degradation. This approach enables the offline agent to quickly adapt to the real-world environment, providing efficient and effective online fine-tuning. Additionally, we directly leverage LAPO with two Q networks for offline-to-online training and use the comparison with our ENOTO-LAPO method to further verify the effectiveness of our ENOTO framework. The results including some ablation studies can be found in the appendix.

\section{Conclusions and Limitations}

In this work, we have demonstrated that Q-ensembles can be efficiently leveraged to alleviate unstable training and performance drop, and serve as a more flexible constraint method for online fine-tuning in various settings. Based on this observation, we propose Ensemble-based Offline-to-Online (ENOTO) RL Framework, which enables many pessimistic offline RL algorithms to perform optimistic online fine-tuning and improve their performance efficiently while maintaining stable training process. The proposed framework is straightforward and can be combined with many existing offline RL algorithms. We instantiate ENOTO with different combinations and conducted experiments on a wide range of tasks to demonstrate its effectiveness.

Despite the promising results, there are some limitations to our work that should be acknowledged. First, although ENOTO is designed to be a flexible plugin for various offline RL algorithms, it may require further modifications to achieve optimal performance in different contexts. For instance, adjusting the weight coefficient of the BC item may result in better fine-tuning performance for TD3+BC~\cite{fujimoto2021minimalist}. Second, the computational cost of ensembles and uncertainty estimates may limit the scalability of ENOTO to large-scale problems. Future work could investigate ways to reduce the computational overhead by using deep ensembles~\cite{fort2019deep} or ensemble distillation~\cite{hinton2015distilling}, while maintaining the performance by using Bayesian compression~\cite{louizos2017bayesian} or variational approximations~\cite{kingma2013auto}. These methods could make ENOTO more scalable and practical for large-scale problems and real-world applications, enabling the development of more efficient and reliable offline-to-online RL systems.

\clearpage

\appendix

\section*{Acknowledgments}

This work is supported by the National Natural Science Foundation of China (Grant Nos. 92370132, 62106172), the Science and Technology on Information Systems Engineering Laboratory (Grant Nos. WDZC20235250409, 6142101220304), and the Xiaomi Young Talents Program of Xiaomi Foundation.

\bibliographystyle{named}
\bibliography{ijcai24}

\clearpage

\appendix

\section{Related Works}

\paragraph{Offline RL}

Offline RL algorithms focus on training RL agents with pre-collected datasets. However, these algorithms face the challenge of distribution shift between the behavior policy and the policy being learned, which can cause issues due to out-of-distribution (OOD) actions sampled from the learned policy and passed into the learned critic. To mitigate this problem, prior methods constrain the learned policy to stay close to the behavior policy via explicit policy regularization~\cite{fujimoto2019off,wu2019behavior}, via implicit policy constraints~\cite{kostrikov2021offline,chen2022latent}, by leveraging auxiliary behavioral cloning losses~\cite{fujimoto2021minimalist}, by penalizing the Q-value of OOD actions to prevent selecting them~\cite{kumar2020conservative,an2021uncertainty,bai2022pessimistic,yang2022rorl}, or through model-based training with conservative penalties~\cite{yu2020mopo,kidambi2020morel}. Among the above methods, we choose several representative algorithms such as CQL~\cite{kumar2020conservative} and LAPO~\cite{chen2022latent} to be the base methods in the offline component of our framework due to their wide applicability and superior performance.

\paragraph{Offline-to-Online RL}

Offline-to-online RL refers to the process of improving the well-trained offline policy by incorporating online interactions. Directly applying pre-trained offline policy to the online fine-tuning stage may lead to poor performance due to excess conservatism~\cite{nair2020awac,lee2022offline,zhao2022adaptive}. To adapt offline RL algorithms to the online environment, several modifications are required. AWAC~\cite{nair2020awac} is the first algorithm proposed to perform well in the offline-to-online RL setting, which forces the policy to imitate actions with high advantage estimates. AW-Opt~\cite{lu2021aw} improves upon AWAC by incorporating positive sample filtering and hybrid actor-critic exploration during the online stage. BR~\cite{lee2022offline} trains an additional neural network to prioritize samples in order to effectively use new data as well as near-on-policy samples from the offline dataset. PEX~\cite{zhang2023policy} proposes a policy expansion approach for offline-to-online RL, which trains more policies from scratch in the online phase and combine them with the pre-trained offline policy to make decisions jointly. Cal-QL~\cite{nakamoto2023cal} is a recent offline-to-online RL method learning a conservative value function initialization that underestimates the value of the learned policy from offline data, while also being calibrated, in the sense that the learned Q-values are at a reasonable scale. A concurrent work with us is O3F~\cite{mark2022fine}, which also aims to eliminate pessimism in online learning. However, our research takes a more holistic perspective by connecting the pessimistic offline learning and optimistic online phases through ensemble modeling, proposing a more applicable framework that encapsulates the implementation of O3F as a special case.

\paragraph{Q-Ensembles in RL}

Q-Ensemble methods have been widely utilized to enhance the performance of RL~\cite{osband2016deep,fujimoto2018addressing,ciosek2019better,chen2021randomized,lee2021sunrise}. TD3~\cite{fujimoto2018addressing} leverages an ensemble of two value functions and uses their minimum for computing the target value during Bellman error minimization. REDQ~\cite{chen2021randomized} minimizes over a random subset of Q-functions in the target to reduce over-estimation bias. For offline RL, a number of works have extended this to propose backing up minimums or lower confidence bound estimates over larger ensembles~\cite{fujimoto2019off,wu2019behavior,agarwal2020optimistic,an2021uncertainty,bai2022pessimistic,yang2022rorl}. In particular, EDAC~\cite{an2021uncertainty} achieves impressive performance by simply increasing the number of Q-networks along with the clipped Q-learning and further proposes to reduce the required number of ensemble networks through ensemble gradient diversification for the purpose of reducing computational cost. PBRL~\cite{bai2022pessimistic} and RORL~\cite{yang2022rorl} both employ an ensemble of bootstrapped Q-functions for uncertainty quantification and perform pessimistic updates to penalize Q functions with high uncertainties. Recent work~\cite{ghasemipour2022so} advocates for using independently learned ensembles without sharing target values and optimizing a policy based on the lower confidence bound of predicted action values. In our work, we discover that Q-ensembles are highly effective in addressing the performance degradation that occurs in the offline-to-online setting, and we advocate for using Q-ensembles for both offline and online algorithms to achieve steady and sample-efficient offline-to-online RL.

\paragraph{Exploration Mechanisms}

Various exploration approaches have been proposed to accelerate the efficiency of online training in recent years. These methods can typically be divided into two main categories following~\cite{exploraton2023jianye}: uncertainty-oriented exploration and intrinsic motivation-oriented exploration. The former employs heuristic design to formulate various intrinsic motivations for exploration, based on factors such as visitation count~\cite{ostrovski2017count,tang2017exploration}, curiosity~\cite{savinov2018episodic}, information gain~\cite{houthooft2016vime}, etc. Uncertainty-oriented exploration methods adopt the principle of optimism in the face of uncertainty and use Q-ensembles to encourage agents to explore areas with higher uncertainty~\cite{ecoffet2021first,lee2021sunrise,ovde}. In this paper, we investigate uncertainty-based exploration mechanisms under the unified framework of Q-ensembles to enhance the performance during the online fine-tuning phase.

\section{Environment Settings}
\label{app:env_setting}

\paragraph{MuJoCo Gym}

We investigate three MuJoCo locomotion tasks, namely HalfCheetah, Walker2d, and Hopper \cite{todorov2012mujoco}. The goal of each task is to move forward as fast as possible, while keeping the control cost minimal. For each task, we consider four types of datasets. The random datasets consist of policy rollouts generated by random policies. The medium datasets contain rollouts from medium-level policies. The medium-replay datasets encompass all samples collected during the training of a medium-level agent from scratch. In the case of the medium-expert datasets, half of the data comprises rollouts from medium-level policies, while the other half consists of rollouts from expert-level policies. In this study, we exclude the random and the expert datasets, as in typical real-world scenarios, we rarely use a random policy or have an expert policy for system control. We utilize the v2 version of each dataset.

\paragraph{Antmaze}

We investigate the Antmaze navigation tasks that involve controlling an 8-DoF ant quadruped robot to navigate through mazes and reach a desired goal. The agent receives sparse rewards of +1/0 based on whether it successfully reaches the goal or not. We study each method using the following datasets from D4RL \cite{fu2020d4rl}: large-diverse, large-play, medium-diverse, medium-play, umaze-diverse, and umaze. The difference between diverse and play datasets is the optimality of the trajectories they contain. The diverse datasets consist of trajectories directed towards random goals from random starting points, whereas the play datasets comprise trajectories directed towards specific locations that may not necessarily correspond to the goal. We use the v1 version of each dataset.

\section{Experiment Details}
\label{app:exp_detail}

\paragraph{Baselines}

For CQL, SAC and AWAC, we use the implementation provided by~\cite{seno2022d3rlpy}: \url{https://github.com/takuseno/d3rlpy} with default hyperparameters. For CQL-N and SAC-N, we keep the default setting from the CQL and SAC experiments other than the ensemble size N. For Balanced Replay (BR), as the official implementation provided by the author of~\cite{lee2022offline} does not contain the offline pre-training part, we implement BR based on d3rlpy. For LAPO-N, we extend the official implementation provided by the author of~\cite{chen2022latent}: \url{https://github.com/pcchenxi/LAPO-offlineRL} to easily adjust the size of ensemble. For PEX and IQL, we use the original implementation provided by the author of~\cite{zhang2023policy}: \url{https://github.com/Haichao-Zhang/PEX}. For Cal-QL, we use the implementation provided by~\cite{tarasov2022corl}: \url{https://github.com/tinkoff-ai/CORL} with default hyperparameters. While we do not utilize the code provided by the original paper for certain methods, some of the our results obtained using the employed code demonstrate superior performance compared to those reported in the original paper. Moreover, when compared to the results provided in the original paper, our proposed ENOTO framework consistently outperforms them. We list the hyperparameters for these methods in Table~\ref{Table:hyper}.

\begin{table*}[h]
\caption{Hyperparameters used in the D4RL MuJoCo experiments}
\label{sample-table}
\vskip 0.15in
\begin{center}
\begin{small}
\begin{tabular}{llllllllll}
    \toprule
    \textbf{Hyperparameters} & \textbf{CQL-N} & \textbf{SAC-N} & \textbf{LAPO-N} & \textbf{AWAC} & \textbf{BR} & \textbf{Cal-QL} & \textbf{PEX} & \textbf{IQL} \\
    \midrule
    policy learning rate & 3e-5 & 3e-5 & 2e-4 & 3e-4 & 3e-5 & 1e-4 & 3e-4 & 3e-4 \\
    critic learning rate & 3e-4 & 3e-4 & 2e-4 & 3e-4 & 3e-4 & 3e-4 & 3e-4 & 3e-4 \\
    alpha learning rate & 1e-4 & 1e-4 & - & - & 3e-4 & 5e-3 & - & - \\
    VAE learning rate & - & - & 2e-4 & - & - & - & - & - \\
    value learning rate & - & - & - & - & - & - & 3e-4 & 3e-4 \\
    ensemble size & 10 & 10 & 10 & 2 & 5 & 2 & 2 & 2 \\
    batch size & 256 & 256 & 512 & 1024 & 256 & 256 & 256 & 256 \\
    \bottomrule
    \label{Table:hyper}
\end{tabular}
\end{small}
\end{center}
\vskip -0.1in
\end{table*}

\paragraph{Offline Pre-training}

For all experiments, we conduct each algorithm for 1M training steps with 5 different seeds, following the common practice in offline RL works. Specifically, for the D4RL Antmaze tasks, IQL and PEX apply a reward modification by subtracting 1 from all rewards, as described in \url{https://github.com/tinkoff-ai/CORL/issues/14}. This modification effectively introduces a survival penalty that encourages the agent to complete the maze as quickly as possible. Additionally, LAPO multiplies all rewards by 100, which also enhances the distinction between the rewards for completing tasks and the rewards for unfinished tasks. These reward transformation techniques prove to be crucial for achieving desirable performance on the Antmaze tasks.

\paragraph{Online Fine-tuning}

For all experiments, we report the online fine-tuning performance over 250K timesteps with 5 seeds. Specifically, our framework loads all pre-trained networks, including the policy network, ensemble Q network and ensemble target Q network, while appending the necessary temperature hyperparameter for SAC to facilitate further fine-tuning. In the Antmaze environment, we maintain the same reward modification as the offline phase during training but keep the rewards unchanged during evaluation. To ensure a fair comparison with IQL and PEX, which utilize offline data, we also load LAPO and ENOTO-LAPO with offline data for online fine-tuning.

\section{Additional Results}
\label{app:add_results}

In this section, we provide more experiments and detailed results to help understand our proposed ENOTO framework more comprehensively.

\subsection{Ablation on Offline Data}
\label{app:ablation_offline_data}

\begin{figure*}[h]
    \centering
    \includegraphics[width=0.3\textwidth]{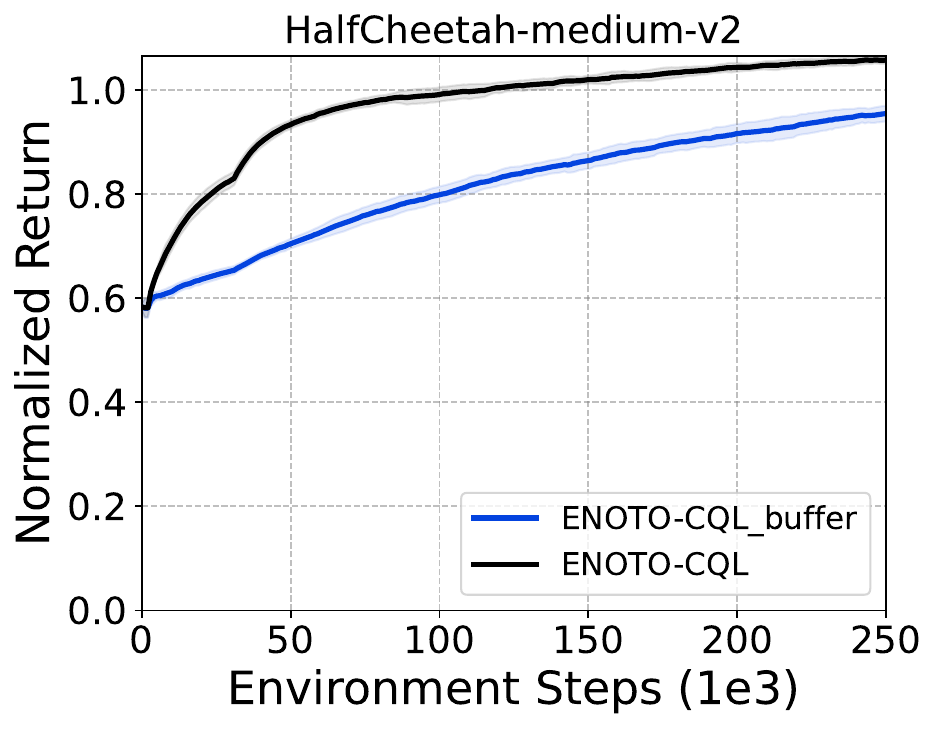}
    \includegraphics[width=0.3\textwidth]{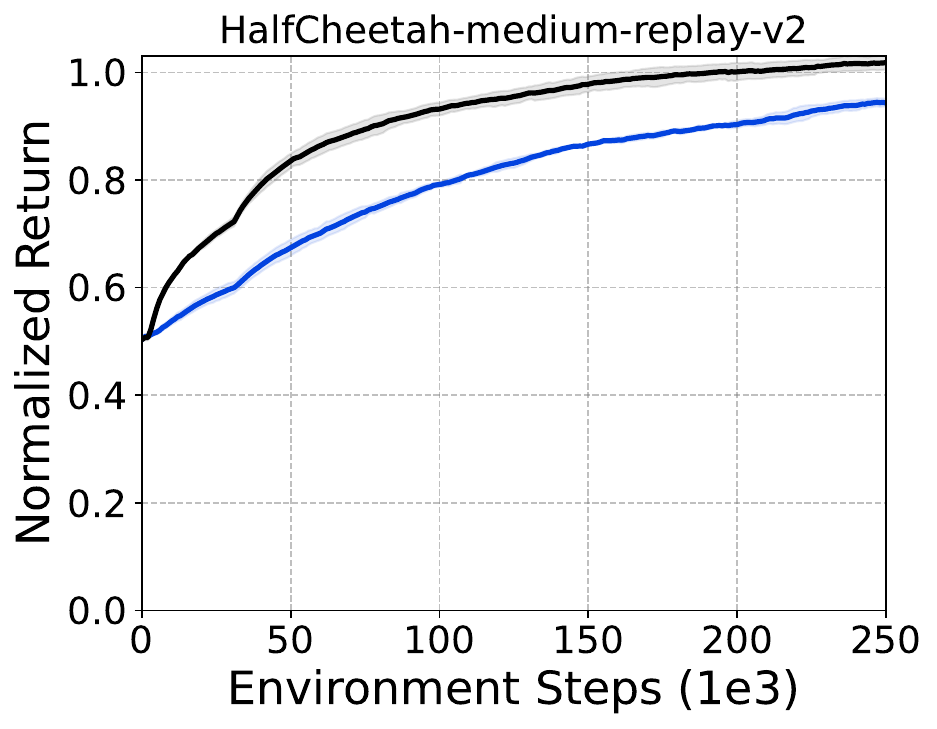}
    \includegraphics[width=0.3\textwidth]{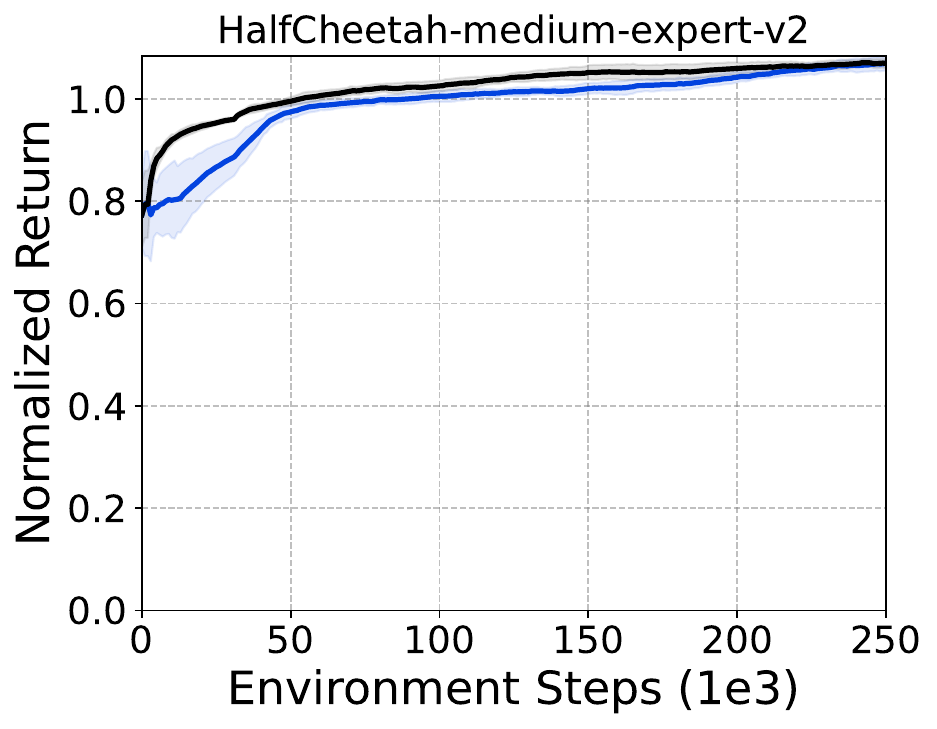}
    \includegraphics[width=0.3\textwidth]{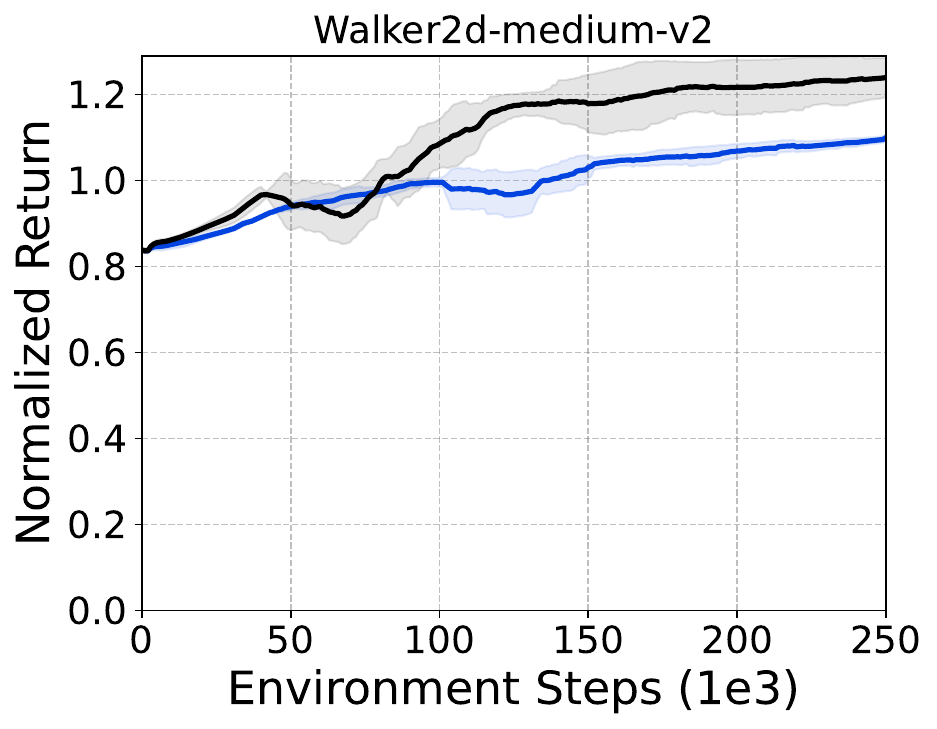}
    \includegraphics[width=0.3\textwidth]{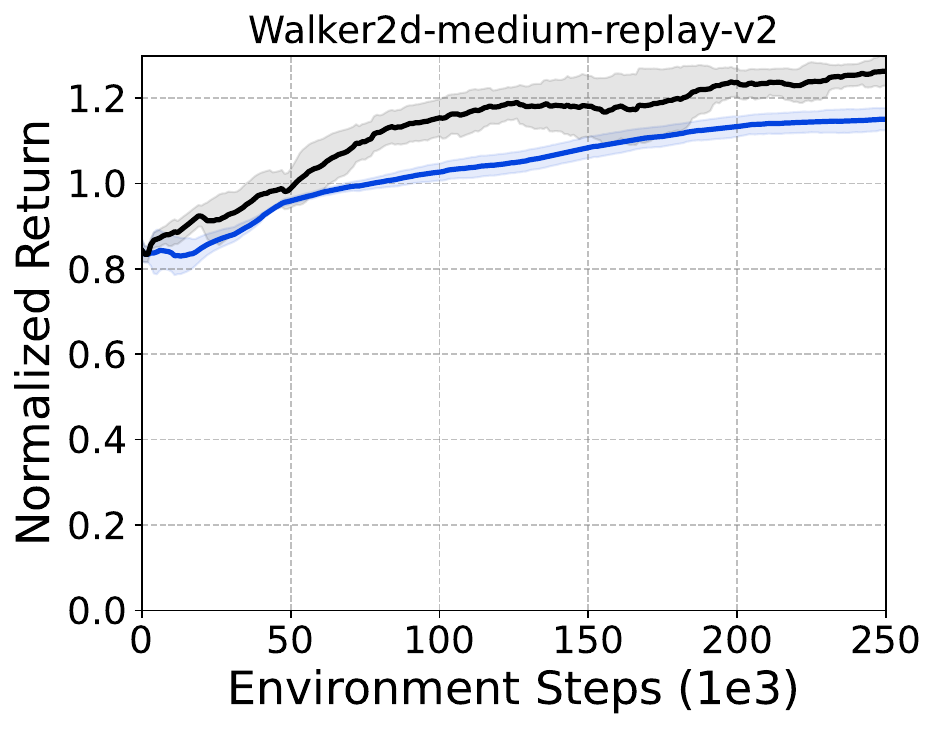}
    \includegraphics[width=0.3\textwidth]{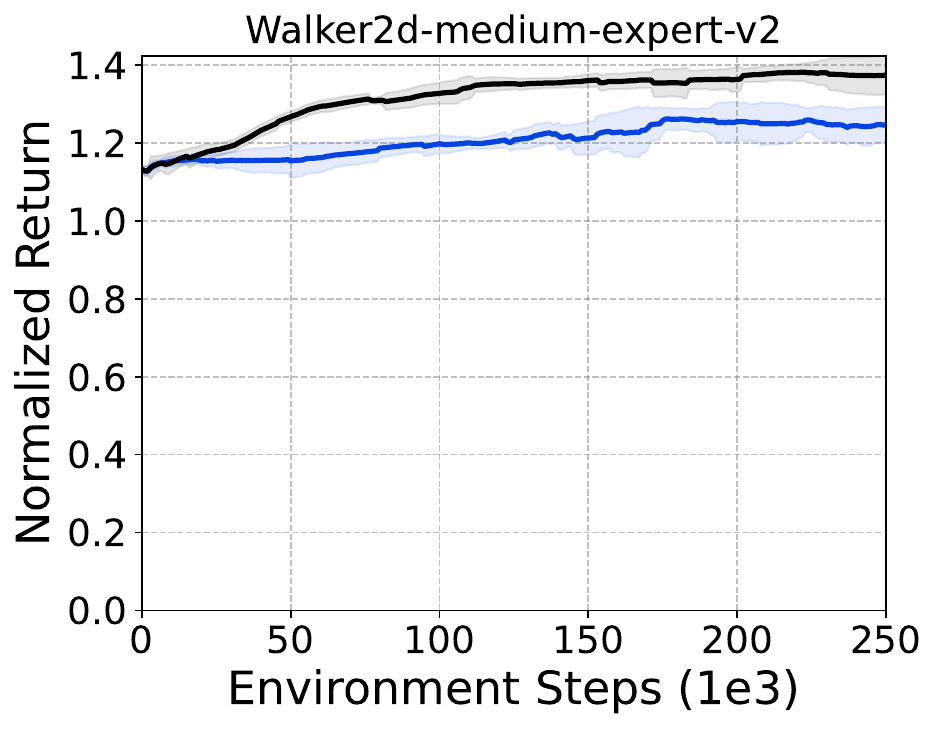}
    \includegraphics[width=0.3\textwidth]{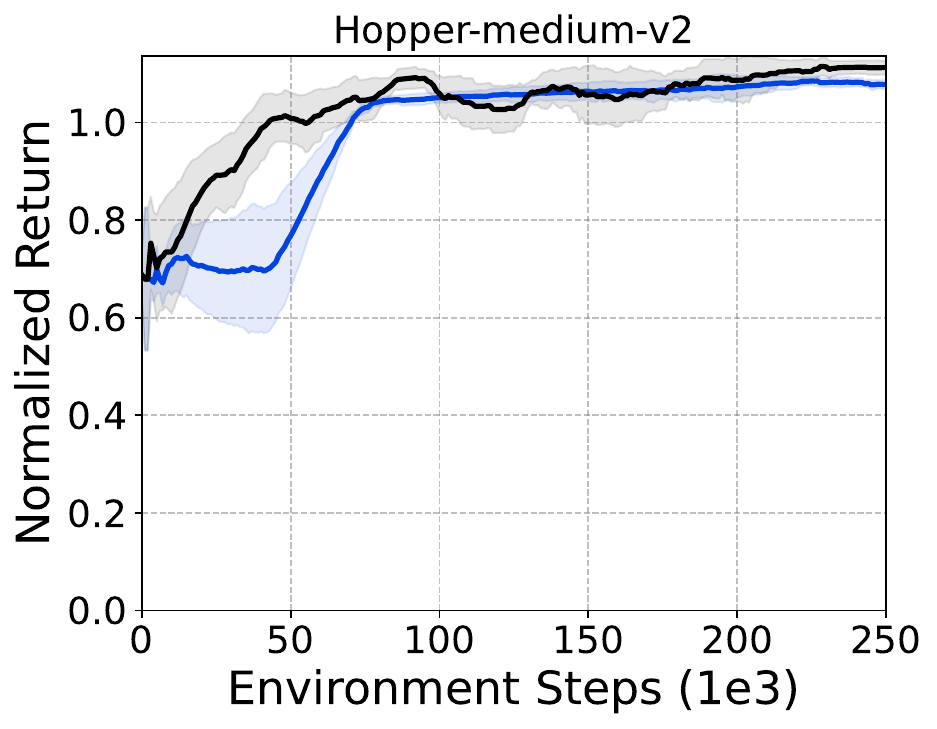}
    \includegraphics[width=0.3\textwidth]{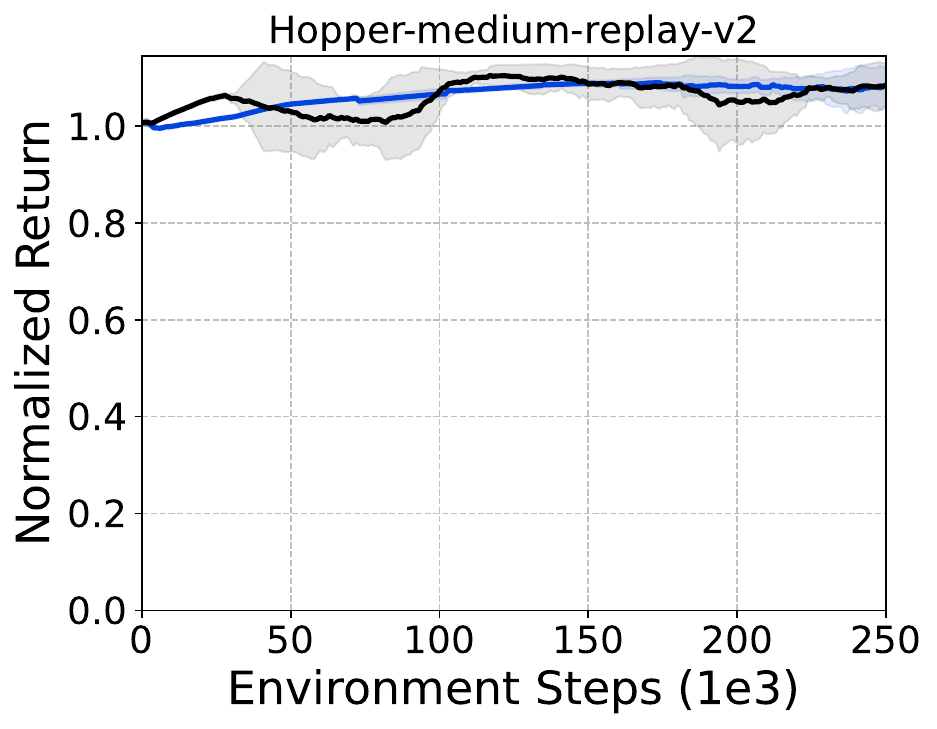}
    \includegraphics[width=0.3\textwidth]{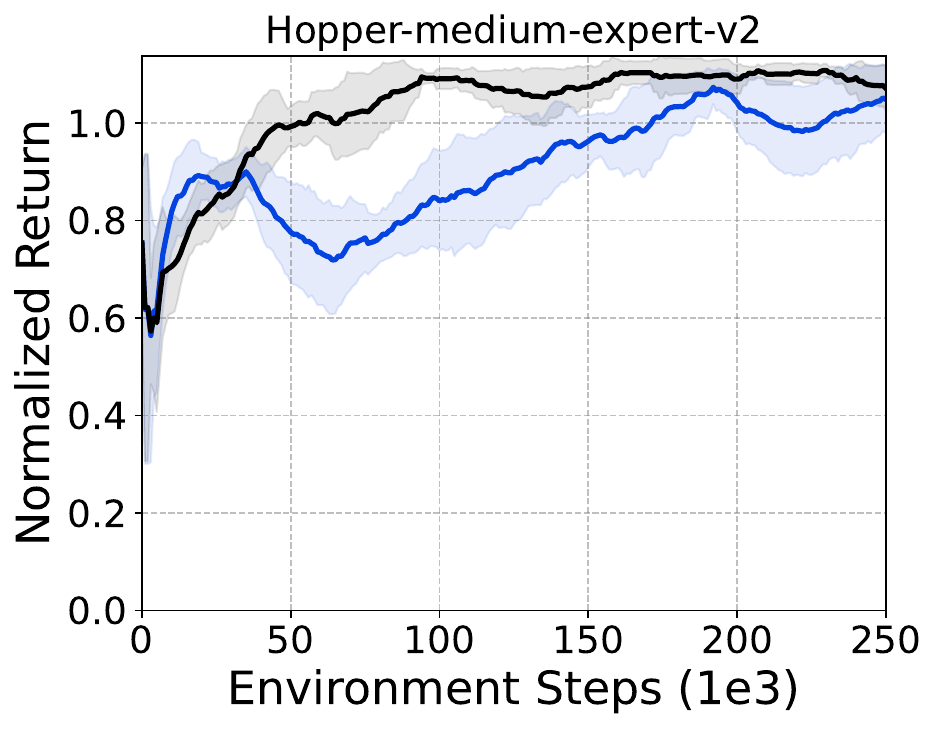}
    \caption{Ablation on offline data. The solid lines and shaded regions represent mean and standard deviation, respectively, across five runs.}
    \label{fig:Appendix_exp_ablation_data}
\end{figure*}

We conduct an ablation study to investigate the impact of using offline data during the online fine-tuning phase for all MuJoCo locomotion tasks, as shown in Fig.~\ref{fig:Appendix_exp_ablation_data}. Our results show that ENOTO-CQL\_buffer, which initializes the online buffer with offline data, exhibits slow performance improvement, while discarding the offline data allows it to achieve higher sample efficiency. This suggests that although many offline-to-online RL methods utilize offline data to alleviate performance degradation, it can adversely affect their online sample efficiency. In contrast, ENOTO-CQL successfully avoids significant performance drop even without using offline data, thereby enhancing learning efficiency during the online stage.

\subsection{Comparison of LAPO and ENOTO-LAPO}
\label{app:exp_lapo}

\begin{figure*}[h]
    \centering
    \includegraphics[width=0.3\textwidth]{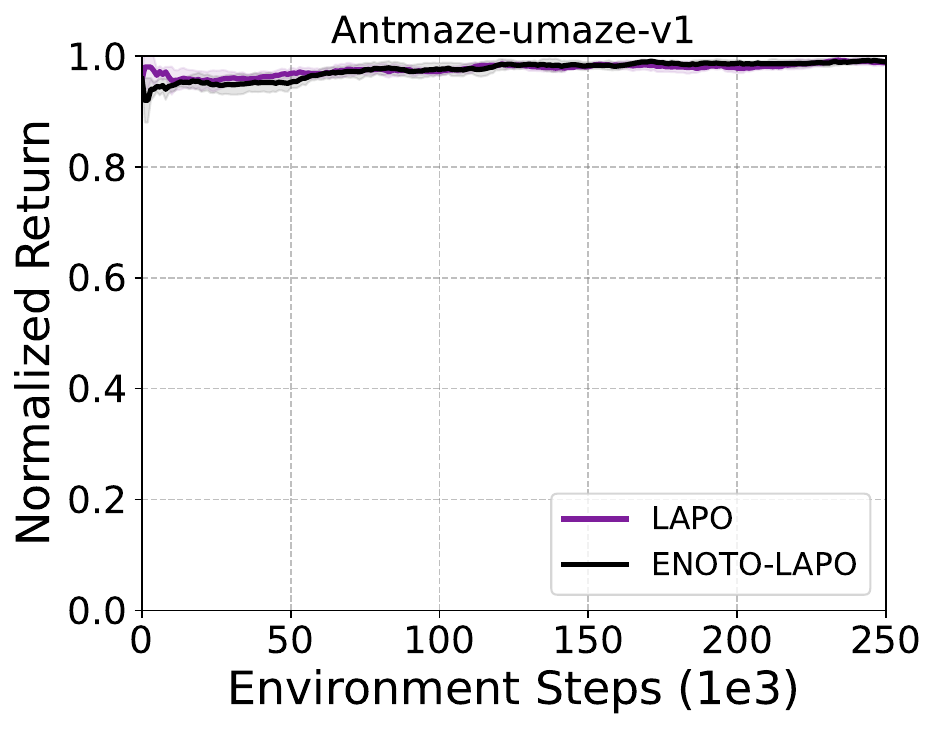}
    \includegraphics[width=0.3\textwidth]{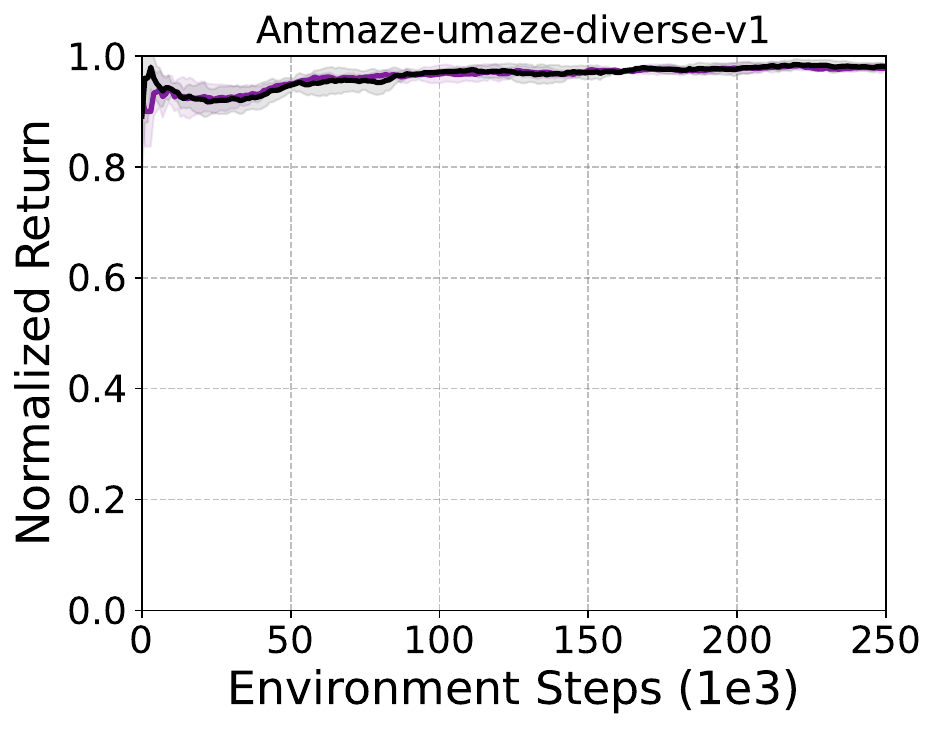}
    \includegraphics[width=0.3\textwidth]{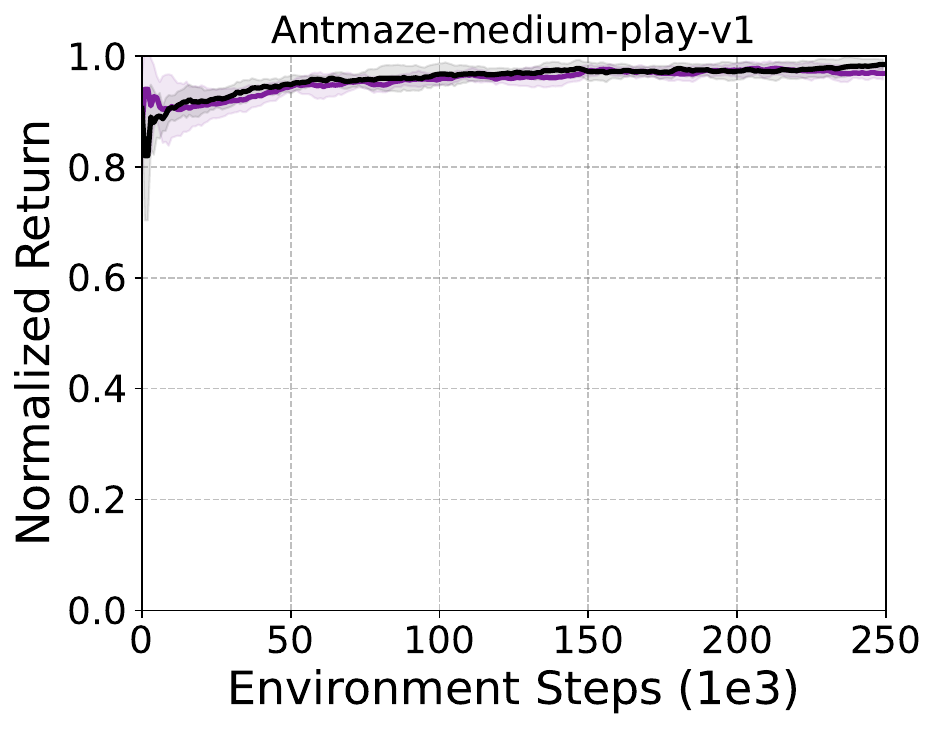}
    \includegraphics[width=0.3\textwidth]{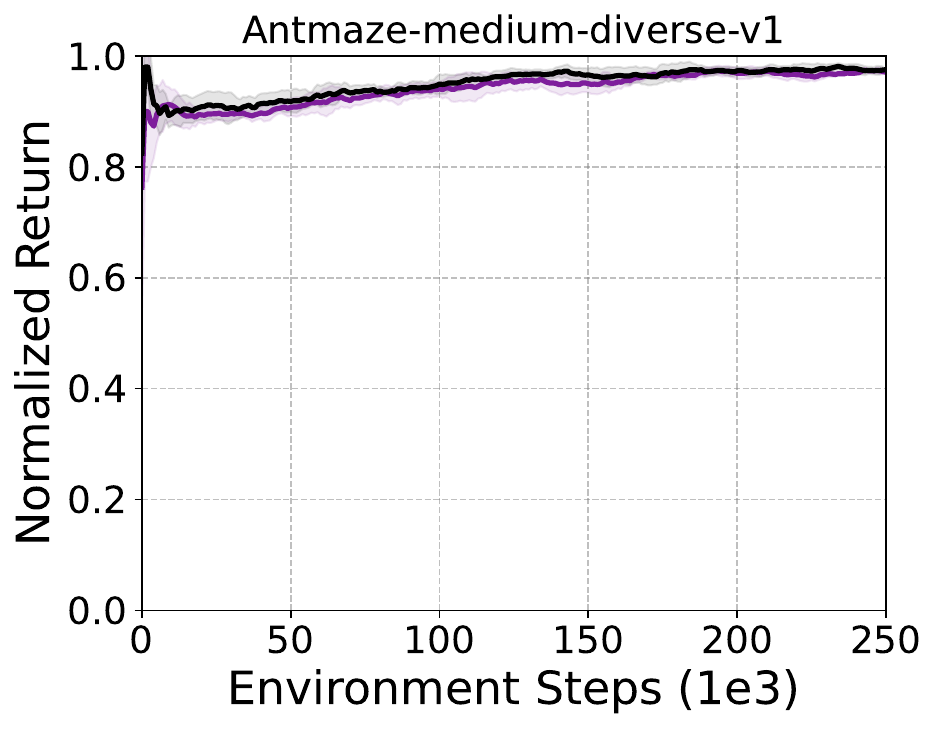}
    \includegraphics[width=0.3\textwidth]{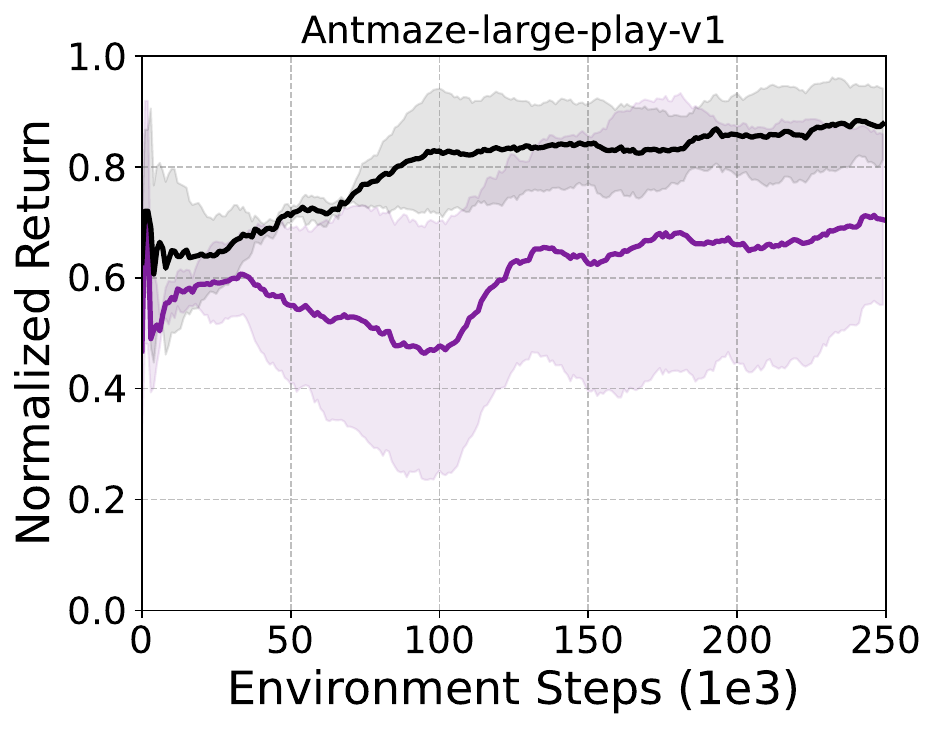}
    \includegraphics[width=0.3\textwidth]{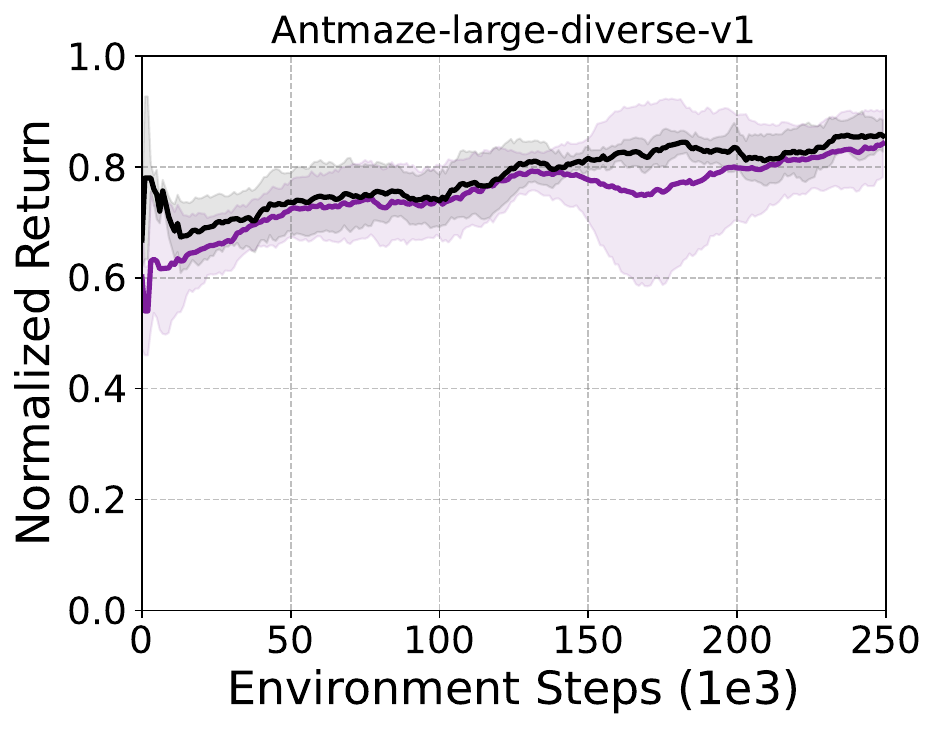}
    \caption{Online learning curves of LAPO and ENOTO-LAPO across five seeds on Antmaze tasks. The solid lines and shaded regions represent mean and standard deviation, respectively.}
    \label{fig:Appendix_exp_antmaze_lapo}
\end{figure*}

On the Antmaze tasks, we have conducted a comparative analysis between our ENOTO-LAPO method and several offline-to-online RL methods. To further validate the effectiveness of our ENOTO framework, we directly utilize LAPO with two Q networks for offline-to-online training and compare it with our ENOTO-LAPO method. The results are shown in Fig.~\ref{fig:Appendix_exp_antmaze_lapo}. As original LAPO can achieve near-optimal performance in simple environments such as umaze and medium mazes during the offline stage, the online fine-tuning performance of both LAPO and ENOTO-LAPO is comparable. However, in the more challenging large maze environment, directly using the offline pre-trained LAPO agent for online fine-tuning leads to slow performance improvement. By employing our proposed ENOTO framework, we demonstrate that ENOTO-LAPO can not only enhance the offline performance of LAPO, but also facilitate more rapid performance improvement while maintaining the offline performance without degradation. This approach enables the offline agent to quickly adapt to the real-world environment, providing efficient and effective online fine-tuning.

\subsection{Visualization and Analysis}

\begin{figure*}[t]
    \centering
    \includegraphics[width=0.35\textwidth]{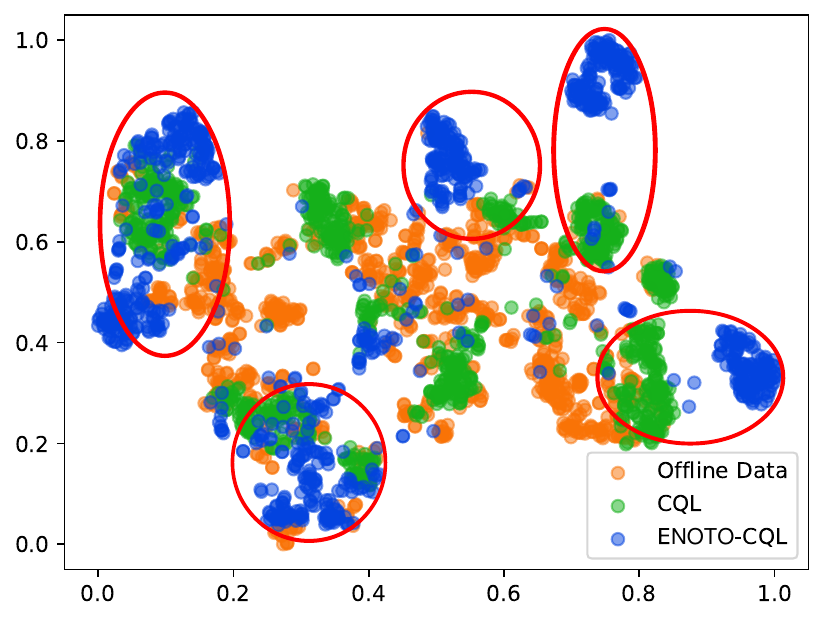}
    \caption{Visualization of the distribution of states generated by \seabornBlue{ENOTO-CQL}, \seabornGreen{CQL} in online phase and states in \seabornOrange{offline dataset}.}
    \label{Fig:tsne}
\end{figure*}

To better understand the training efficiency of ENOTO in comparison to traditional pessimistic offline RL algorithms, we compare the distribution of states generated by CQL, ENOTO-CQL in the online phase, and the distribution of states from the offline dataset. To visualize the results clearly, we plot the distribution with t-Distributed Stochastic Neighbor Embedding (t-SNE)~\cite{hinton2002stochastic}. The results are shown in Fig.~\ref{Fig:tsne}, it can be found that both the distribution of ENOTO-CQL states and CQL states bear some similarity to the distribution of offline states. However, the online states accessed by CQL are located on the edge of the offline area, but most still overlap with the offline states. On the other hand, the online states accessed by ENOTO-CQL deviate further from the offline states. With our ensemble-based design for optimistic exploration, ENOTO empowers the offline agent to explore more states beyond those contained in the offline dataset. This capability allows for swift adaptation to online environments and facilitates rapid performance improvement.

\subsection{Detailed Results of ENOTO Components}
\label{app:detail}

In this section, we provide all learning curves of ENOTO components that are restricted by the length of the text in the main paper.

\paragraph{Q-Ensembles}

\begin{figure*}[h]
    \centering
    \includegraphics[width=0.3\textwidth]{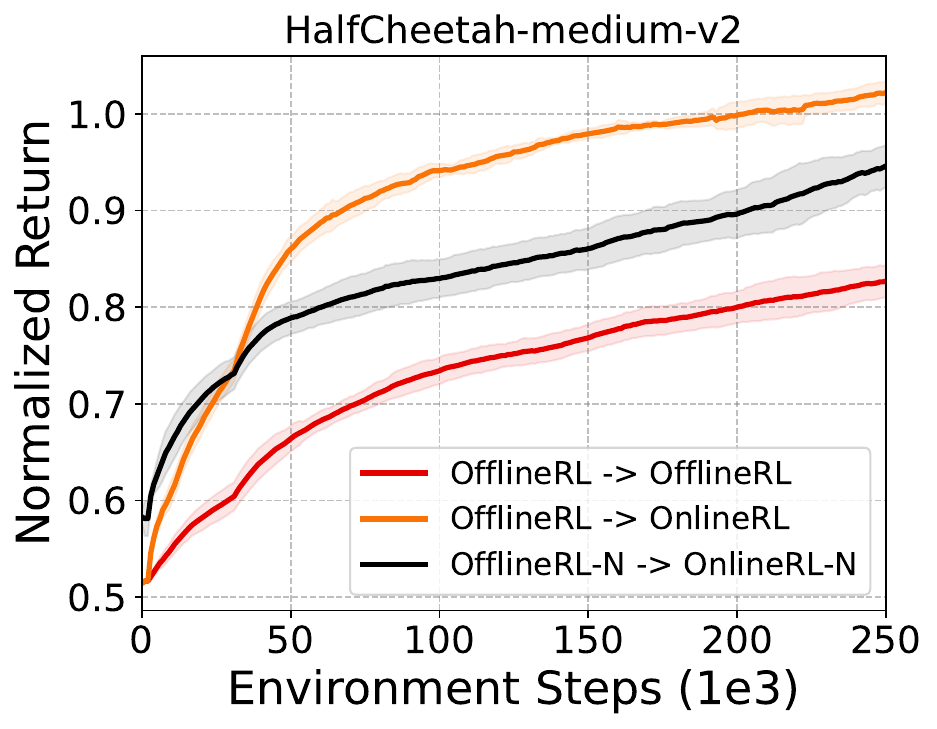}
    \includegraphics[width=0.3\textwidth]{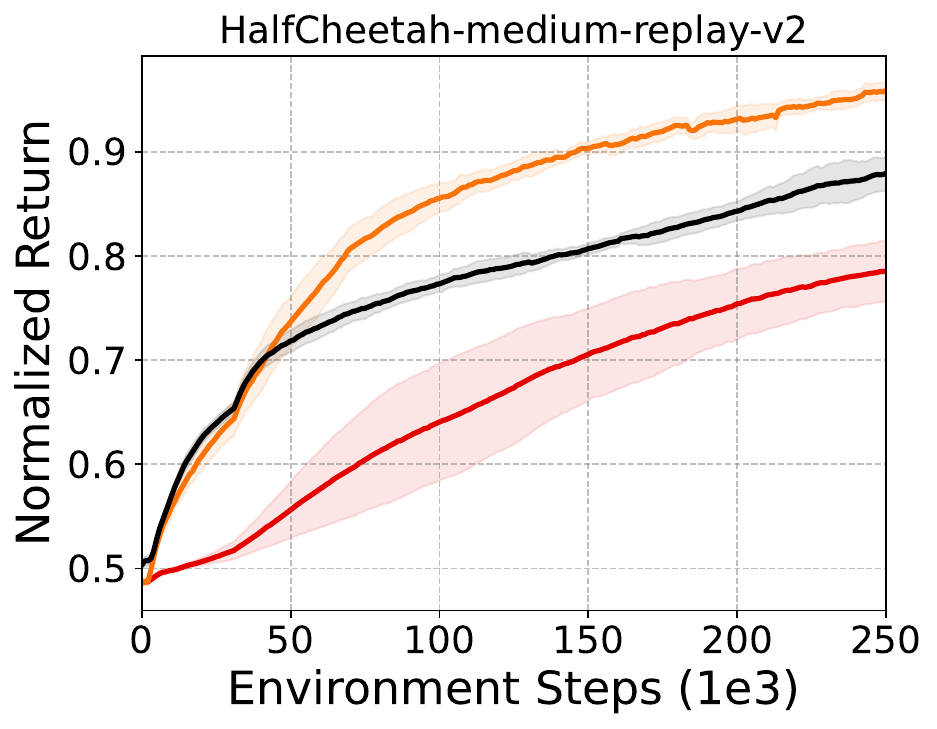}
    \includegraphics[width=0.3\textwidth]{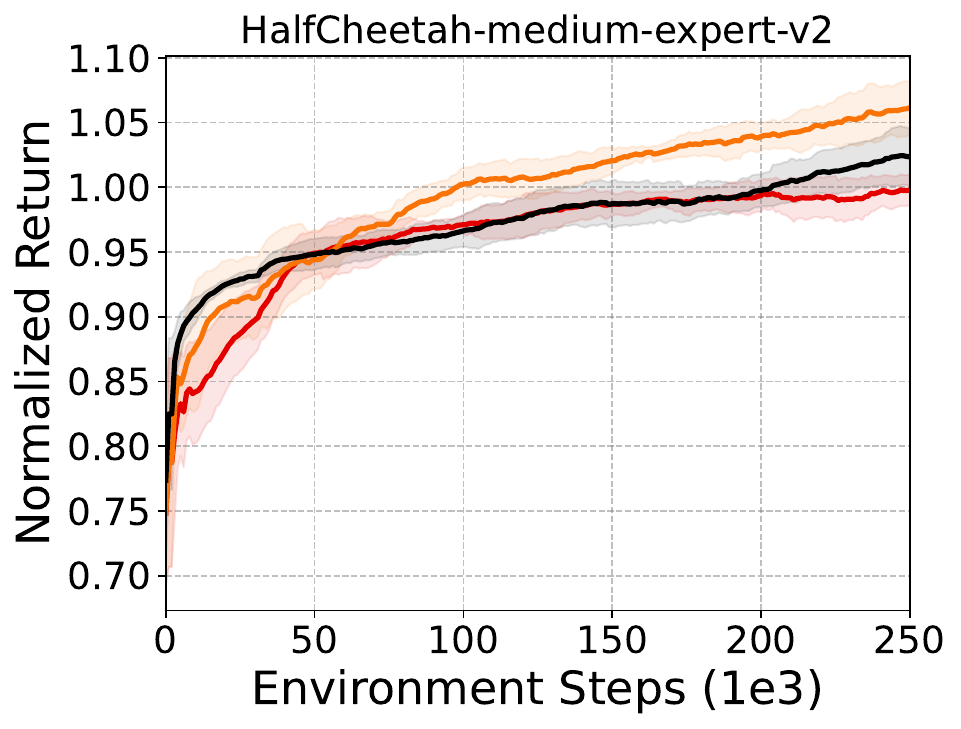}
    \includegraphics[width=0.3\textwidth]{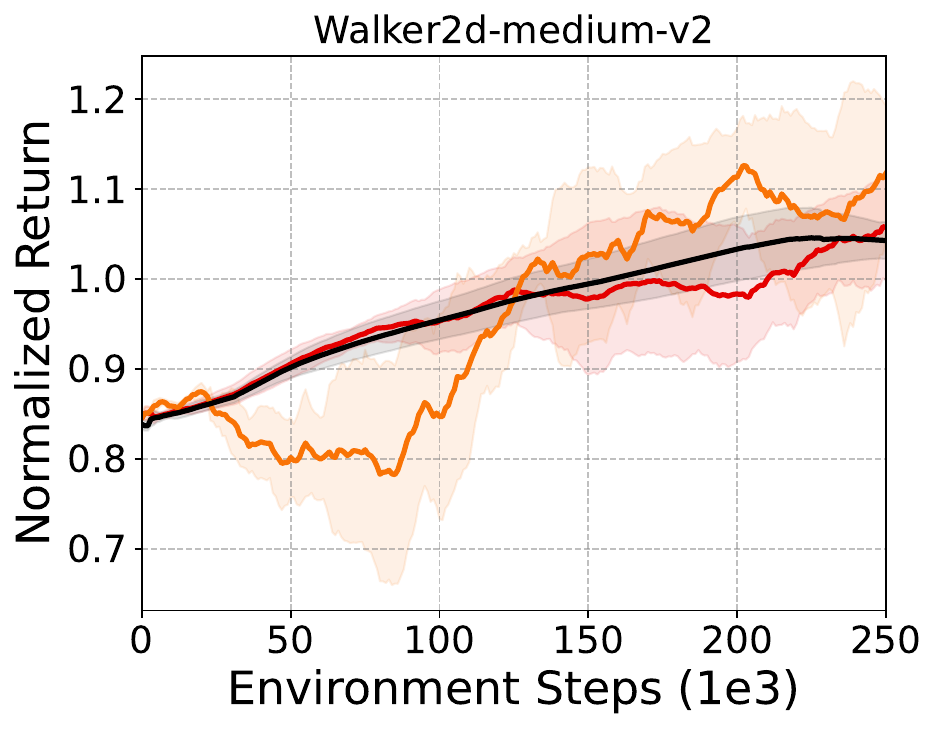}
    \includegraphics[width=0.3\textwidth]{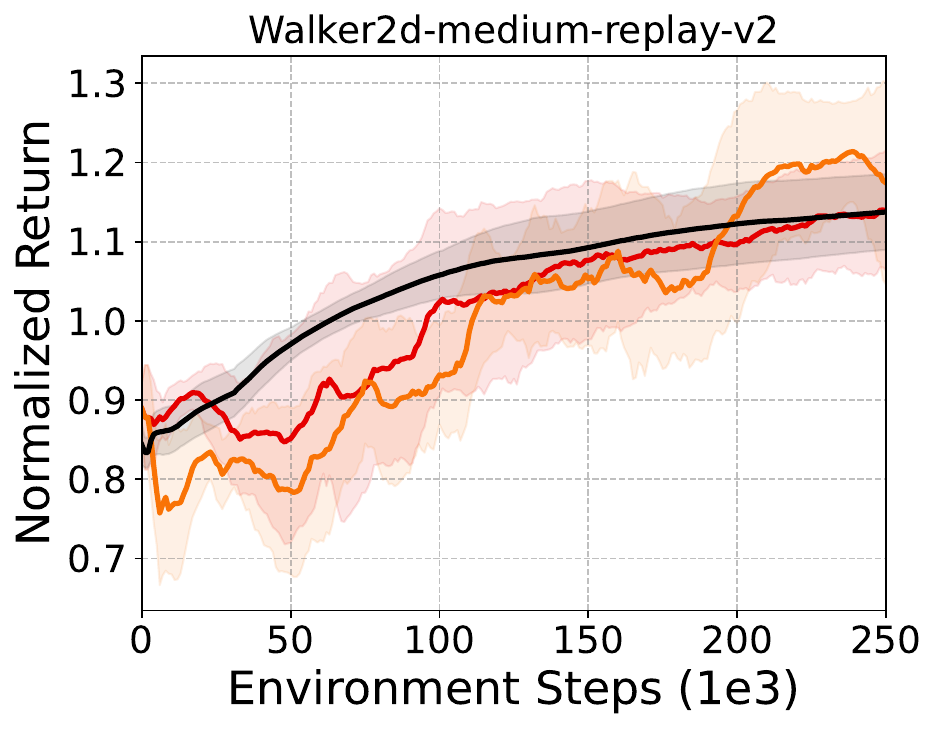}
    \includegraphics[width=0.3\textwidth]{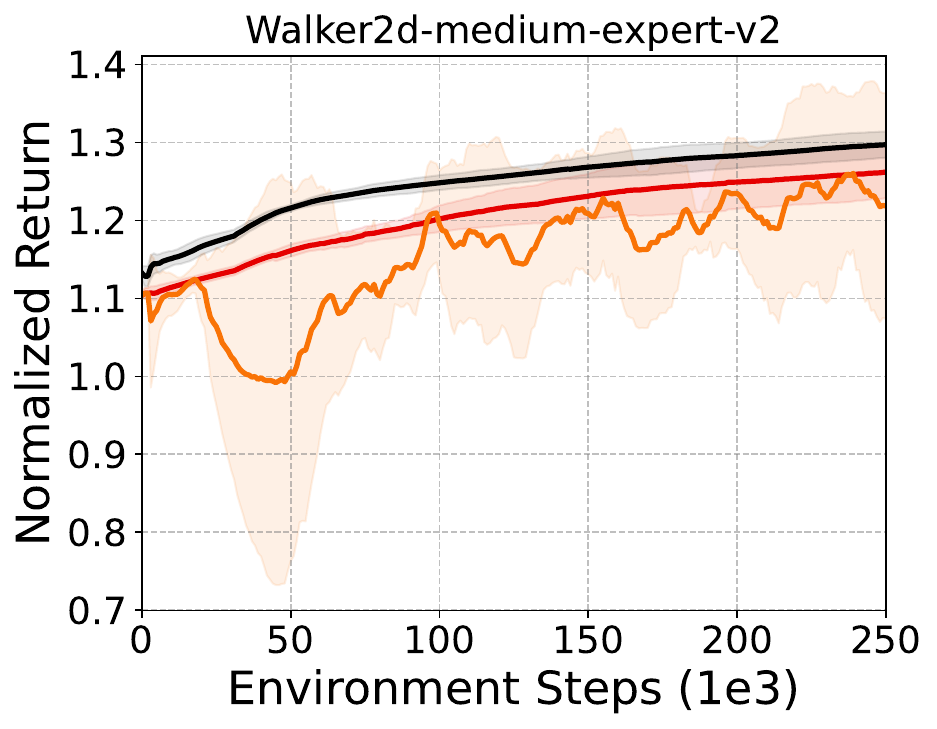}
    \includegraphics[width=0.3\textwidth]{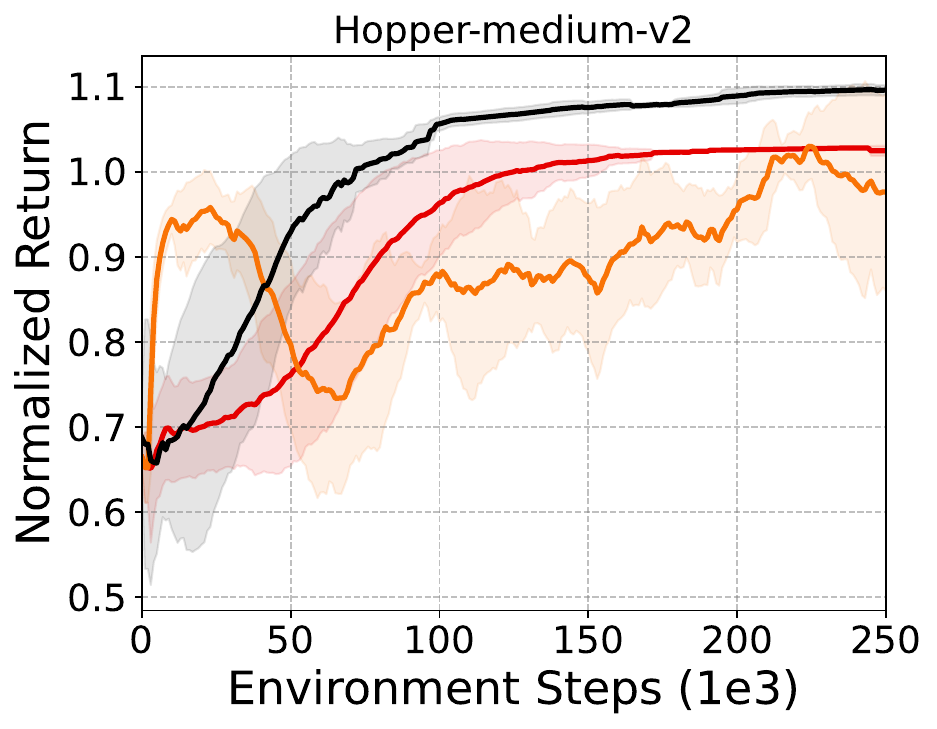}
    \includegraphics[width=0.3\textwidth]{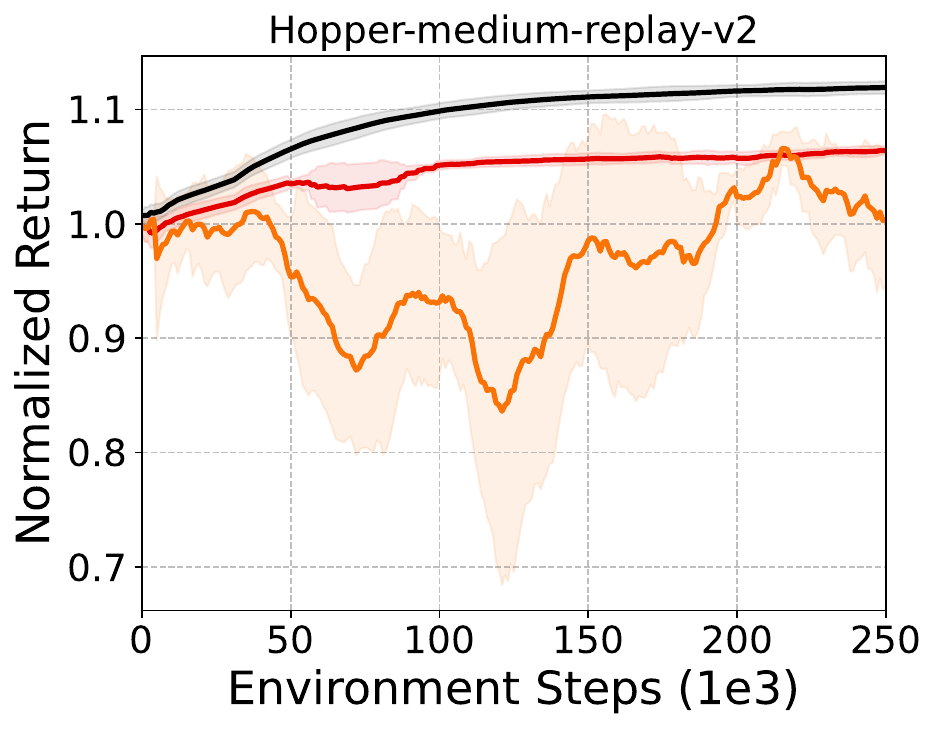}
    \includegraphics[width=0.3\textwidth]{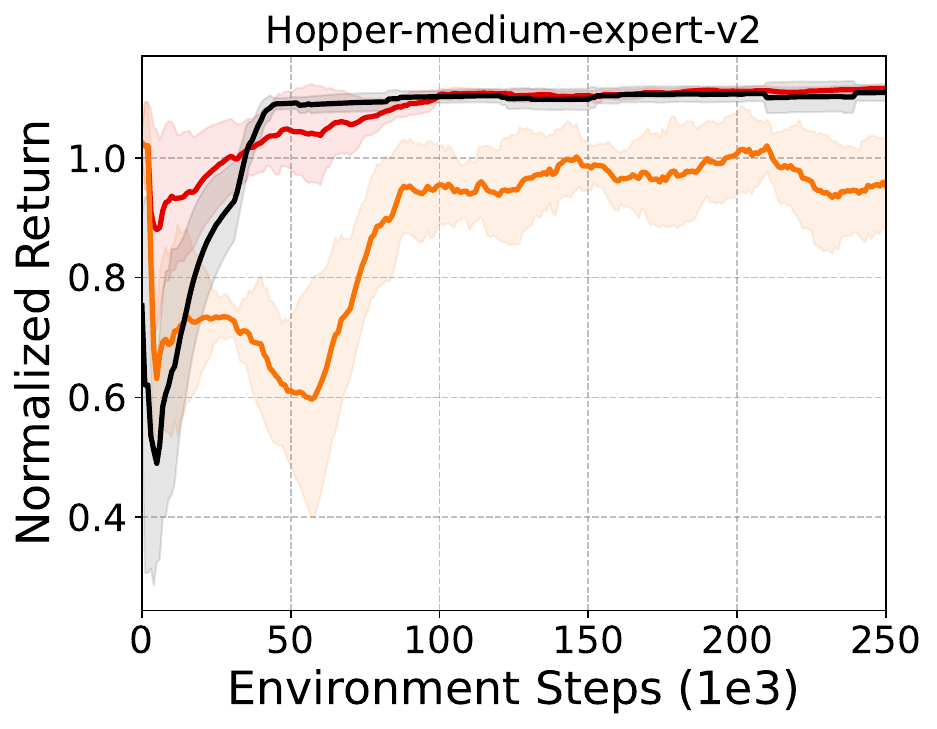}
    \caption{Online learning curves of different offline-to-online approaches across five seeds on MuJoCo locomotion tasks. The solid lines and shaded regions represent mean and standard deviation, respectively.}
    \label{fig:Appendix_method_ensemble}
\end{figure*}

Fig.~\ref{fig:Appendix_method_ensemble} illustrates the performance of various offline-to-online RL approaches on MuJoCo locomotion tasks. It is evident that the \textit{OfflineRL $\to$ OnlineRL} method exhibits the best performance in the HalfCheetah environment. However, it demonstrates unstable learning in the more complex environments of Walker2d and Hopper. On the other hand, the \textit{OfflineRL $\to$ OfflineRL} approach remains stable but shows slower asymptotic performance. In contrast, the \textit{OfflineRL-N $\to$ OnlineRL-N} method no longer experiences performance collapse after transitioning to online fine-tuning, and its training process is relatively stable across all tasks. Additionally, \textit{OfflineRL-N $\to$ OnlineRL-N} achieves superior fine-tuned performance compared to \textit{OfflineRL $\to$ OfflineRL}. It is worth noting that the Hopper-medium-expert-v2 dataset represents a special case where most considered offline-to-online RL methods exhibit varying degrees of performance drop, as depicted in this figure and subsequent figures. Nevertheless, our ENOTO framework consistently achieves state-of-the-art performance in comparison to all baseline methods across most tasks.

\paragraph{Loosing Pessimism}

\begin{figure*}[h]
    \centering
    \includegraphics[width=0.3\textwidth]{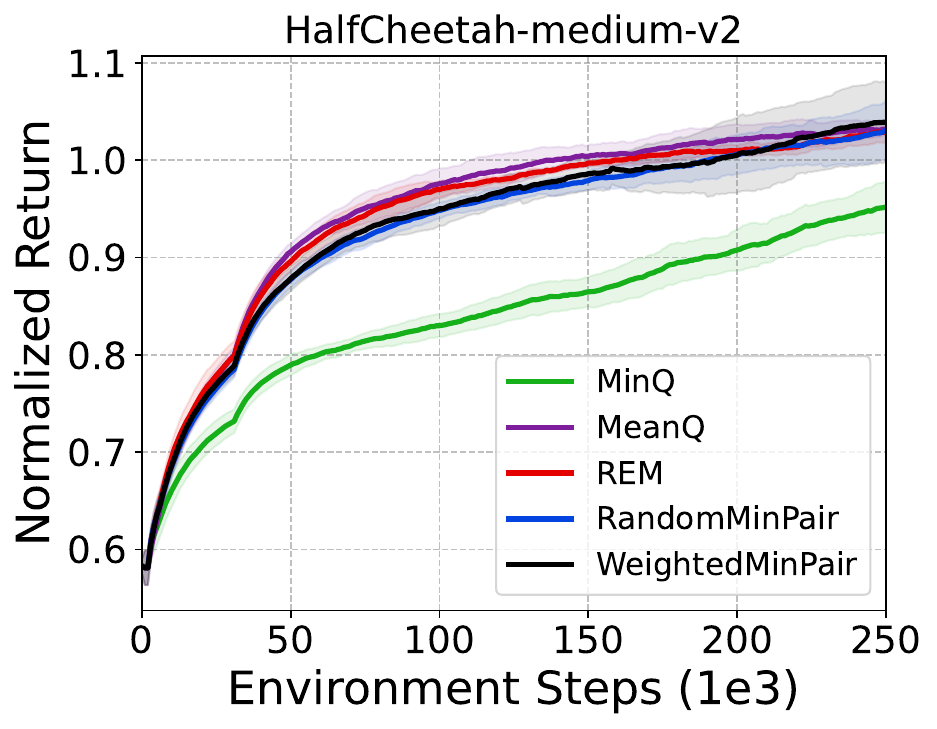}
    \includegraphics[width=0.3\textwidth]{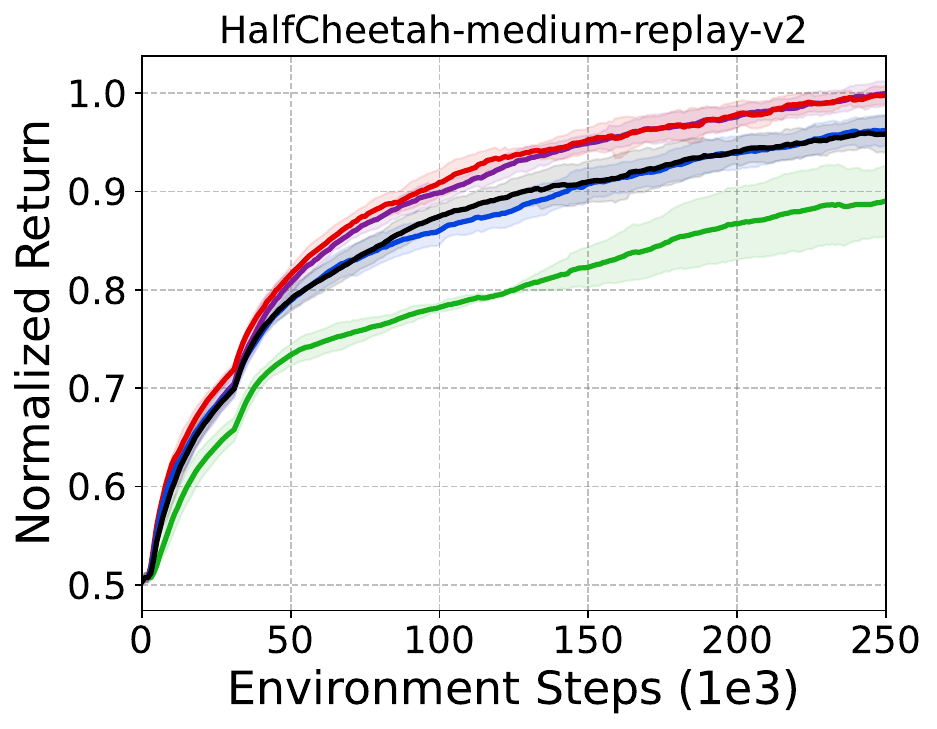}
    \includegraphics[width=0.3\textwidth]{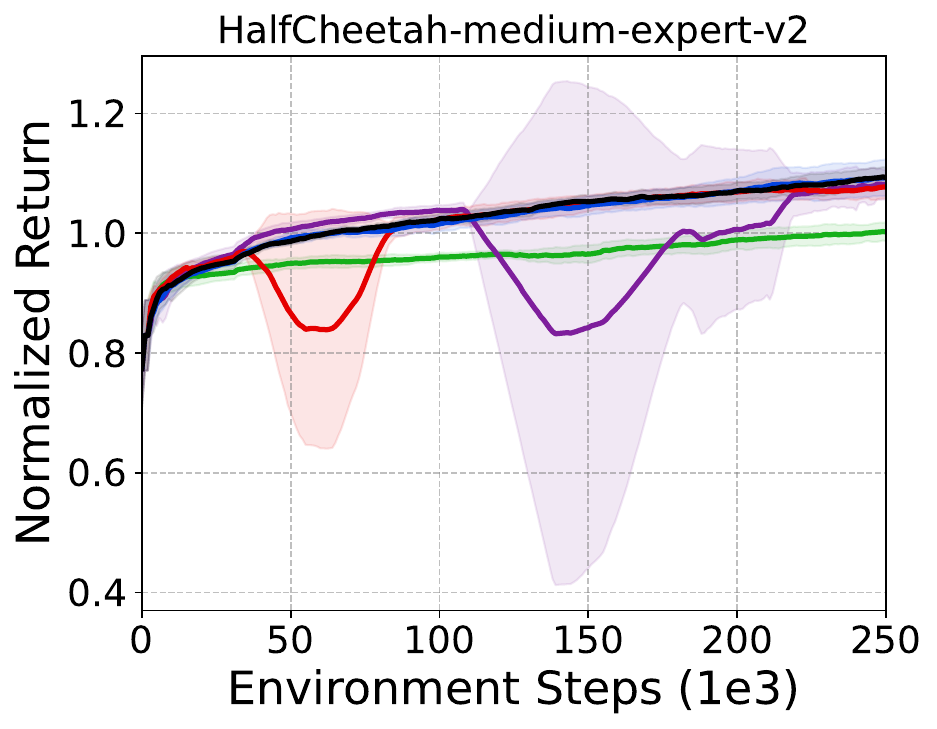}
    \includegraphics[width=0.3\textwidth]{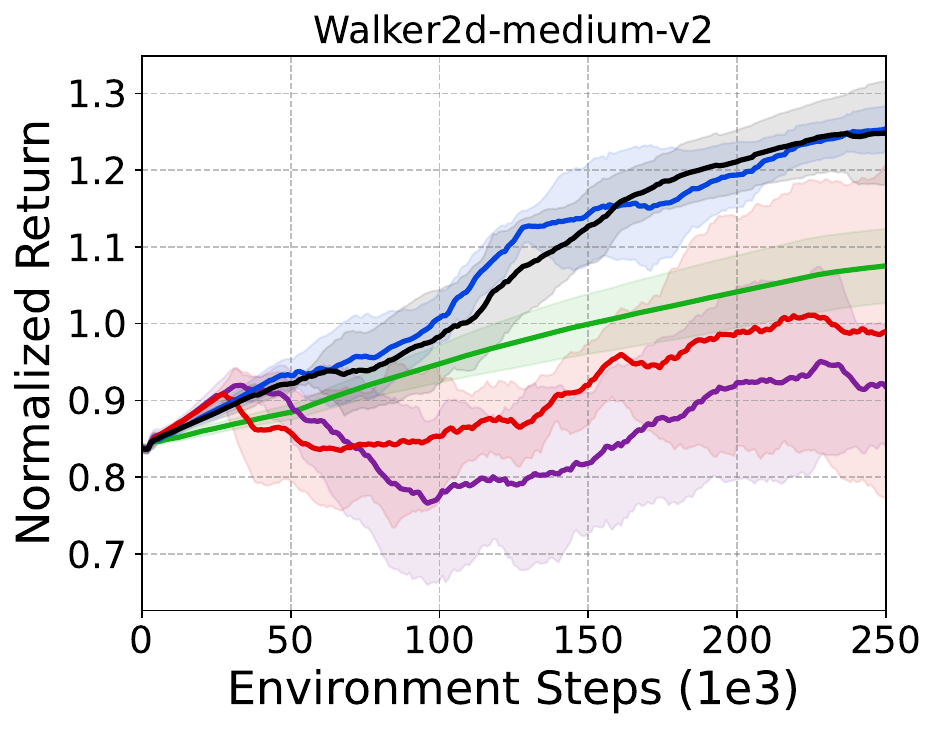}
    \includegraphics[width=0.3\textwidth]{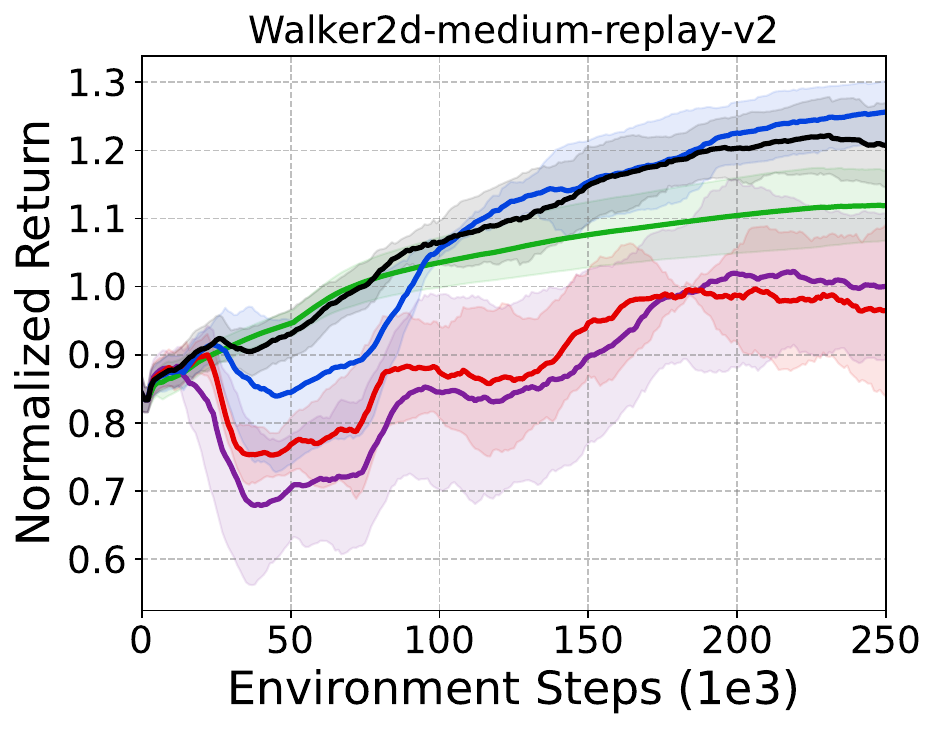}
    \includegraphics[width=0.3\textwidth]{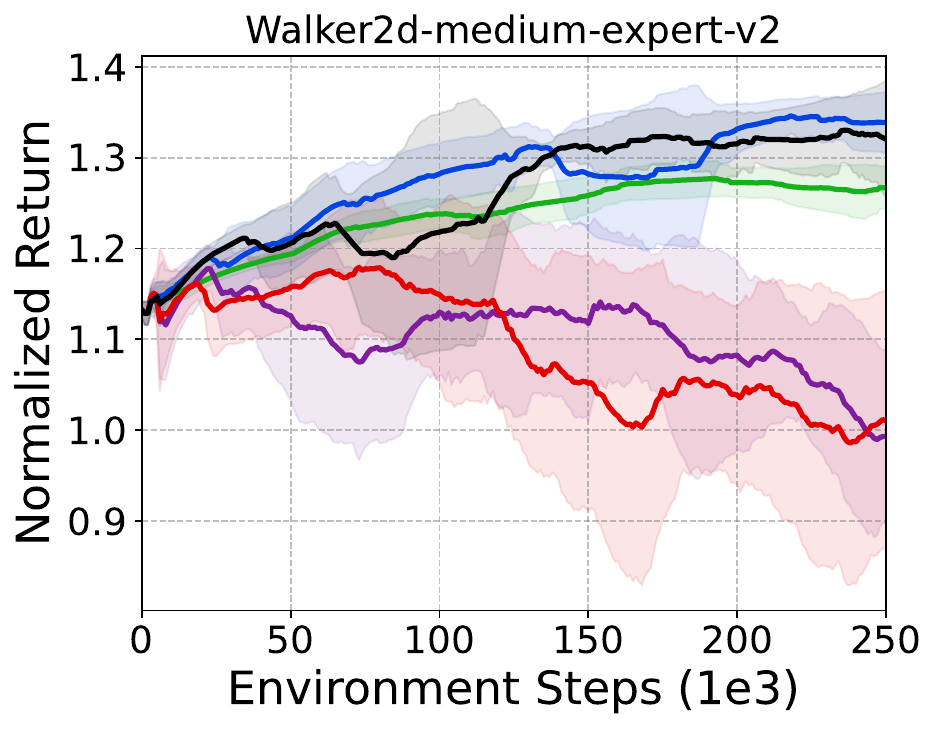}
    \includegraphics[width=0.3\textwidth]{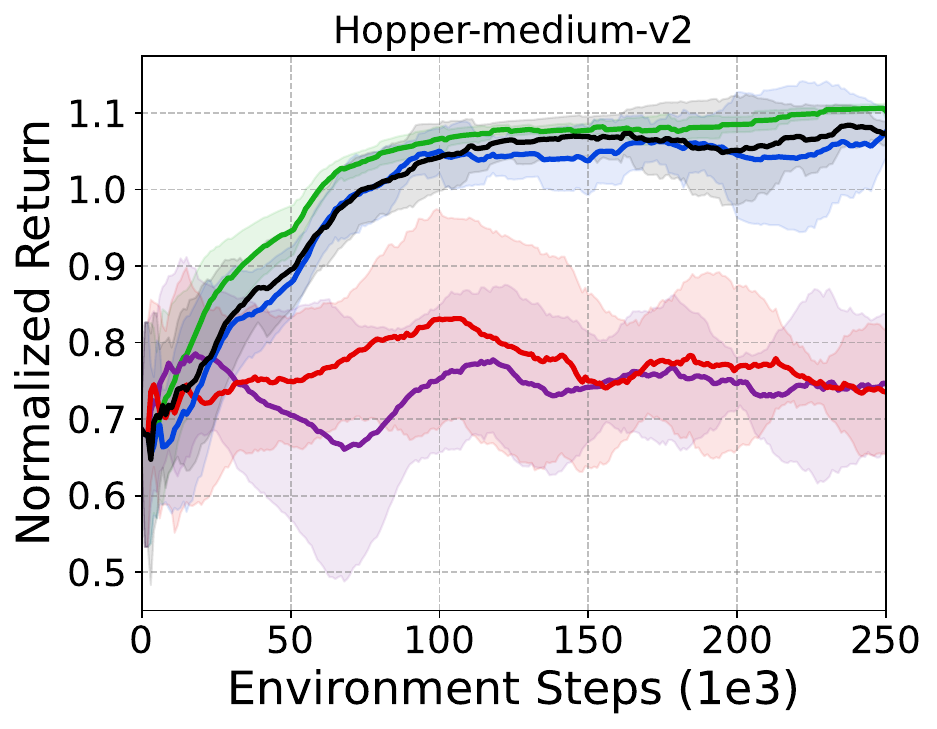}
    \includegraphics[width=0.3\textwidth]{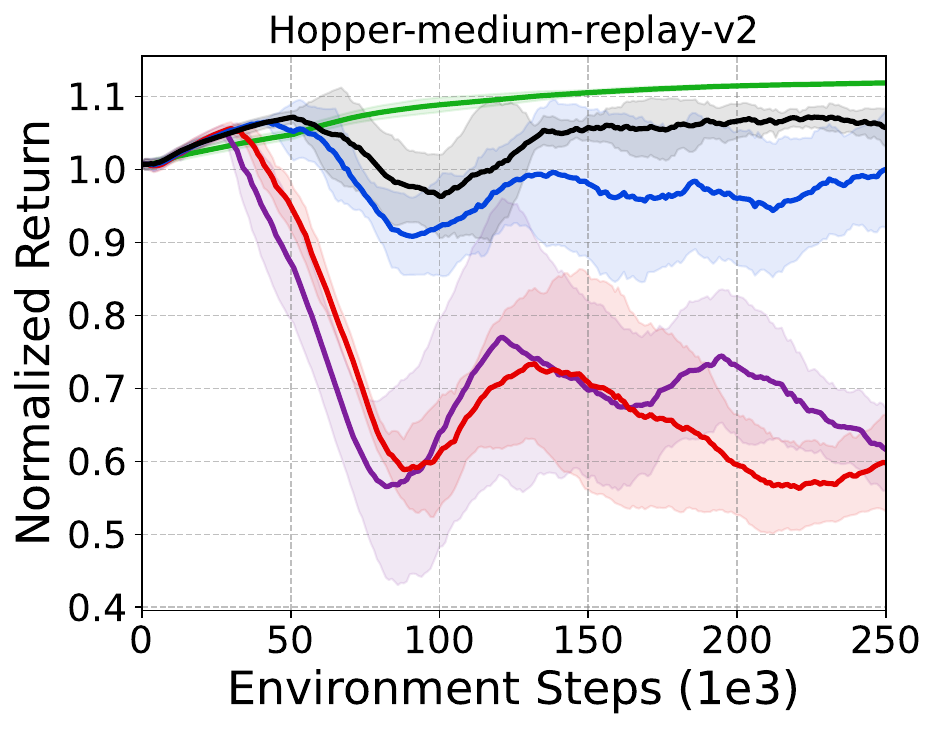}
    \includegraphics[width=0.3\textwidth]{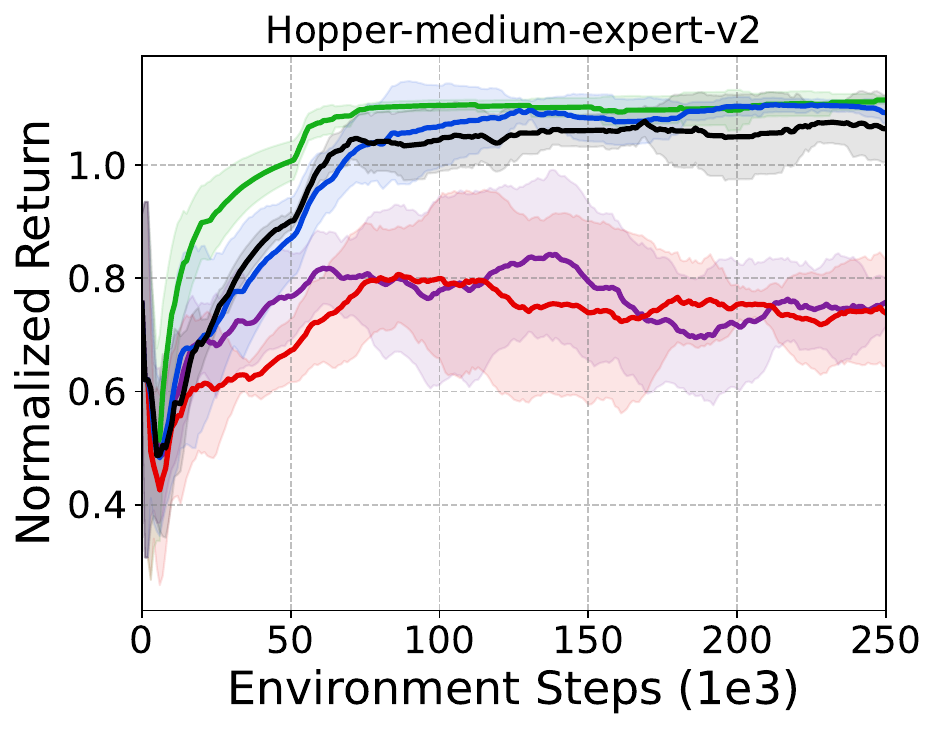}
    \caption{Online learning curves of \textit{OnlineRL-N} using different Q-target computation methods across five seeds on MuJoCo locomotion tasks. The solid lines and shaded regions represent mean and standard deviation, respectively.}
    \label{fig:Appendix_method_targetq}
\end{figure*}

Fig.~\ref{fig:Appendix_method_targetq} displays the performance of \textit{OnlineRL-N} utilizing different Q-target computation methods on MuJoCo locomotion tasks. It is evident that MinQ exhibits remarkable stability across all tasks, albeit with slower performance improvement in the HalfCheetah and Walker2d environments. MeanQ and REM demonstrate excellent performance in the HalfCheetah environment, but struggle to improve in the more challenging environments of Walker2d and Hopper, and their learning process is characterized by instability. In contrast, RandomMinPair and WeightedMinPair showcase superior performance across most tasks, with the exception of the Hopper-medium-replay-v2 dataset where they exhibit slight instability in learning. Among these two methods, WeightedMinPair demonstrates slightly better stability and performance, thus we select it as the component of our final ENOTO framework and present the experiments related to RandomMinPair in the appendix.

\paragraph{Optimistic Exploration}

\begin{figure*}[h]
    \centering
    \includegraphics[width=0.3\textwidth]{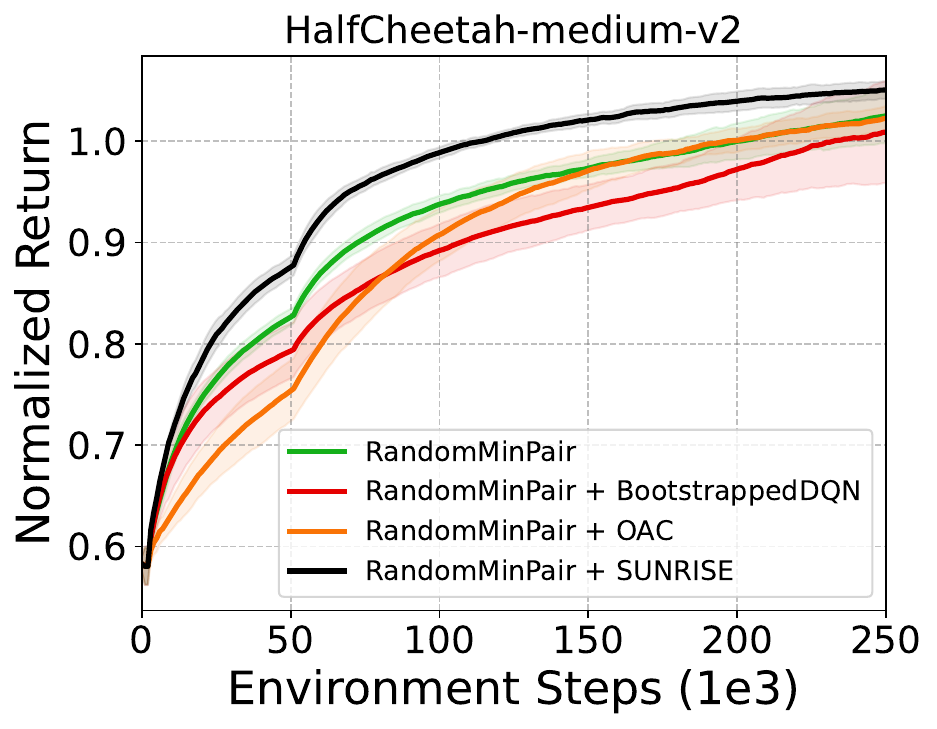}
    \includegraphics[width=0.3\textwidth]{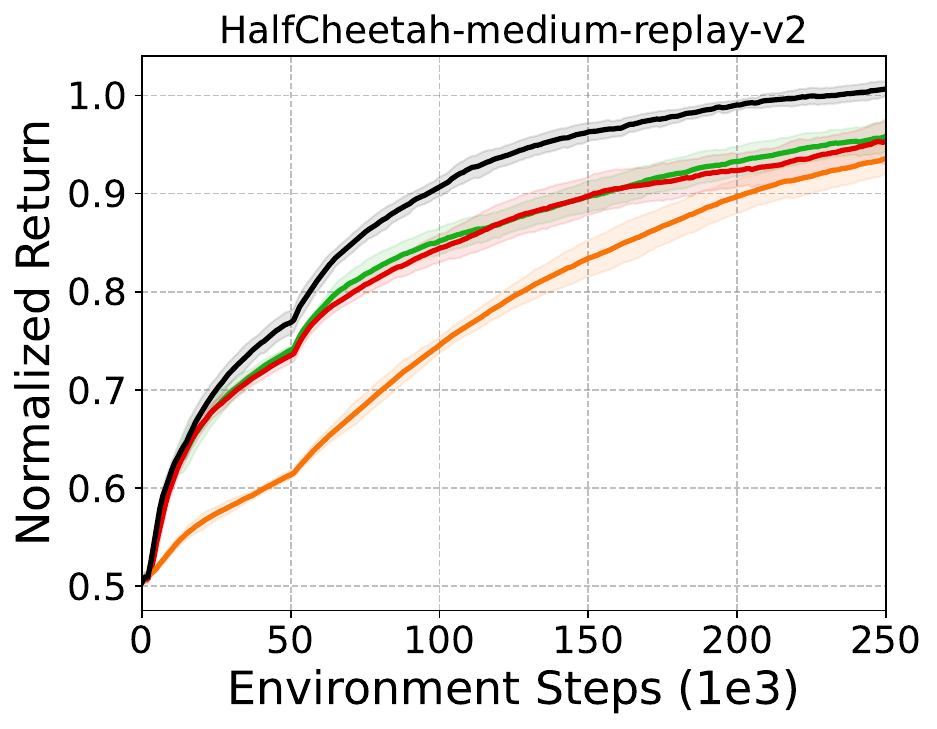}
    \includegraphics[width=0.3\textwidth]{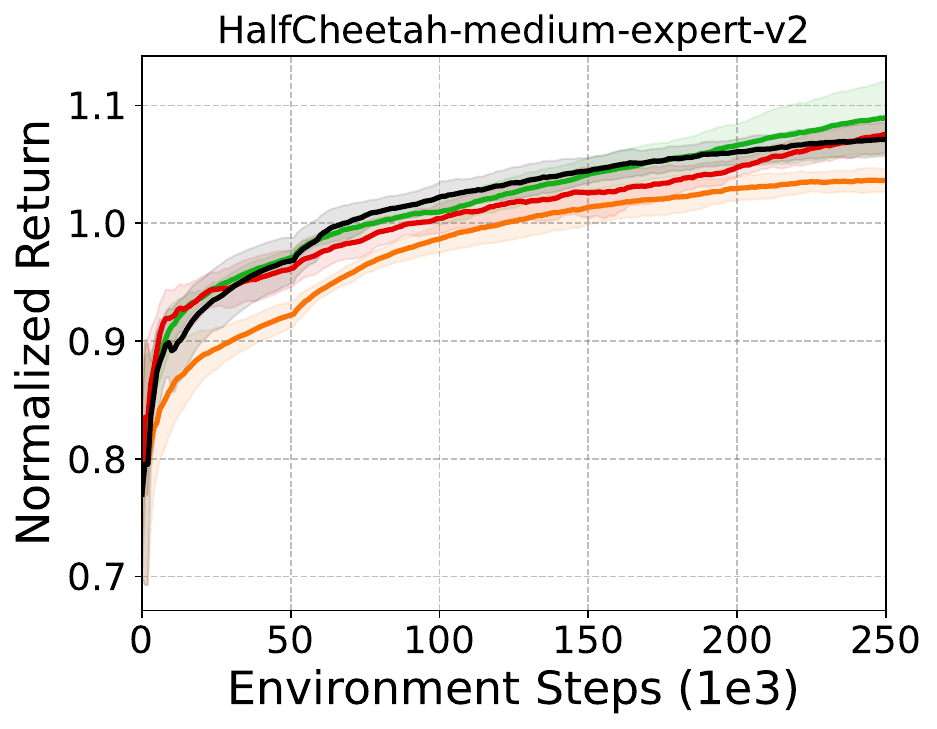}
    \includegraphics[width=0.3\textwidth]{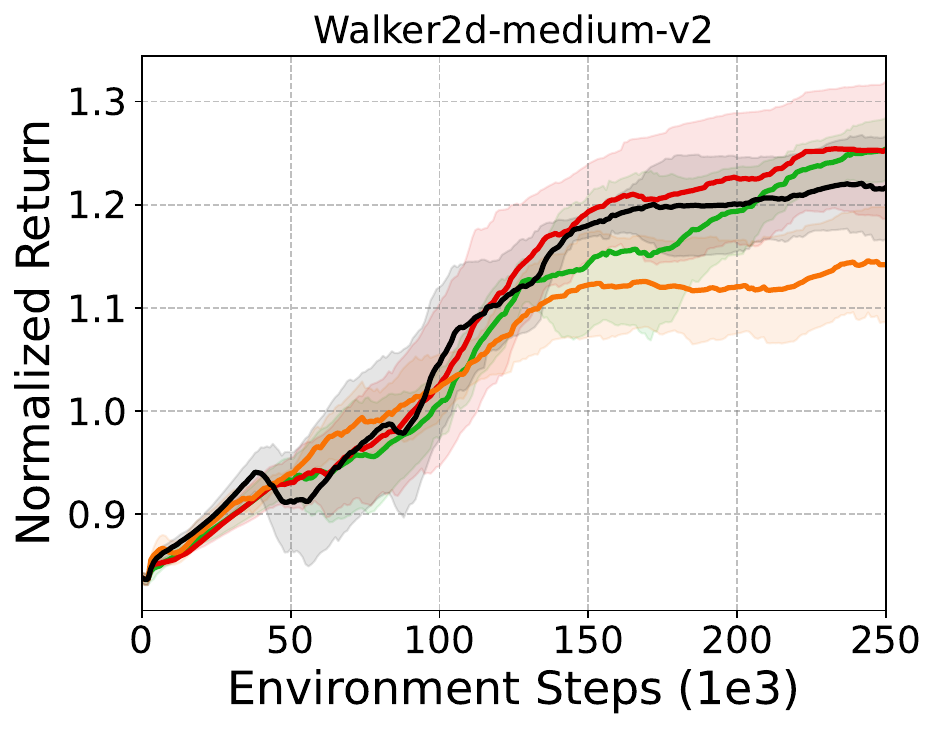}
    \includegraphics[width=0.3\textwidth]{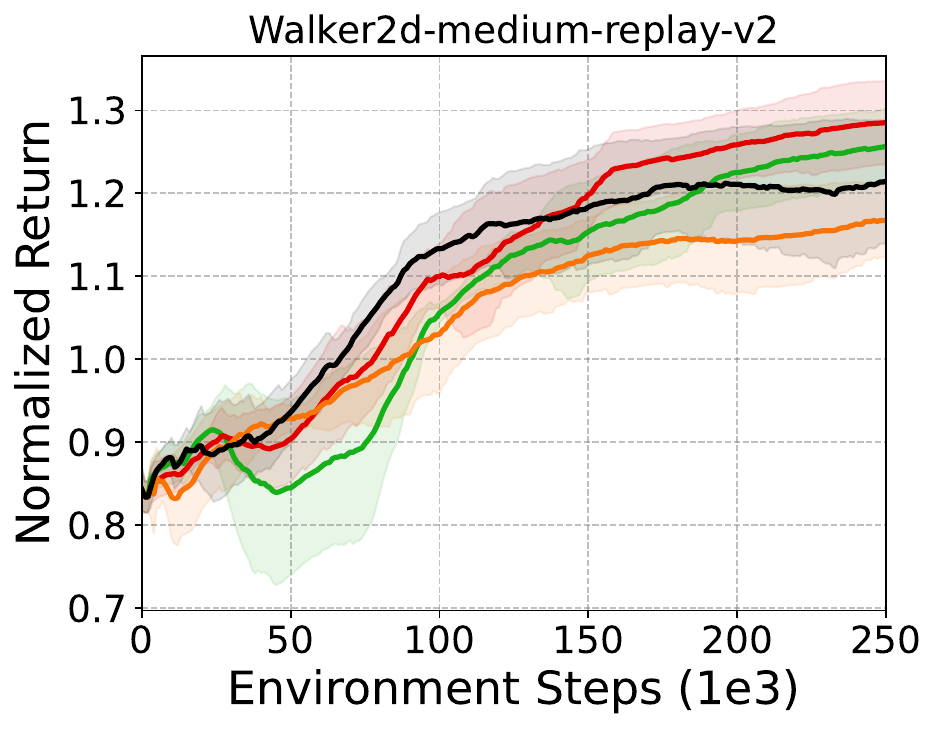}
    \includegraphics[width=0.3\textwidth]{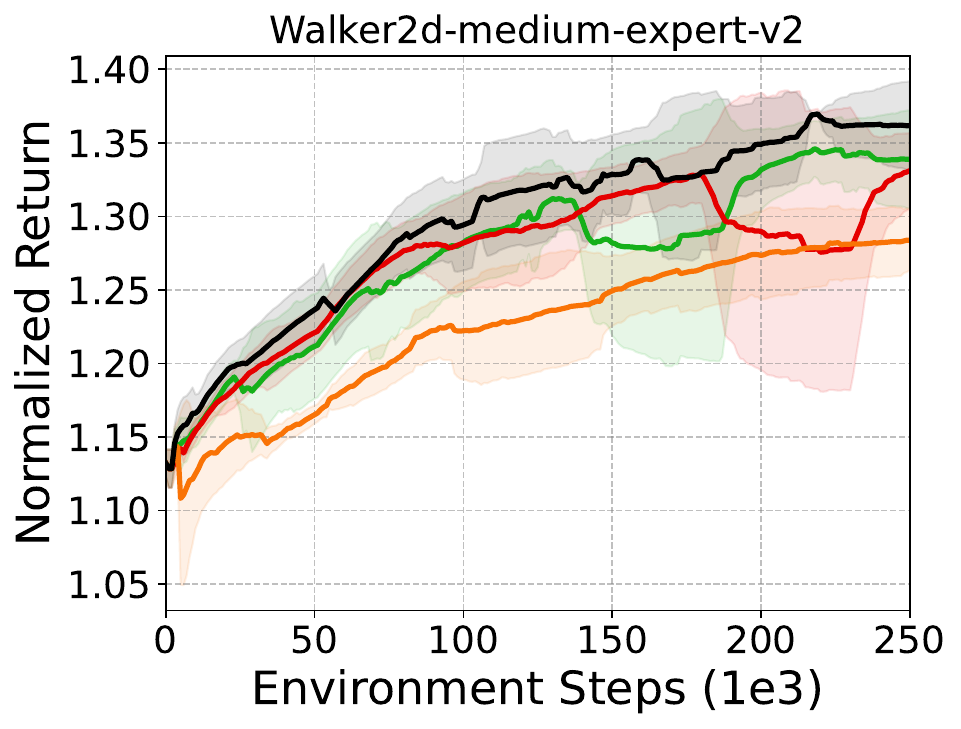}
    \includegraphics[width=0.3\textwidth]{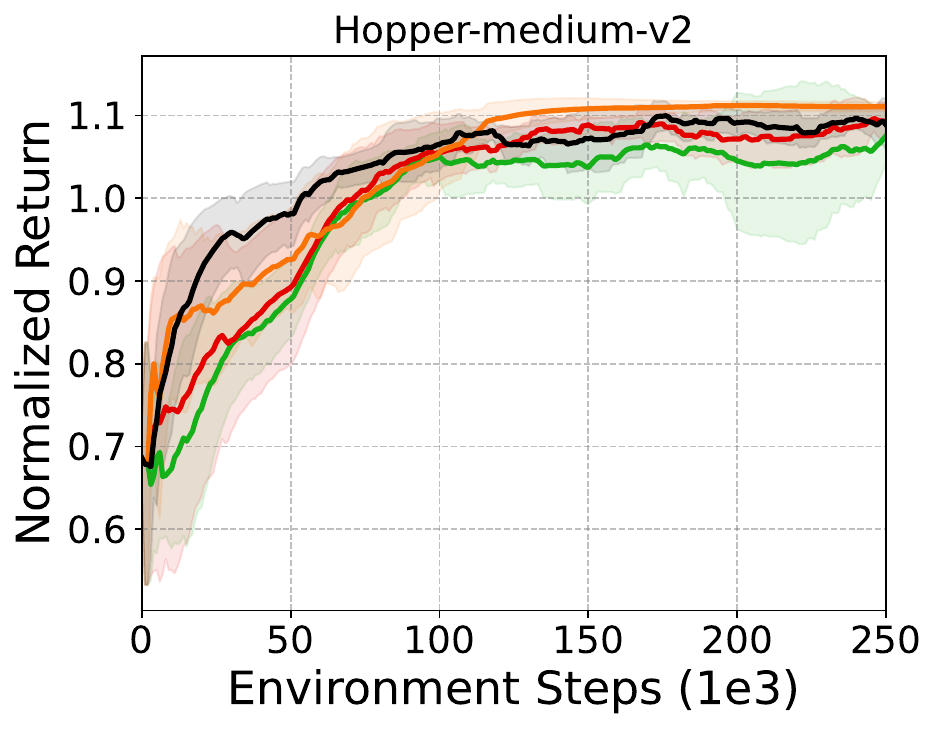}
    \includegraphics[width=0.3\textwidth]{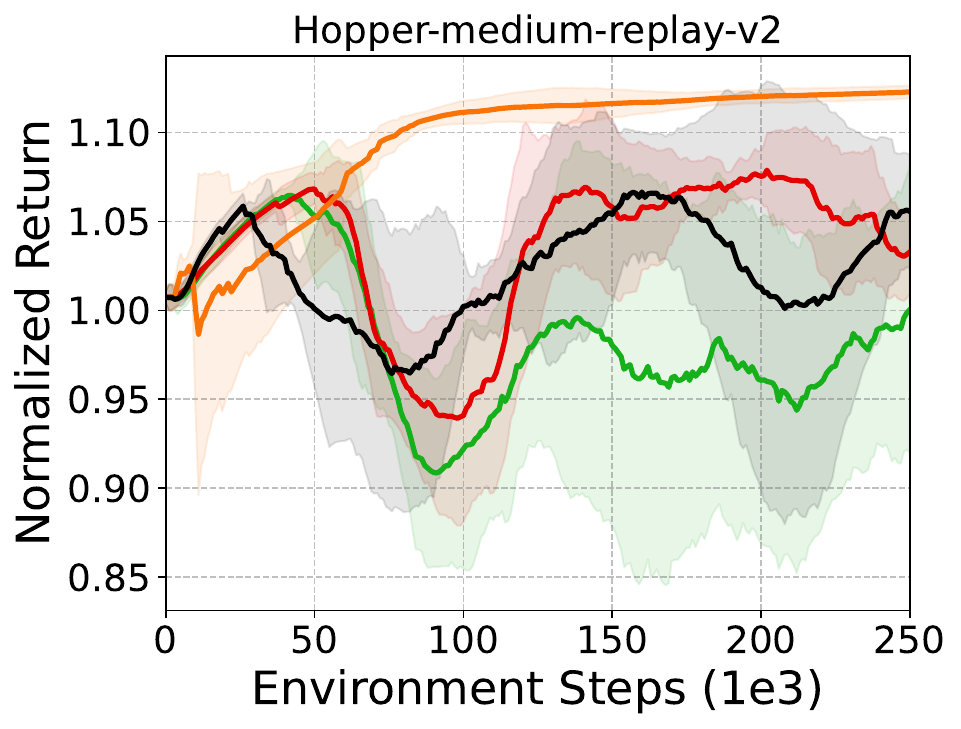}
    \includegraphics[width=0.3\textwidth]{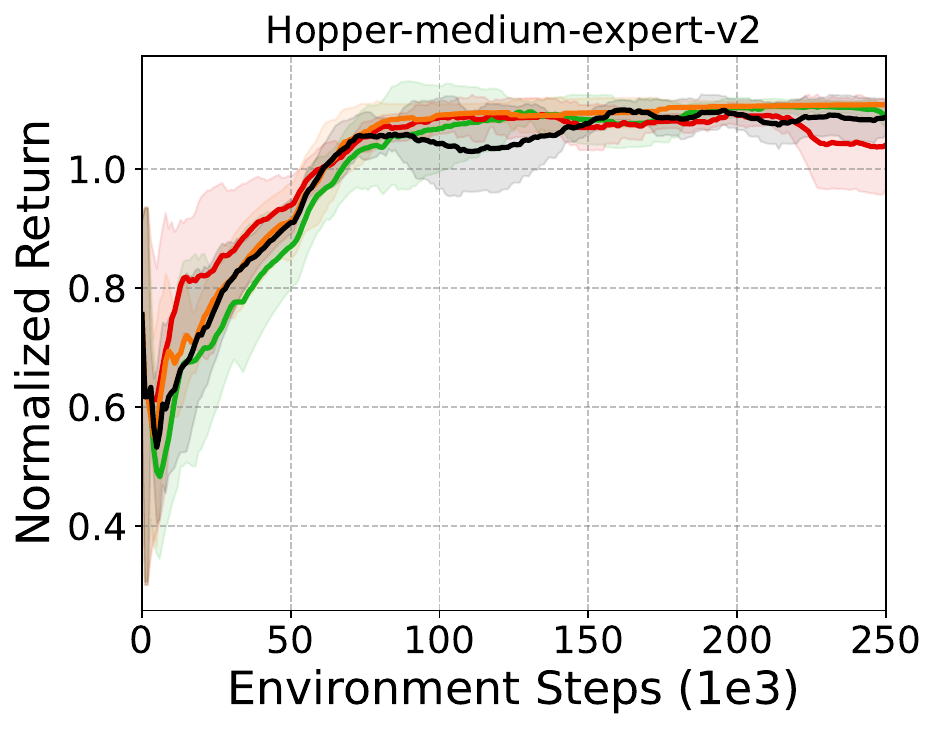}
    \caption{Online learning curves of \textit{OnlineRL-N + RandomMinPair} using different exploration methods across five seeds on MuJoCo locomotion tasks. The solid lines and shaded regions represent mean and standard deviation, respectively.}
    \label{fig:Appendix_method_exploration_random}
\end{figure*}

Fig.~\ref{fig:Appendix_method_exploration_weighted} and Fig.~\ref{fig:Appendix_method_exploration_random} present the performance of \textit{OnlineRL-N + WeightedMinPair} and \textit{OnlineRL-N + RandomMinPair} using different exploration methods on MuJoCo locomotion tasks, respectively. These two figures exhibit similar observations. In the Hopper environment, OAC achieves the best performance, but its performance improvement in HalfCheetah and Walker2d is relatively slow. The use of Bootstrapped DQN leads to minimal improvement in performance, while SUNRISE enhances the learning efficiency of \textit{OnlineRL-N + WeightedMinPair} across most tasks, with the exception of the Hopper-medium-replay-v2 dataset where they exhibit slight instability in learning.

\begin{figure*}[h]
    \centering
    \includegraphics[width=0.3\textwidth]{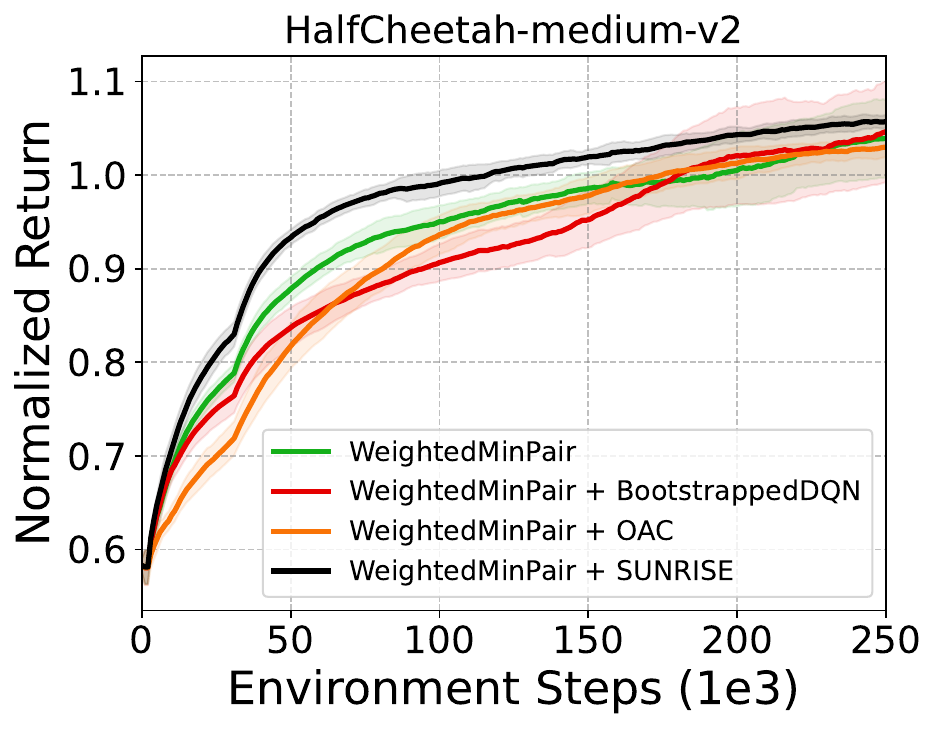}
    \includegraphics[width=0.3\textwidth]{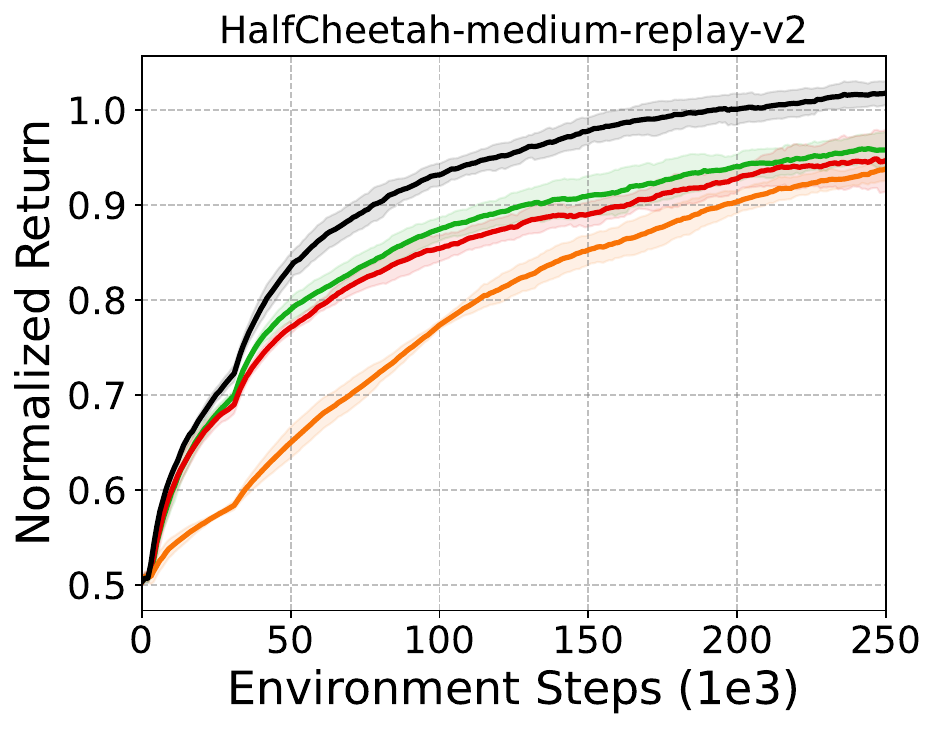}
    \includegraphics[width=0.3\textwidth]{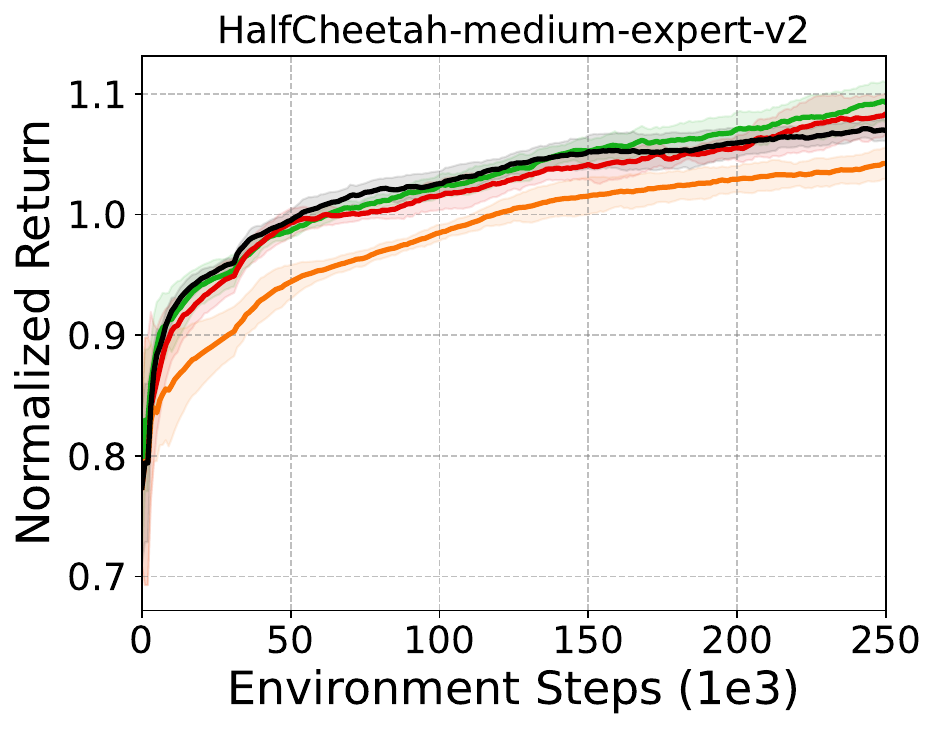}
    \includegraphics[width=0.3\textwidth]{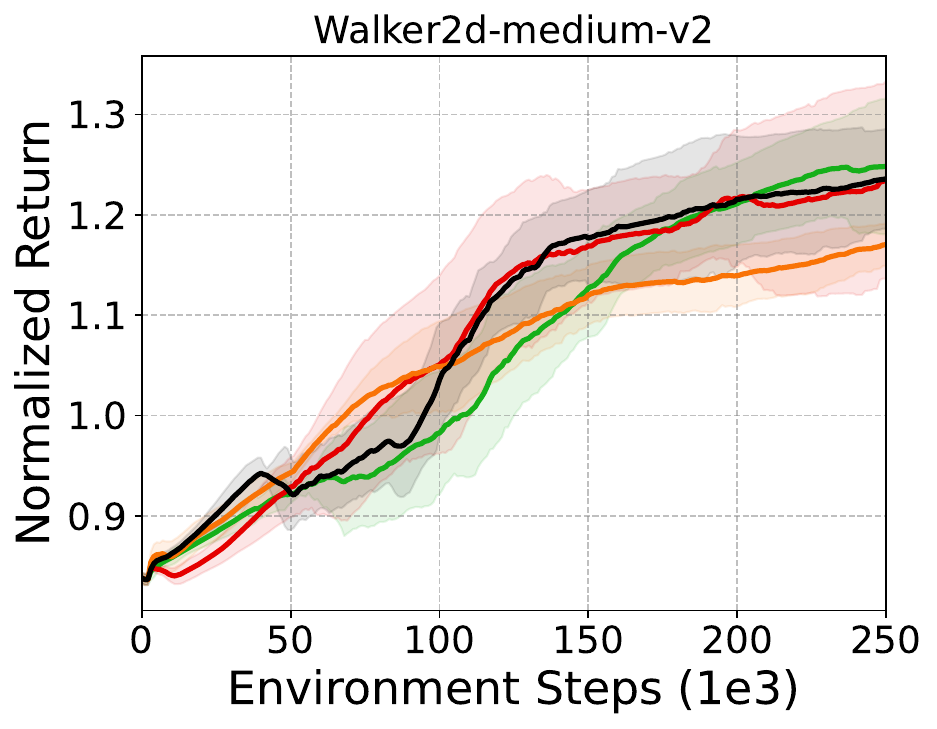}
    \includegraphics[width=0.3\textwidth]{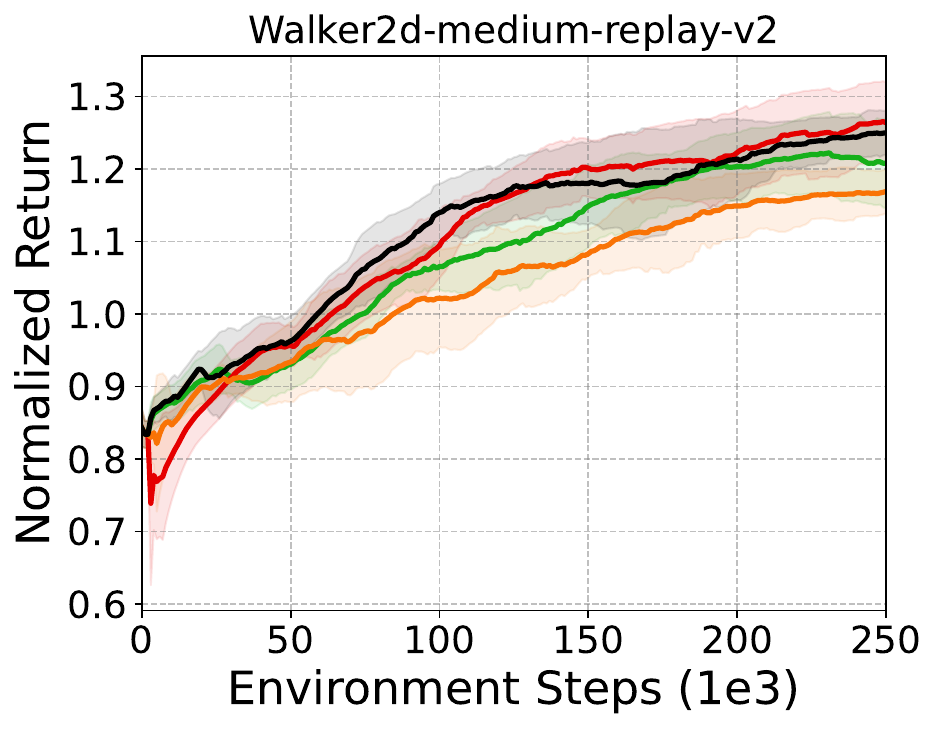}
    \includegraphics[width=0.3\textwidth]{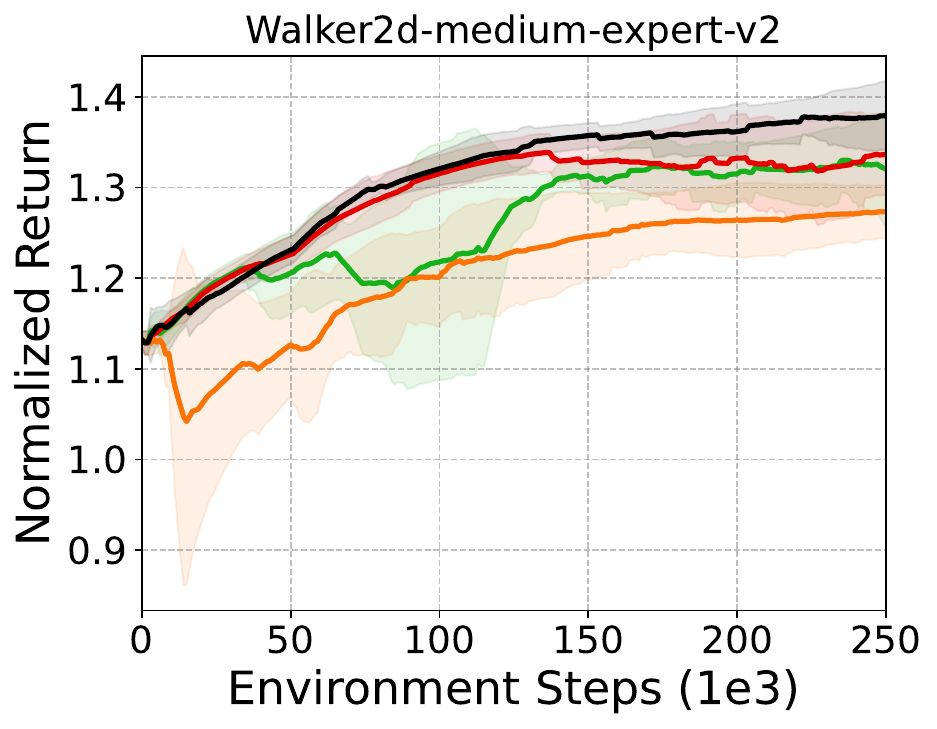}
    \includegraphics[width=0.3\textwidth]{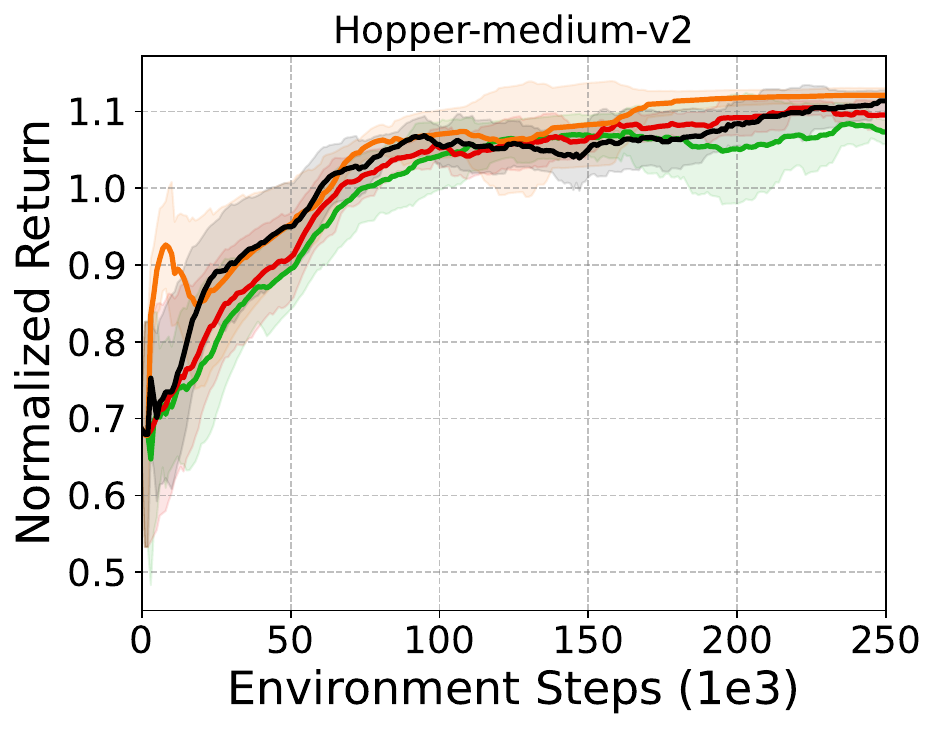}
    \includegraphics[width=0.3\textwidth]{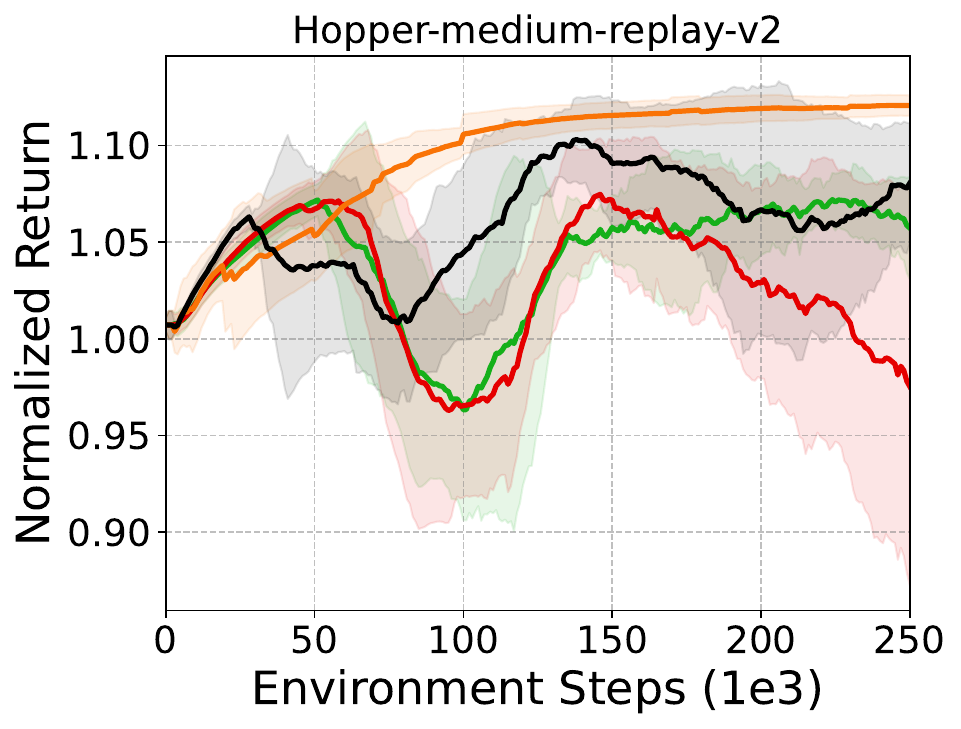}
    \includegraphics[width=0.3\textwidth]{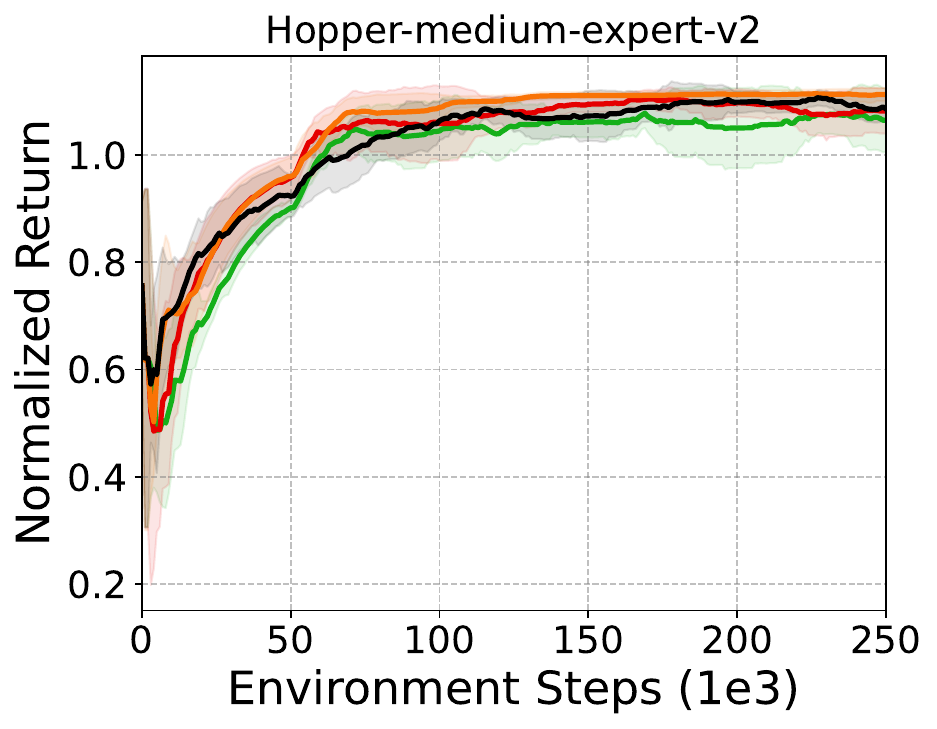}
    \caption{Online learning curves of \textit{OnlineRL-N + WeightedMinPair} using different exploration methods across five seeds on MuJoCo locomotion tasks. The solid lines and shaded regions represent mean and standard deviation, respectively.}
    \label{fig:Appendix_method_exploration_weighted}
\end{figure*}

\subsection{Offline Performance}

In this section, we provide the offline learning curves of different methods on locomotion and navigation tasks.

\paragraph{Locomotion Tasks}

\begin{figure*}[h]
    \centering
    \includegraphics[width=0.3\textwidth]{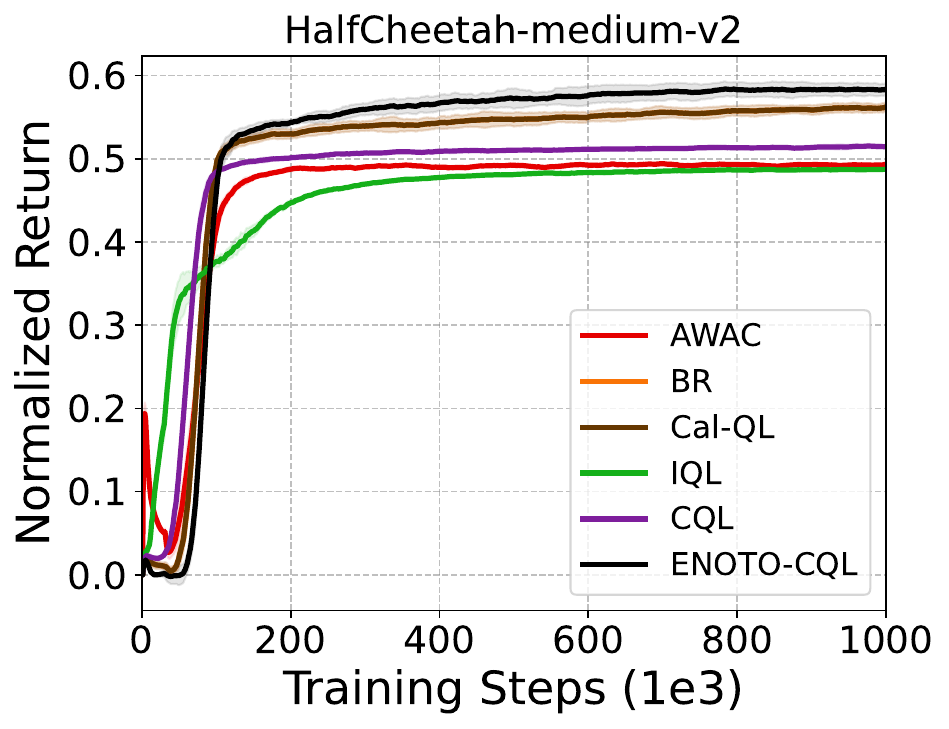}
    \includegraphics[width=0.3\textwidth]{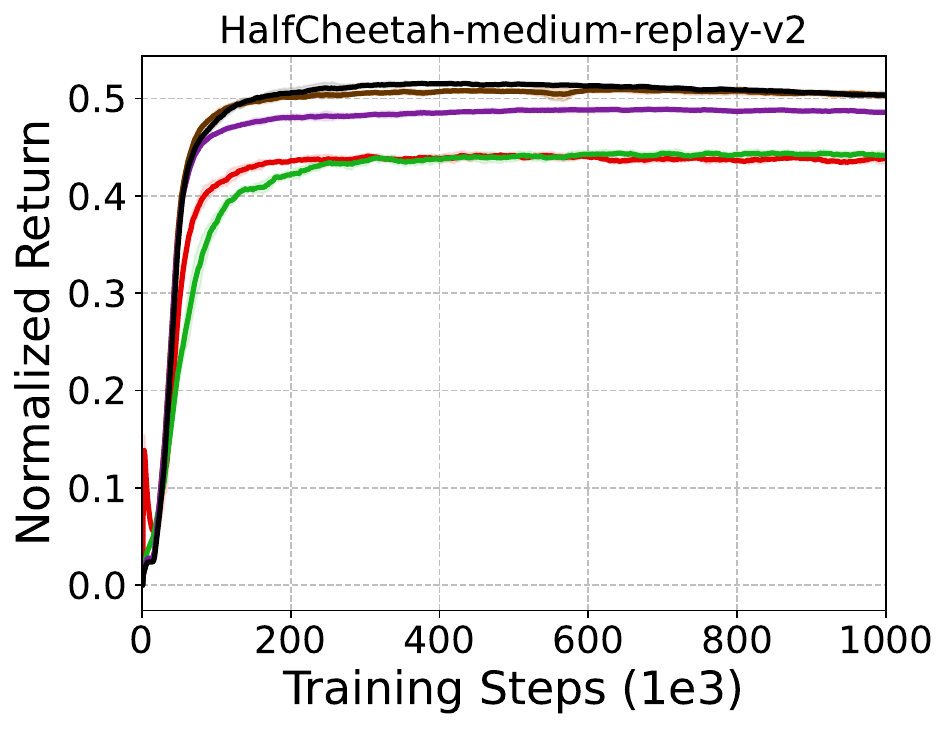}
    \includegraphics[width=0.3\textwidth]{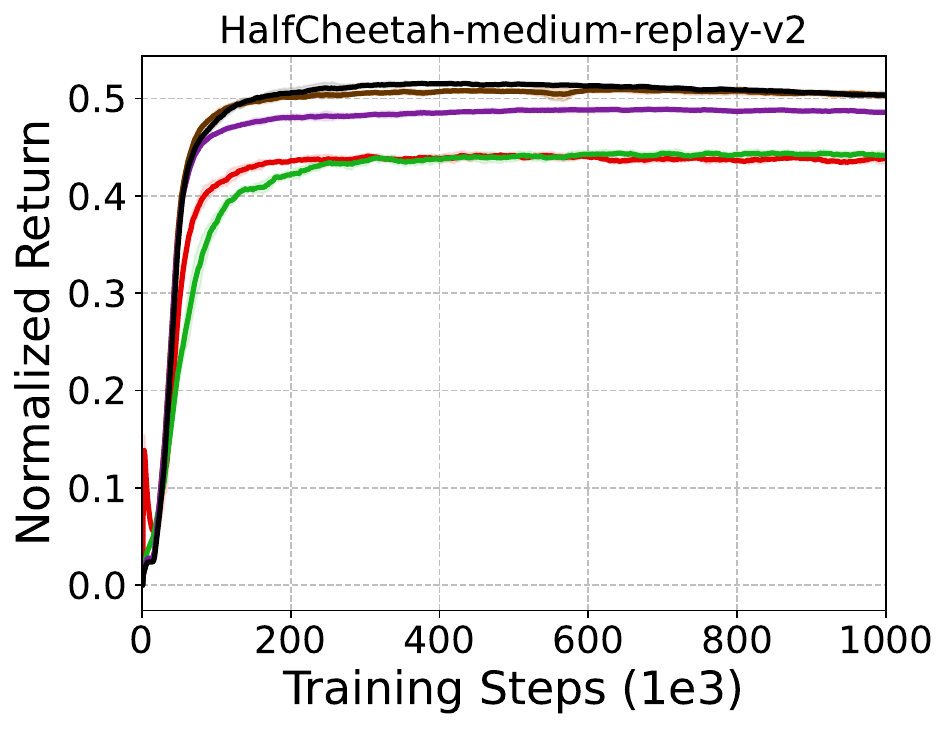}
    \includegraphics[width=0.3\textwidth]{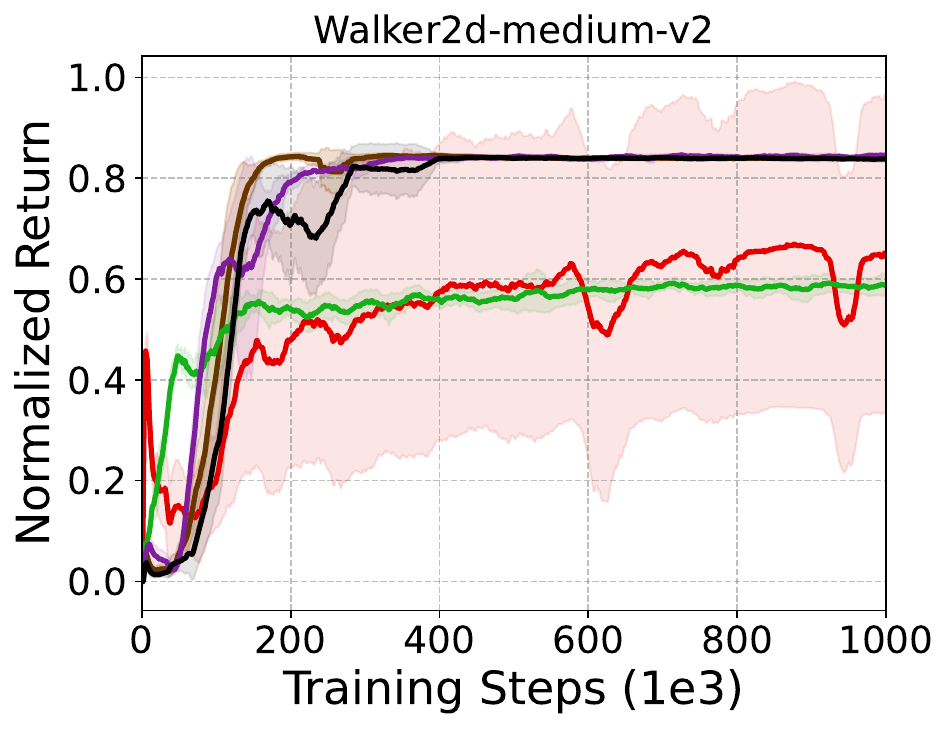}
    \includegraphics[width=0.3\textwidth]{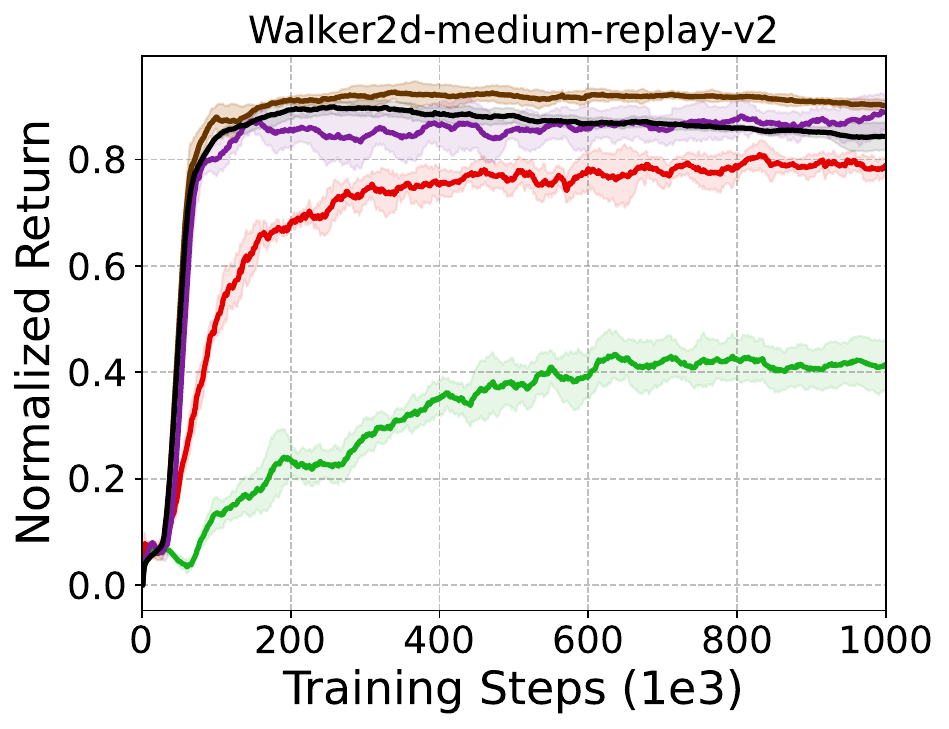}
    \includegraphics[width=0.3\textwidth]{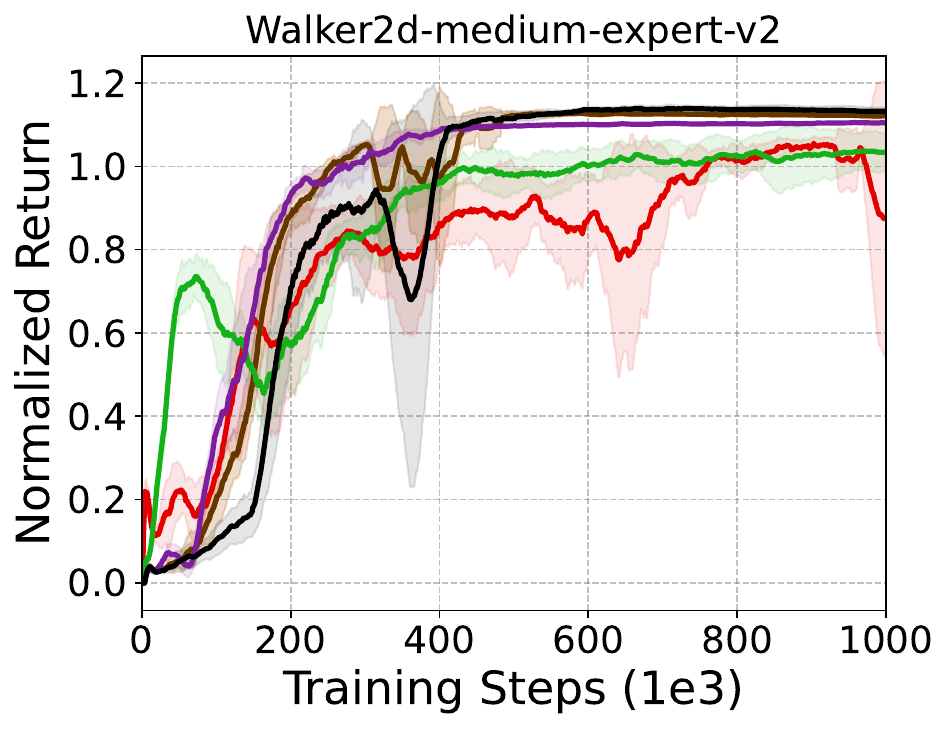}
    \includegraphics[width=0.3\textwidth]{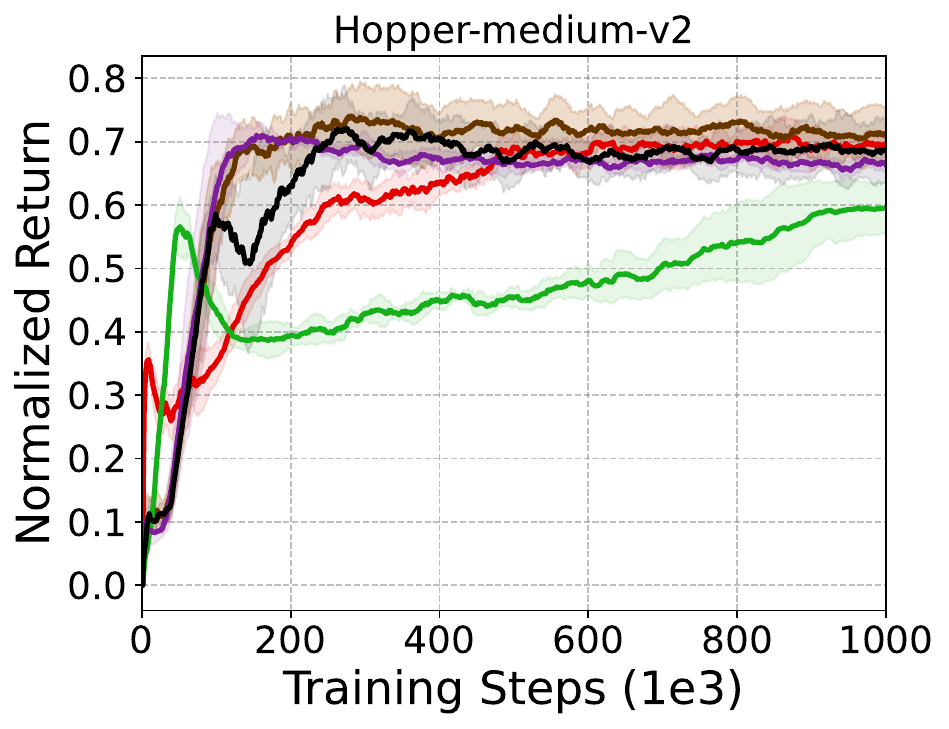}
    \includegraphics[width=0.3\textwidth]{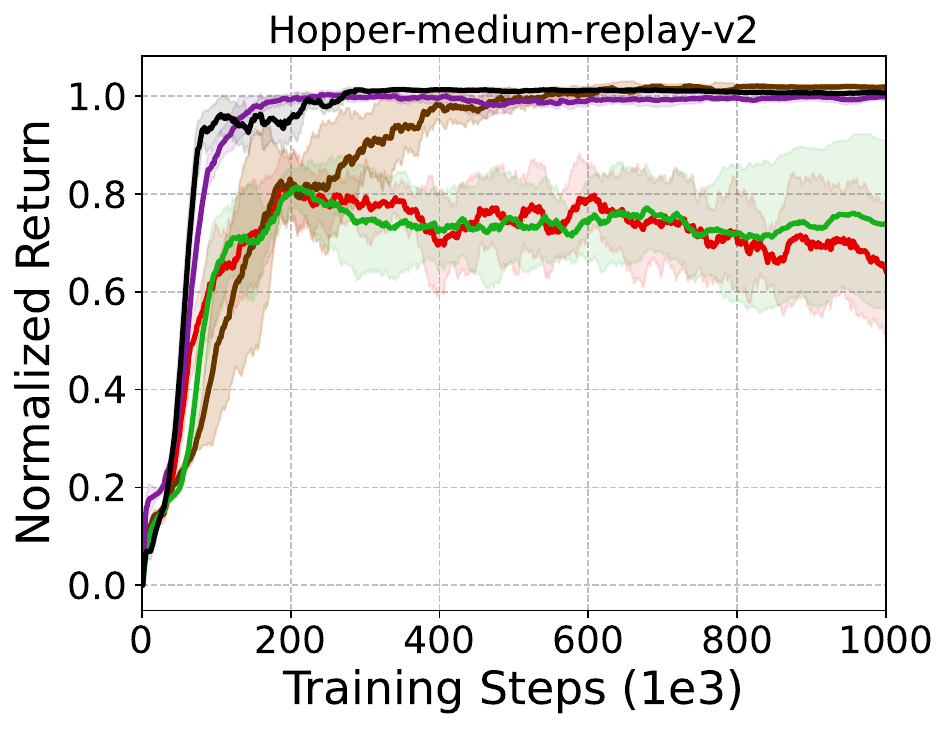}
    \includegraphics[width=0.3\textwidth]{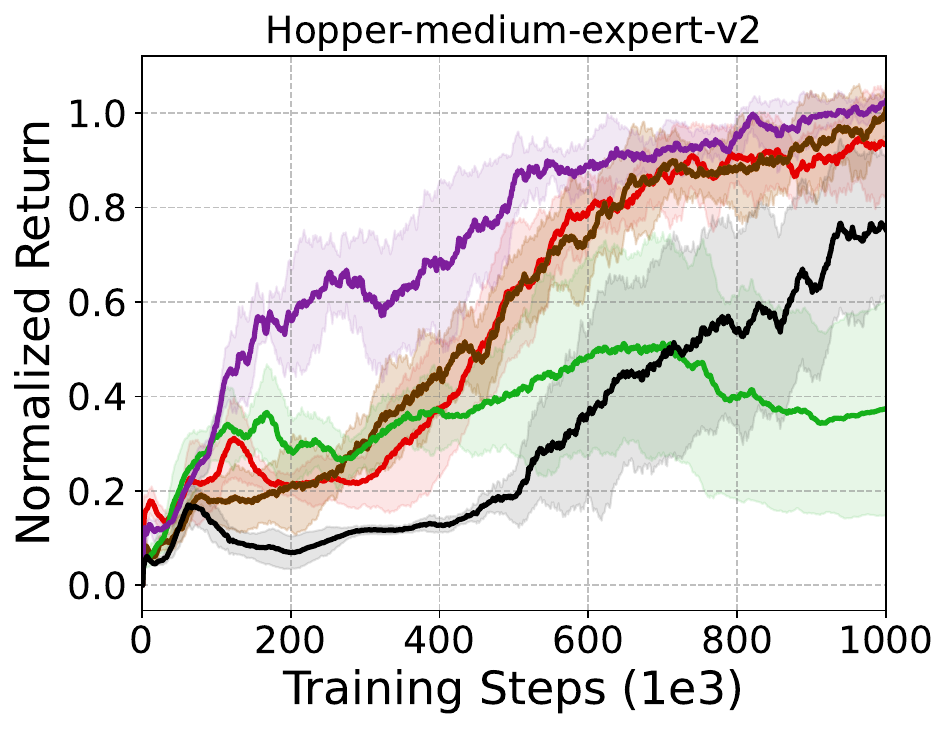}
    \caption{Offline learning curves of different methods across five seeds on MuJoCo locomotion tasks. The solid lines and shaded regions represent mean and standard deviation, respectively.}
    \label{fig:Appendix_exp_mujoco_offline}
\end{figure*}

Fig.~\ref{fig:Appendix_exp_mujoco_offline} shows the offline performance of different methods on MuJoCo locomotion tasks. We observe that on certain datasets such as HalfCheetah-medium-v2 and HalfCheetah-medium-replay-v2, ENOTO-CQL exhibits a slight performance improvement compared to CQL. This indicates that the introduction of Q-ensembles indeed has some benefits for the performance in the offline stage. However, it is important to note that the use of Q-ensembles can impact the convergence speed. For instance, on the Hopper-medium-expert-v2 dataset, ENOTO-CQL demonstrates a noticeably slower convergence speed compared to CQL. Nevertheless, if both algorithms are allowed to continue training, for example, for 3M training steps, we believe that the performance of ENOTO-CQL can still surpass that of CQL.

Although different offline RL algorithms may have varying final performance in the offline stage, with some methods potentially performing worse than ENOTO-CQL, our comparisons in the online stage remain fair. On one hand, the introduction of Q-ensembles can enhance the performance of existing algorithms in the offline stage, which is a inherent advantage of Q-ensembles. On the other hand, higher performance in the offline stage can actually lead to performance drop in the online stage, while lower performance in the offline stage is less prone to such drop. Our proposed ENOTO framework ensures that the offline policy maintains high performance from the offline stage, and achieves rapid performance improvement in the online stage without encountering performance drop.

\paragraph{Navigation Tasks}

\begin{figure*}[h]
    \centering
    \includegraphics[width=0.3\textwidth]{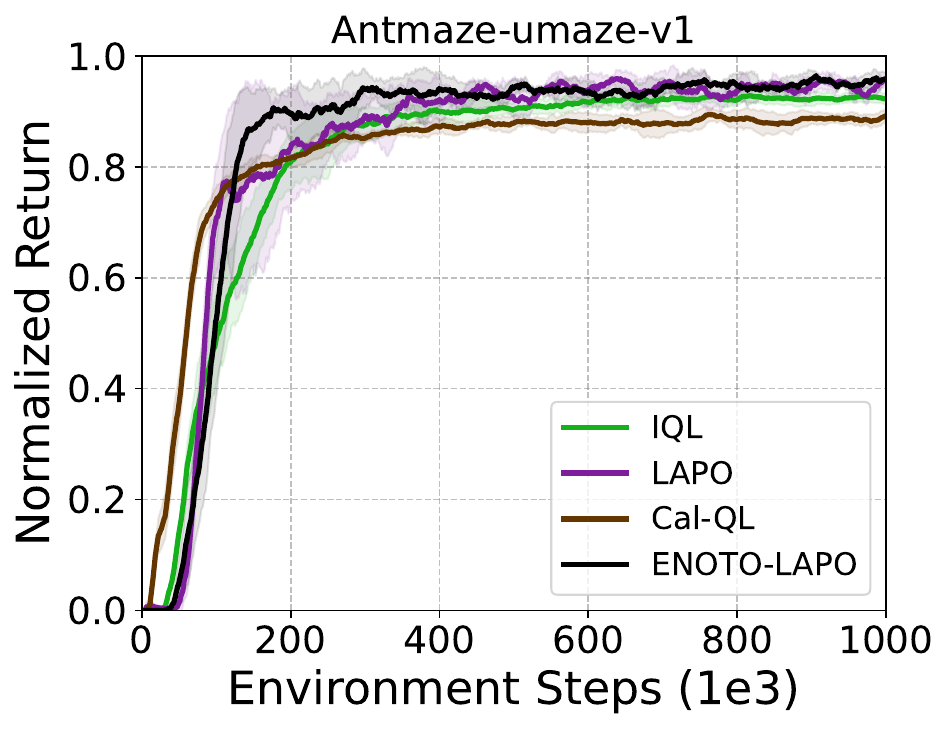}
    \includegraphics[width=0.3\textwidth]{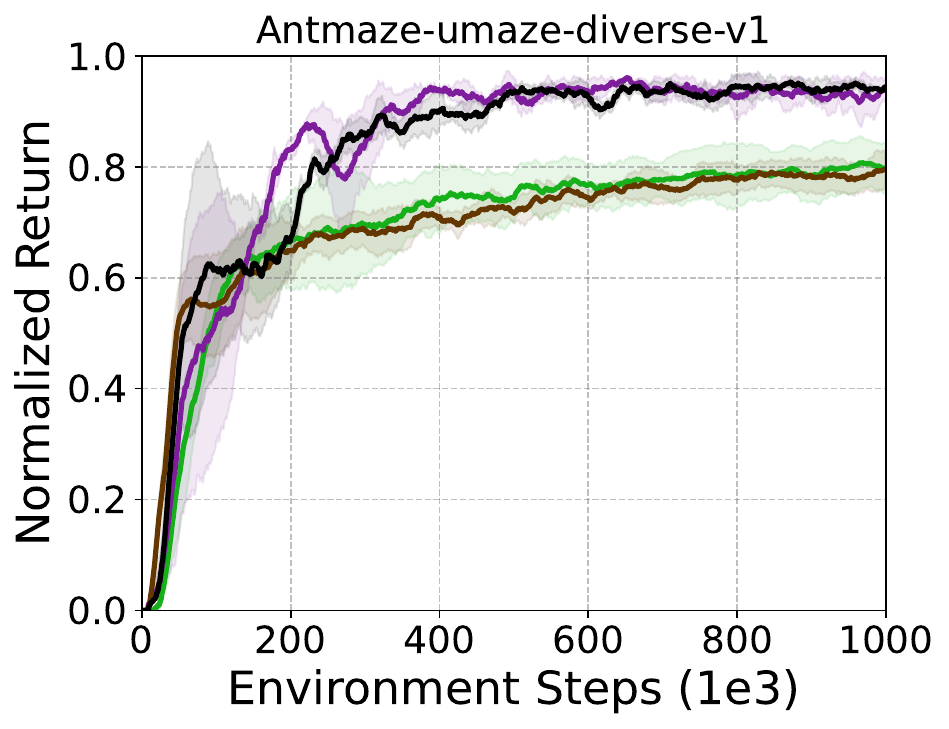}
    \includegraphics[width=0.3\textwidth]{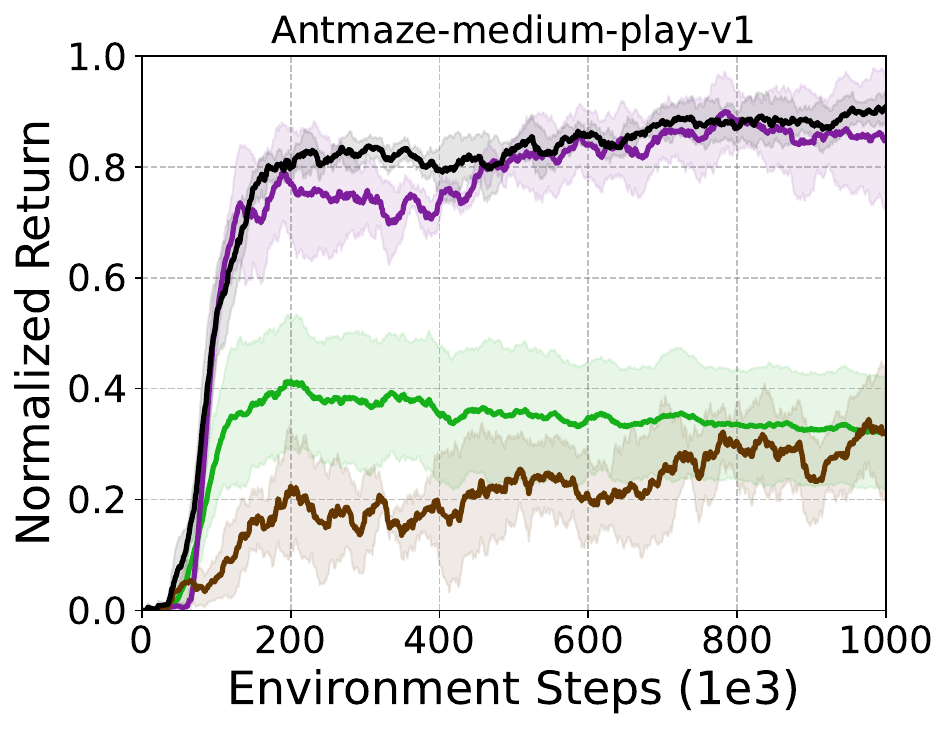}
    \includegraphics[width=0.3\textwidth]{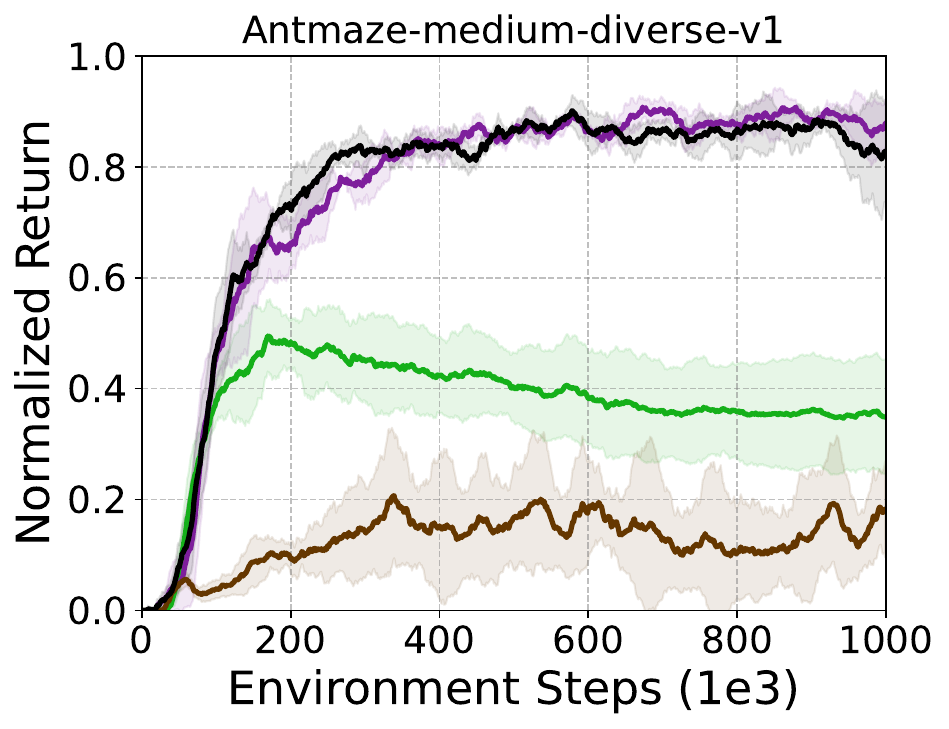}
    \includegraphics[width=0.3\textwidth]{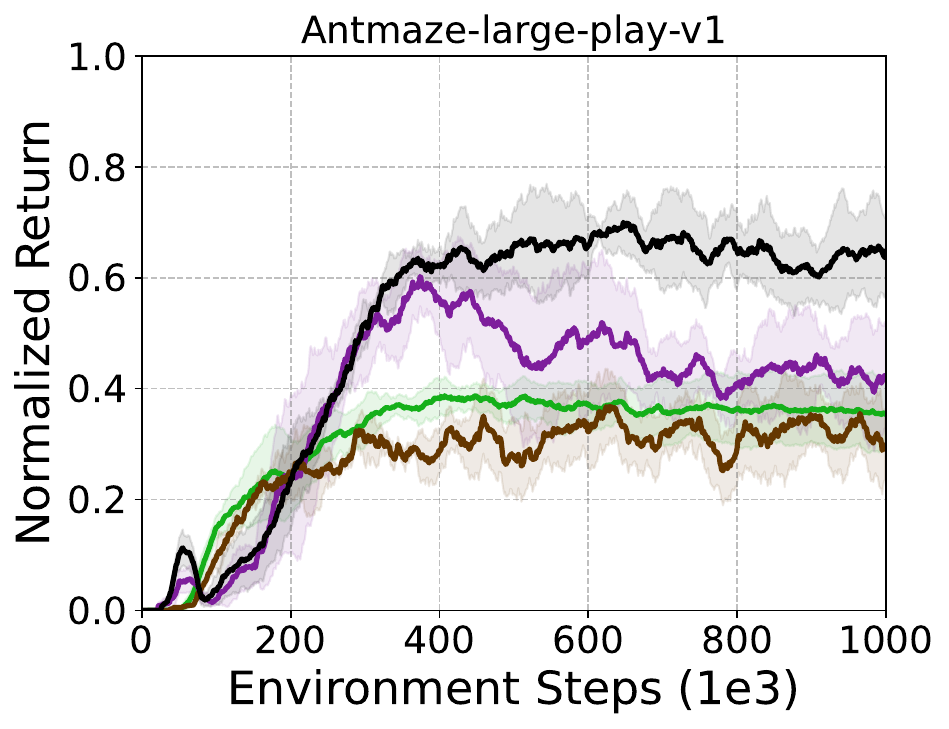}
    \includegraphics[width=0.3\textwidth]{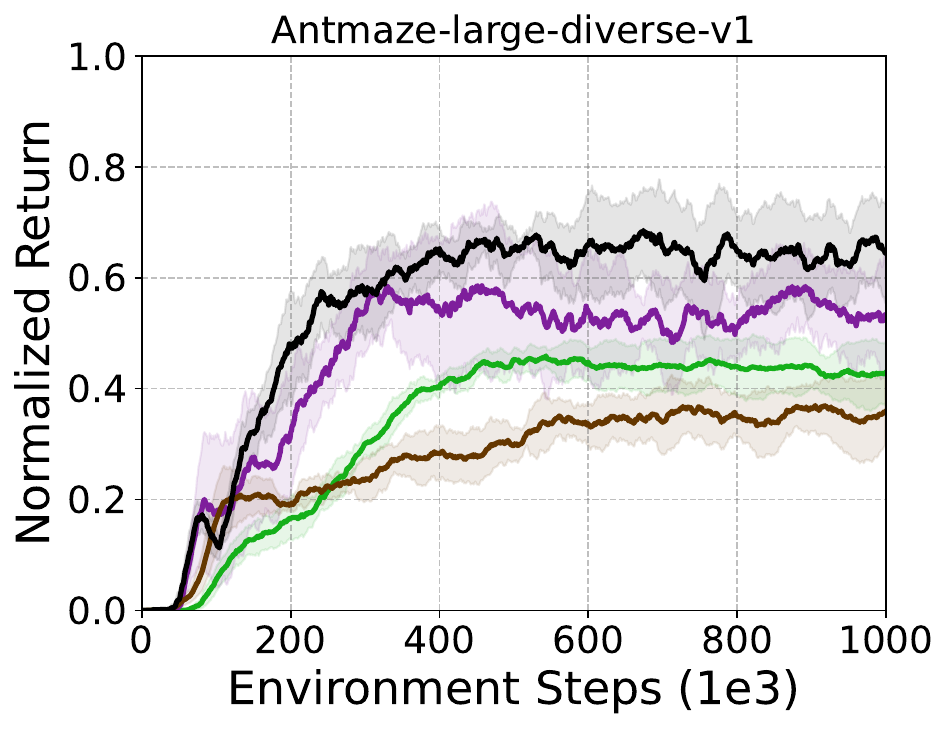}
    \caption{Offline learning curves of different methods across five seeds on Antmaze navigation tasks. The solid lines and shaded regions represent mean and standard deviation, respectively.}
    \label{fig:Appendix_exp_antmaze_offline}
\end{figure*}

Fig.~\ref{fig:Appendix_exp_antmaze_offline} displays the offline performance of various methods on Antmaze navigation tasks. Firstly, we observe that LAPO outperforms IQL in terms of offline performance, providing a higher starting point for the online phase. This is particularly evident in the umaze and medium maze environments, where LAPO nearly reaches the performance ceiling. Regarding LAPO and ENOTO-LAPO, since LAPO achieves near-optimal performance in simple environments such as umaze and medium mazes, their offline performance is comparable. However, in the more challenging large maze environment, the inclusion of Q-ensembles enables ENOTO-LAPO to surpass LAPO in terms of performance.

\end{document}